\documentclass[sigconf]{acmart}

\usepackage{multirow}
\usepackage{graphicx}
\usepackage{subfigure}

\AtBeginDocument{%
\providecommand\BibTeX{{%
\normalfont B\kern-0.5em{\scshape i\kern-0.25em b}\kern-0.8em\TeX}}}

\copyrightyear{2020} 
\acmYear{2020} 
\setcopyright{acmcopyright}\acmConference[KDD '20]{Proceedings of the 26th ACM SIGKDD Conference on Knowledge Discovery and Data Mining USB Stick}{August 23--27, 2020}{Virtual Event, USA}
\acmBooktitle{Proceedings of the 26th ACM SIGKDD Conference on Knowledge Discovery and Data Mining USB Stick (KDD '20), August 23--27, 2020, Virtual Event, USA}
\acmPrice{15.00}
\acmDOI{10.1145/3394486.3403076}
\acmISBN{978-1-4503-7998-4/20/08}

\begin{document}
\fancyhead{}
\title{Towards Deeper Graph Neural Networks}


\author{Meng Liu}
\affiliation{%
\institution{Texas A\&M University}
\city{College Station}
\state{TX}
}
\email{mengliu@tamu.edu}

\author{Hongyang Gao}
\affiliation{%
\institution{Texas A\&M University}
\city{College Station}
\state{TX}
}
\email{hongyang.gao@tamu.edu}

\author{Shuiwang Ji}
\affiliation{%
\institution{Texas A\&M University}
\city{College Station}
\state{TX}
}
\email{sji@tamu.edu}



\begin{abstract}


Graph neural networks have shown significant success in the field of graph representation learning. Graph convolutions perform neighborhood aggregation and represent one of the most important graph operations. Nevertheless, one layer of these neighborhood aggregation methods only consider immediate neighbors, and the performance decreases when going deeper to enable larger receptive fields. Several recent studies attribute this performance deterioration to the over-smoothing issue, which states that repeated propagation makes node representations of different classes indistinguishable. In this work, we study this observation systematically and develop new insights towards deeper graph neural networks. First, we provide a systematical analysis on this issue and argue that the key factor compromising the performance significantly is the entanglement of representation transformation and propagation in current graph convolution operations. After decoupling these two operations, deeper graph neural networks can be used to learn graph node representations from larger receptive fields. We further provide a theoretical analysis of the above observation when building very deep models, which can serve as a rigorous and gentle description of the over-smoothing issue. Based on our theoretical and empirical analysis, we propose Deep Adaptive Graph Neural Network (DAGNN) to adaptively incorporate information from large receptive fields. A set of experiments on citation, co-authorship, and co-purchase datasets have confirmed our analysis and insights and demonstrated the superiority of our proposed methods.

\end{abstract}

\ccsdesc[500]{Mathematics of computing~Graph algorithms}
\ccsdesc[500]{Computing methodologies~Artificial intelligence}
\ccsdesc[500]{Computing methodologies~Neural networks}

\keywords{deep learning, graph representation learning, graph neural networks}


\maketitle

\section{Introduction}
Graphs, representing entities and their relationships, are
ubiquitous in the real world, such as social networks, point
clouds, traffic networks, knowledge graphs, and molecular
structures. Recently, many studies focus on developing deep
learning approaches for graph data, leading to rapid development
in the field of graph neural networks. Great successes have been
achieved for many applications, such as node
classification~\cite{kipf2016semi,hamilton2017inductive,velivckovic2017graph,monti2017geometric,gao2018large,xu2018representation,klicpera2018predict,wu2019simplifying},
graph
classification~\cite{gilmer2017neural,ying2018hierarchical,zhang2018end,xu2018powerful,gao2019graph,lee2019self,ma2019graph,Yuan2020StructPool:}
and link prediction~\cite{zhang2017weisfeiler,zhang2018link, cai2020multi}.
Graph convolutions adopt a neighborhood aggregation (or message
passing) scheme to learn node representations by considering the
node features and graph topology information together, among
which the most representative method is Graph Convolutional
Networks (GCNs)~\cite{kipf2016semi}. GCN learns
representation for a node by aggregating representations of its
neighbors iteratively. However, a common challenge faced by GCN
and most other graph convolutions is that one layer of graph
convolutions only consider immediate neighbors and the performance
degrades greatly when we apply multiple layers to leverage large
receptive fields. Several recent works attribute this performance
degradation to the over-smoothing
issue~\cite{li2018deeper,xu2018representation,chen2019measuring},
which states that representations from different classes become
inseparable due to repeated propagation. In this work, we study
this performance deterioration systematically and develop new
insights towards deeper graph neural networks.

We first systematically analyze the performance
degradation when stacking multiple GCN layers by using our
quantitative metric for node representation smoothness
measurement and a data visualization technique. We observe and
argue that the main factor compromising the performance greatly
is the entanglement of representation transformation and propagation.
After decoupling these two operations, it is demonstrated that
deeper graph neural networks can be deployed to learn graph node
representations from larger receptive fields without suffering
from performance deterioration. The over-smoothing issue is shown
to affect performance only when extremely large receptive fields
are utilized. We further give a theoretical analysis of the above
observation when building very deep models, which shows that
graph node representations will become indistinguishable when
depth goes infinity. This aligns with the over-smoothing issue.
The previous descriptions of the over-smoothing issue simplify
the assumption of non-linear activation function~\cite{li2018deeper,xu2018representation} or make approximations
of different probabilities~\cite{xu2018representation}. Our
theoretical analysis can serve as a more rigorous and gentle
description of the over-smoothing issue. Based on our theoretical
and empirical analysis, we propose an efficient and effective network, termed as Deep Adaptive Graph Neural Network, to learn node
representations by adaptively incorporating information from large receptive fields. Extensive experiments on citation,
co-authorship, and co-purchase datasets demonstrate the
reasonability of our insights and the superiority of our proposed network.

\section{Background and Related Works}

First, we introduce our notations used throughout this paper.
Generally, we let bold uppercase letters represent matrices and
bold lowercase letters denote vectors. A graph is formally
defined as $G = (V, E)$, where $V$ is the set
of nodes (vertices) that are indexed from $1$ to $n$ and $E \subseteq V \times V$ is the set of edges between nodes in $V$. $n = |V|$ and $m=|E|$ are the numbers
of nodes and edges, respectively. In this paper, we consider unweighted and undirected graphs. The topology information of
the whole graph is described by the adjacency matrix
$\boldsymbol{A} \in \mathbb{R}^{n \times n}$, where
$\boldsymbol{A}_{(i,j)}=1$ if an edge exists between node $i$ and node $j$, otherwise 0. The diagonal
matrix of node degrees are denoted as $\boldsymbol{D} \in
\mathbb{R}^{n \times n} $, where $\boldsymbol{D}_{(i,i)}=\sum_j
\boldsymbol{A}_{(i,j)}$. $\mathcal{N}_i$ denotes the neighboring
nodes set of node $i$. An attributed graph has a node feature
matrix $\boldsymbol{X} \in \mathbb{R}^{n \times d}$, where each
row $\boldsymbol{x}_i \in \mathbb{R}^d$ represents the feature
vector of node $i$ and $d$ is the dimension of node features.

\subsection{Graph Convolution Operations}

Most popular graph convolution operations follow a neighborhood
aggregation (or message passing) fashion to learn a node
representation by propagating representations of its neighbors
and applying transformation after that. The $\ell$-th layer of
a general graph convolution can be described as
\begin{equation}
\begin{aligned}
\boldsymbol{a}_i^{(\ell)} & = {\rm PROPAGATION}^{(\ell)}\left(\left\{\boldsymbol{x}_i^{(\ell-1)},\{\boldsymbol{x}_j^{(\ell-1)} | j\in \mathcal{N}_i\}\right\}\right)
\\
\boldsymbol{x}_i^{(\ell)} &= {\rm TRANSFORMATION}^{(\ell)}\left(\boldsymbol{a}_i^{(\ell)}\right).
\end{aligned}
\label{eq:gnn}
\end{equation}
$\boldsymbol{x}_i^{(\ell)}$ is the representation of node $i$ at
$l$-th layer and $\boldsymbol{x}_i^{(0)}$ is initialized as node
feature $\boldsymbol{x}_i$. Most graph convolutions, like
GCN~\cite{kipf2016semi}, GraphSAGE~\cite{hamilton2017inductive}, GAT~\cite{velivckovic2017graph}, and GIN~\cite{xu2018powerful}, can be obtained under this
framework by deploying different propagation and transformation
mechanisms.

Without losing generalization, we focus on the Graph
Convolutional Network (GCN)~\cite{kipf2016semi}, the most
representative graph convolution operation, in the following
analysis. The $\ell$-th layer forward-propagation process is
formulated as
\begin{equation}\label{GCN_EQ}
\boldsymbol{X}^{(\ell)} = \sigma\left(\widehat{\boldsymbol{A}} \boldsymbol{X}^{(\ell-1)}\boldsymbol{W}^{(\ell)}\right),
\end{equation}
where $\boldsymbol{X}^{(\ell)} \in \mathbb{R}^{n \times
d^{(\ell)}}$ and $\boldsymbol{X}^{(\ell-1)} \in \mathbb{R}^{n
\times d^{(\ell-1)}}$ are the output and input node
representation matrices of layer $\ell$.
$\widehat{\boldsymbol{A}} =
\widetilde{\boldsymbol{D}}^{-\frac{1}{2}}\widetilde{\boldsymbol{A}}\widetilde{\boldsymbol{D}}^{-\frac{1}{2}}$,
where $\widetilde{\boldsymbol{A}}=\boldsymbol{A}+\boldsymbol{I}$
is the adjacency matrix with added self-connections.
$\widetilde{\boldsymbol{D}}_{(i,i)}=\sum_{j}\widetilde{\boldsymbol{A}}_{(i,j)}$
is the diagonal node degree matrix. $\boldsymbol{W}^{(\ell)} \in
\mathbb{R}^{d^{(\ell-1)} \times d^{(\ell)}}$ is a layer-specific
trainable weight matrix. $\sigma$ is a non-linear activation
function like ReLU~\cite{nair2010rectified}. Intuitively, GCN
learns representation for each node by propagating neighbors'
representations and conducting non-linear transformation after
that. GCN is originally applied for semi-supervised
classification, where only partial nodes have training labels in
a graph. Thanks to the propagation process, representation of a
labeled node carries the information from its neighbors that are
usually unlabeled, thus training signals can be propagated to the
unlabeled nodes.

\subsection{Related Works}
From~Eq.(\ref{GCN_EQ}), one layer GCN only considers immediate
neighbors, i.e. one-hop neighborhood. Multiple layers should be
applied if multi-hop neighborhood is needed. In practice,
however, the performance of GCN degrades greatly when multiple
layers are stacked. Several works reveal that stacking many
layers can bring the over-smoothing issue, which means that
representations of nodes converge to indistinguishable limits. To
our knowledge, ~\cite{li2018deeper} is the first attempt to
demystify the over-smoothing issue in the GCN model. The authors
first demonstrate that the propagation process of the GCN model
is a special symmetric form of Laplacian
smoothing~\cite{taubin1995signal}, which makes the
representations of nodes in the same class similar, thus
significantly easing the classification task. Then they show that
repeatedly stacking many layers may make representations of nodes
from different classes indistinguishable. The same problem is
studied in~\cite{xu2018representation} by analyzing the
connection of nodes' influence distribution and random
walk~\cite{lovasz1993random}. Recently,
SGC~\cite{wu2019simplifying} is proposed by reducing unnecessary
complexity in GCN. The authors show that SGC corresponds to a
low-pass-type filter on the spectral domain, thus deriving
smoothing features across a graph. Another recent
work~\cite{chen2019measuring} verify that smoothing is the nature
of most typical graph convolutions. It is showed that reasonable
smoothing makes graph convolutions work and over-smoothing
results in poor performance.

Due to the potential concern of the over-smoothing issue, a
limited neighborhood is usually used in practice and it is
difficult to extend. However, long-range dependencies should be
taken into consideration, especially for peripheral nodes. Also,
small receptive fields are not enough to propagate training
signals to the whole graph when the number of training nodes is
limited under a semi-supervised learning setting.
~\cite{li2018deeper} applies co-training and self-training to
overcome the limitation of shallow architectures. A smoothness
regularizer term and adaptive edge optimization are proposed in
~\cite{chen2019measuring} to relieve the over-smoothing problem.
Jumping Knowledge Network~\cite{xu2018representation} deploys a
layer-aggregation mechanism to adaptively select a node’s
sub-graph features at different ranges rather than to capture
equally smoothed representations for all nodes.
~\cite{klicpera2018predict} utilizes the relationship between GCN
and PageRank~\cite{page1999pagerank} to develop a propagation
mechanism based on personalize PageRank, which can preserve the
node's local information while gather information from a large
neighborhood. Recently, Geom-GCN~\cite{pei2020geom} and non-local
GNNs~\cite{liu2020non} are proposed to capture long-range
dependencies for disassortative graph by designing non-local
aggregators.

\begin{figure*}
\centering
\subfigure{
\includegraphics[width=0.205\columnwidth]{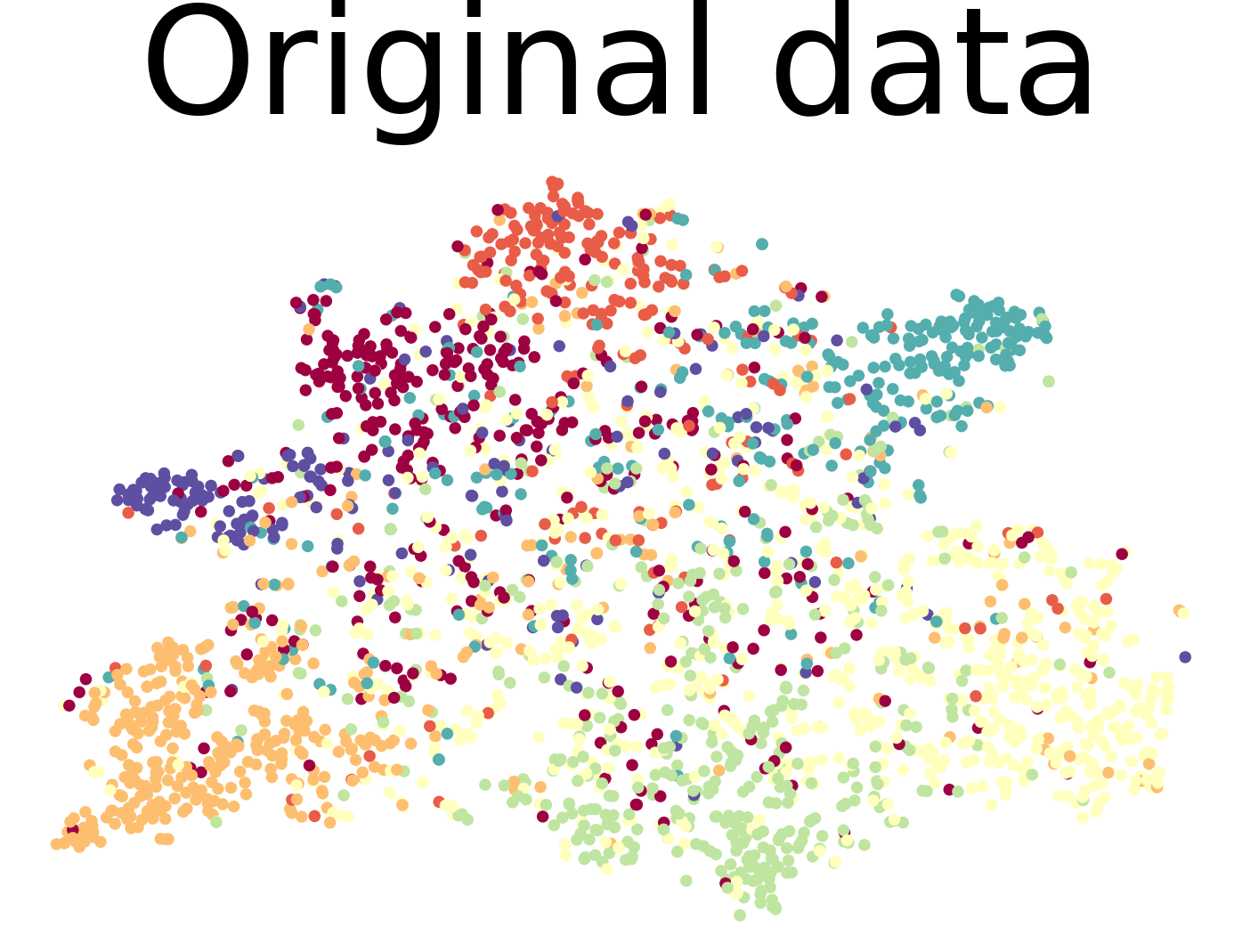}
}
\subfigure{
\includegraphics[width=0.205\columnwidth]{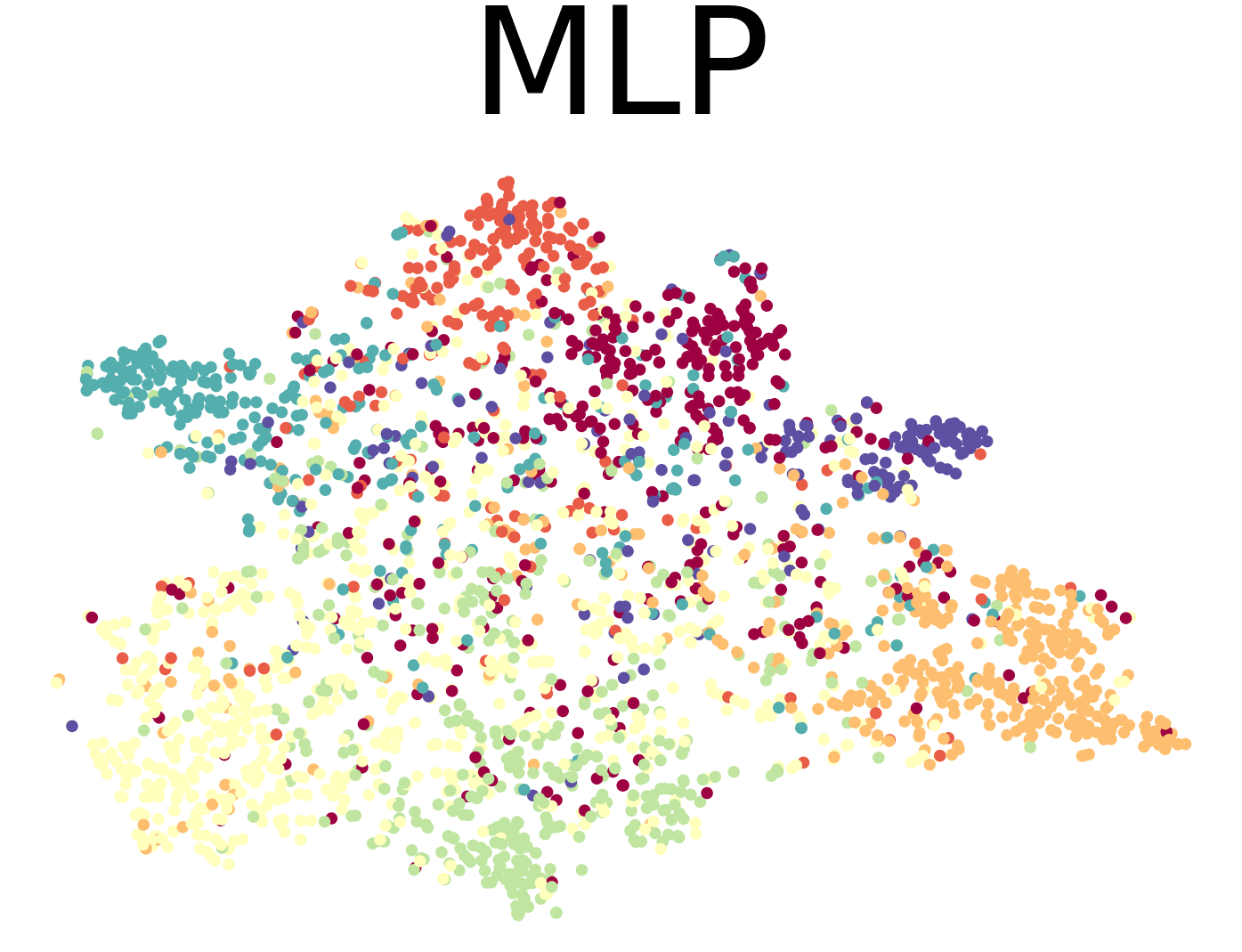}
}
\subfigure{
\includegraphics[width=0.205\columnwidth]{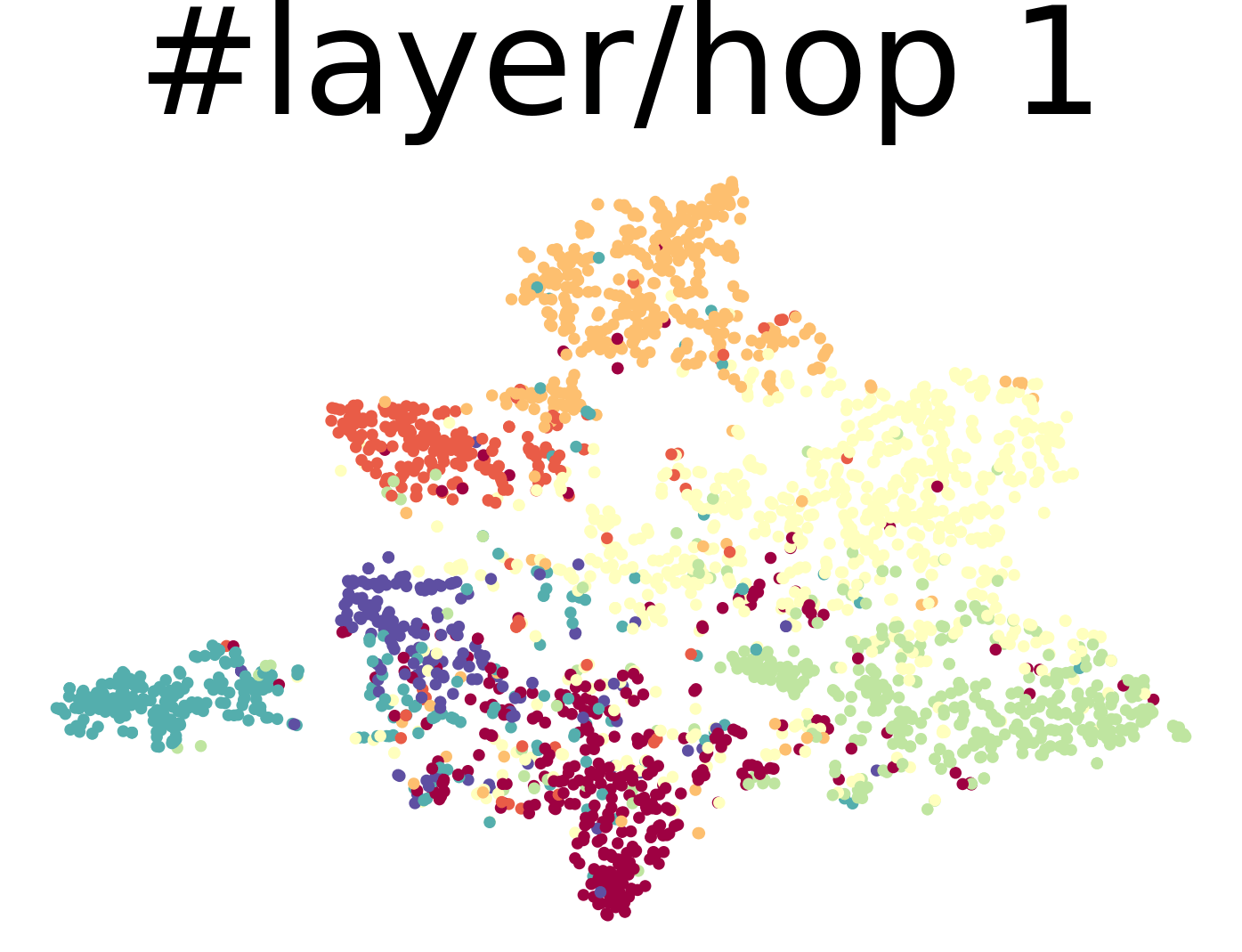}
}
\subfigure{
\includegraphics[width=0.205\columnwidth]{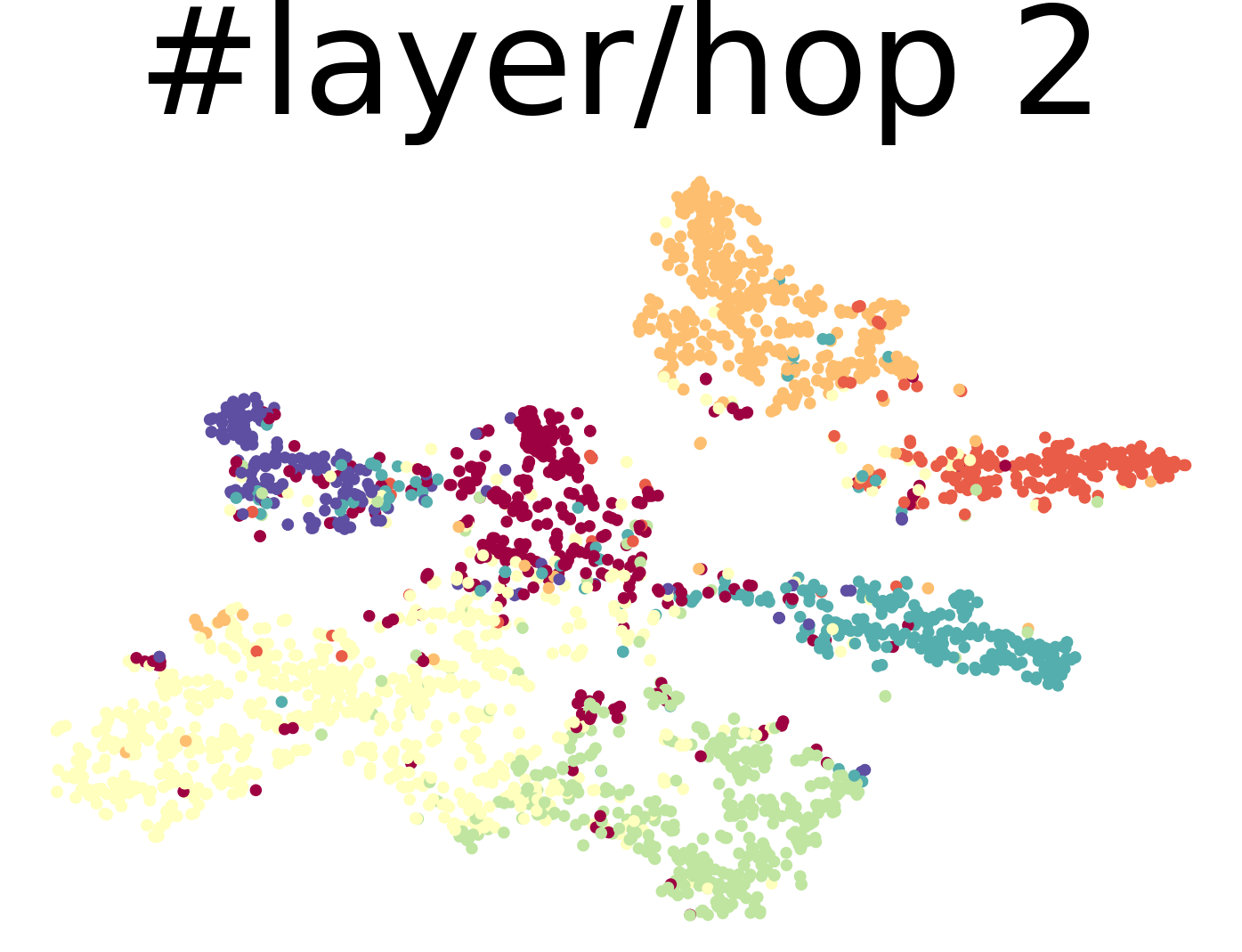}
}
\subfigure{
\includegraphics[width=0.205\columnwidth]{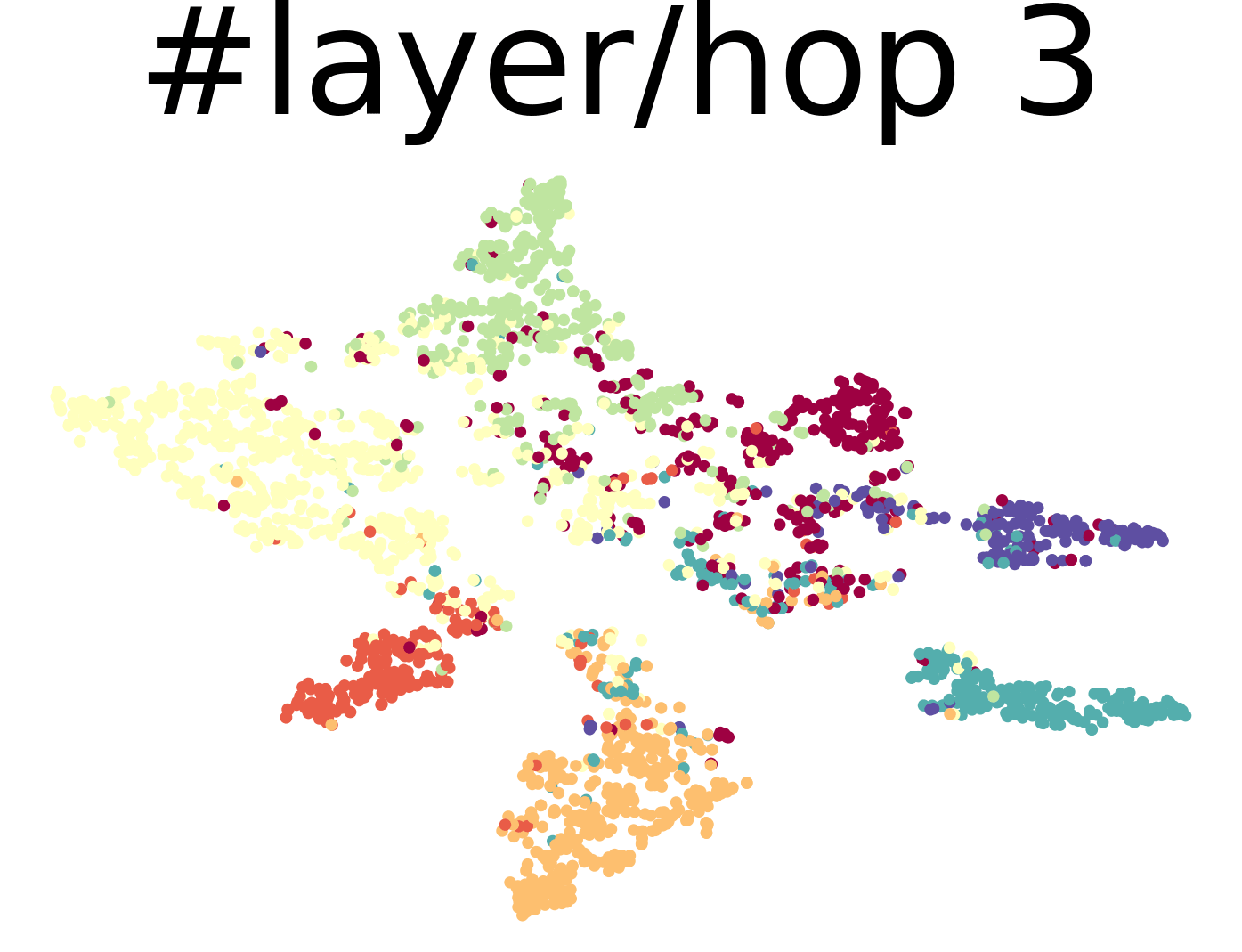}
}
\subfigure{
\includegraphics[width=0.205\columnwidth]{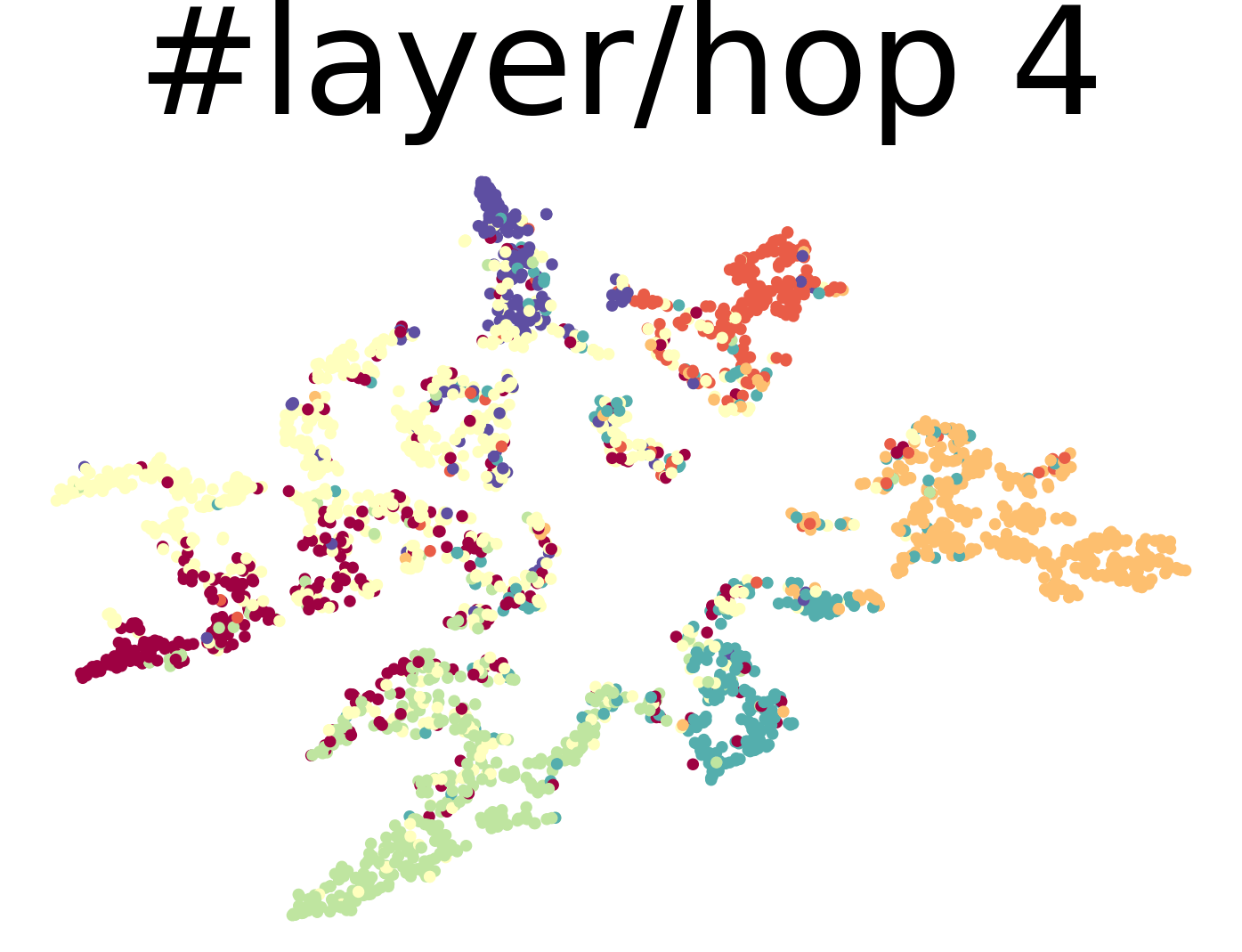}
}
\subfigure{
\includegraphics[width=0.205\columnwidth]{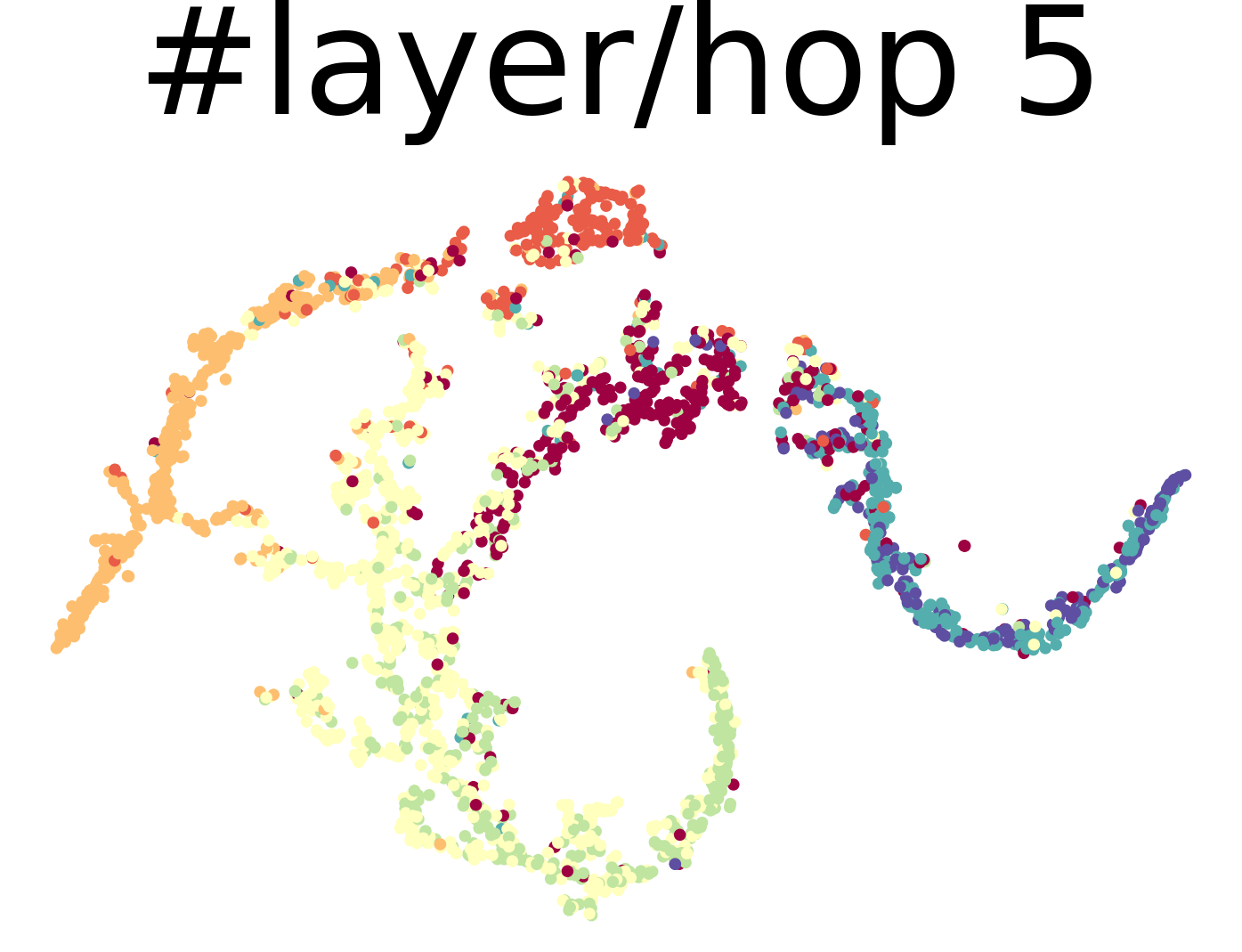}
}
\subfigure{
\includegraphics[width=0.205\columnwidth]{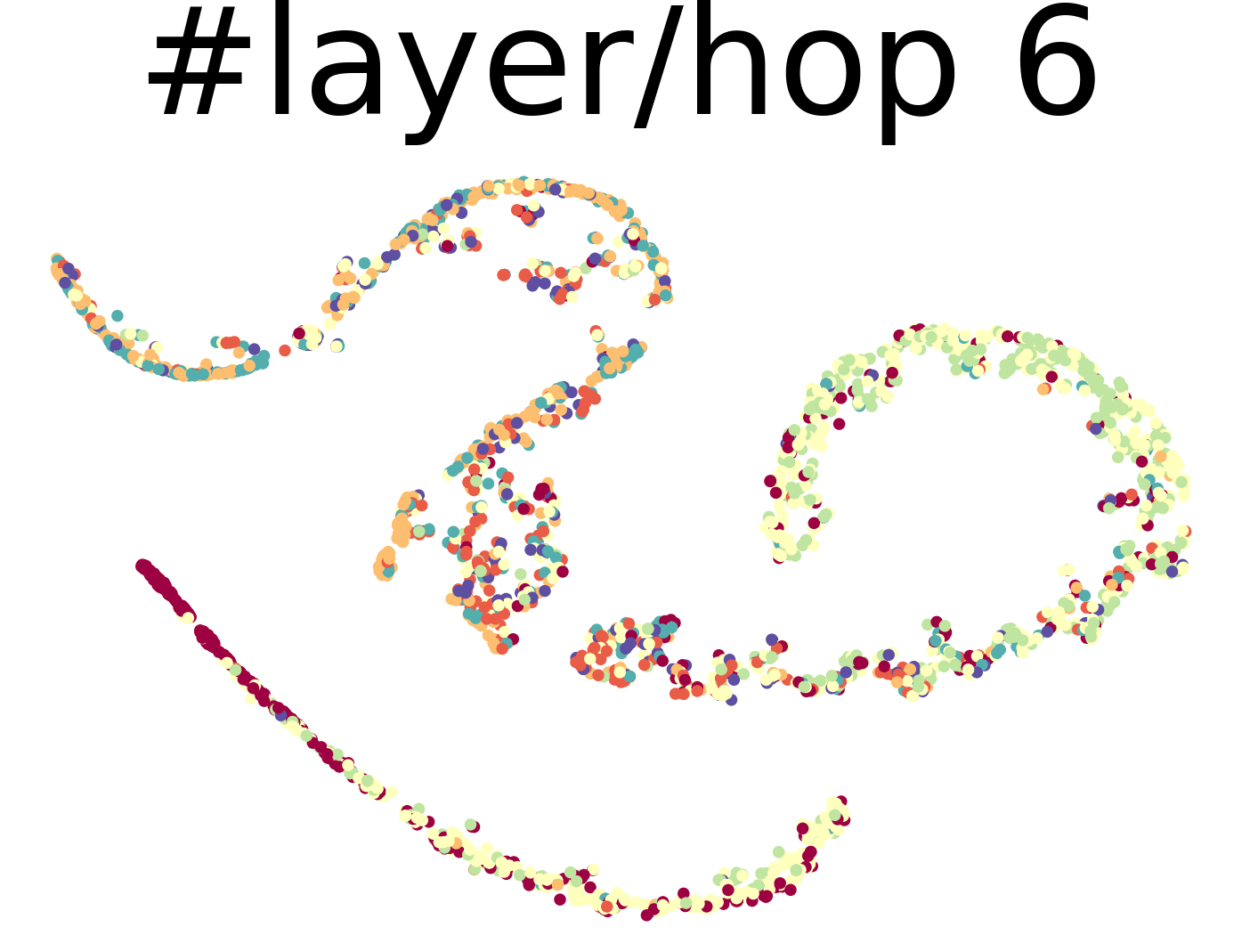}
}
\subfigure{
\includegraphics[width=0.205\columnwidth]{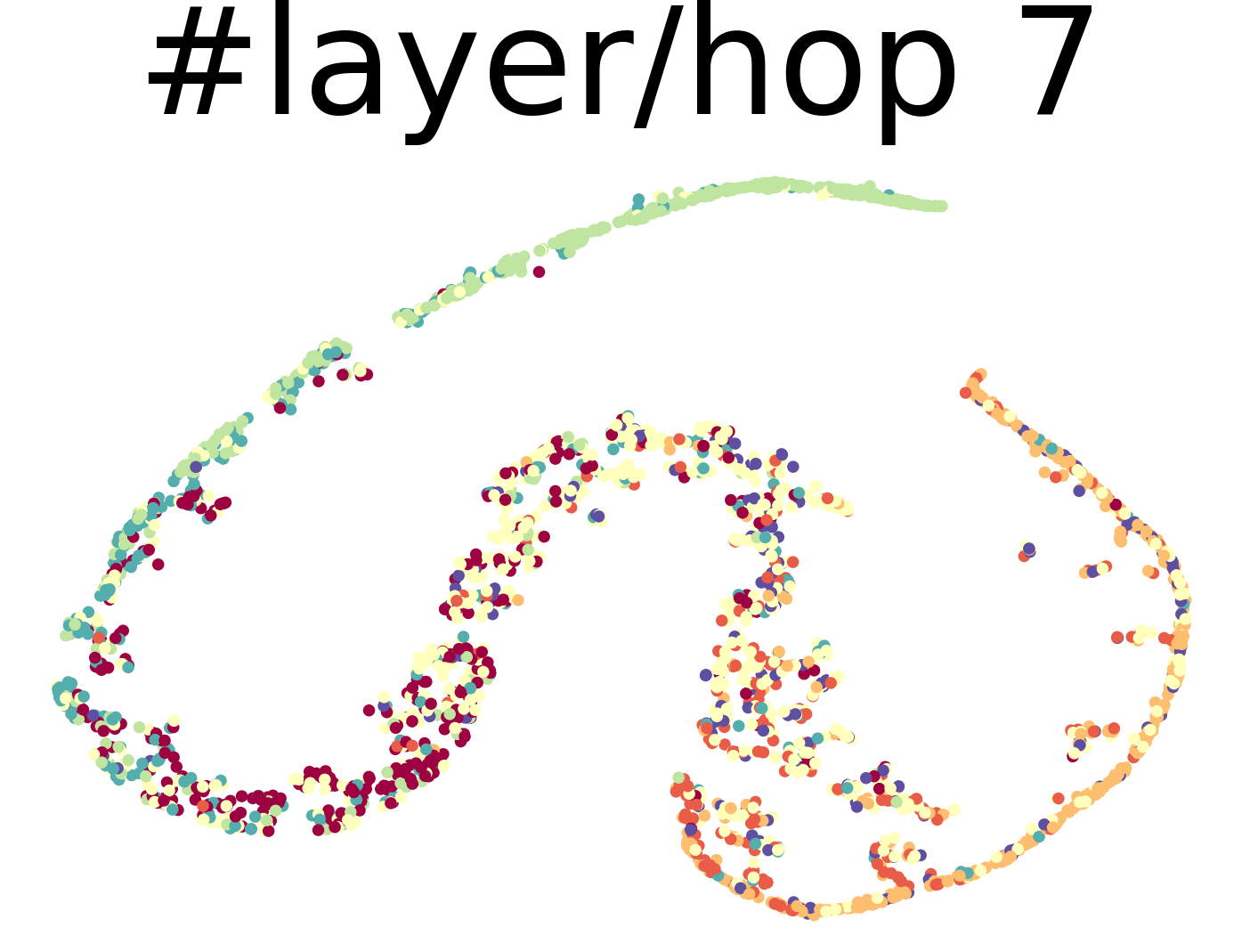}
}
\caption{t-SNE visualization of node representations derived by different numbers of GCN layers on Cora. Colors represent node classes.}
\label{fig:tsne1}
\end{figure*}

\section{Empirical and Theoretical Analysis of Deep GNNs}

In this section, we first propose a quantitative metric to
measure the smoothness of graph node representations. Then we
utilize this metric, along with a data visualization technique,
to rethink the performance degradation when utilizing GCN layer
to build deep graph neural networks. We observe and argue that
the entanglement of representation transformation and propagation is a
prominent factor that compromises the network performance. After
decoupling these two operations, deeper graph neural networks can be built to learn graph
node representations from large receptive fields without
suffering from performance degradation. The over-smoothing issue
is shown to influence the performance only when extremely large
receptive fields are adopted. Further, we provide a theoretical
analysis of the above observation when building very deep models,
which aligns with the conclusion of over-smoothing issue and can
serve as a rigorous description of the over-smoothing issue.

\subsection{Quantitative Metric for Smoothness}

Smoothness is a metric that reflects the similarity of node
representations. Here, we first define a similarity metric
between the representations of node $i$ and node $j$ with their
Euclidean distance:
\begin{equation}
\label{eq:sim}
D(\boldsymbol{x}_i,\boldsymbol{x}_j) = \frac{1}{2}\left\lVert\frac{\boldsymbol{x}_i}{\lVert\boldsymbol{x}_i\rVert}-\frac{\boldsymbol{x}_j}{\lVert\boldsymbol{x}_j\rVert}\right\rVert,
\end{equation}
where $\boldsymbol{x}_i$ is the feature representation of node
$i$ and $\lVert \cdot \rVert$ denotes                                                                                                                                                                                                                                                                                                                                                                                                                                                                                                                                                                                                                                                                                                                                                                                                                                                                                                                                                                                                                                                                                                                                                                                                                                                                                                                                                                                                                                                                                                                                                                                                                                                                                                                                                                                                                                                                                                                                                                                                                                                                                                                                                                                                                                                                                                                                                                                                                                                              the Euclidean norm. The
Euclidean distance is a simple but effective way to measure the
similarity of two representations, especially in high dimensional
space. Smaller Euclidean distance value incidates higher
similarity of two representations. To remove the influence of the
magnitude of feature representations, we use normalized node
representations to compute their Euclidean distance, thus
constraining $D(\boldsymbol{x}_i,\boldsymbol{x}_j)$ in the range
of $[0,1]$.

Based on the similarity metric in~Eq.(\ref{eq:sim}), we further
propose a smoothness metric $SMV_i$ for node $i$, which is
computed as the average distance between node $i$ to other nodes:
\begin{equation}
SMV_i = \frac{1}{n-1}\sum_{j\in V, j\neq i} D(\boldsymbol{x}_i,\boldsymbol{x}_j).
\end{equation}
Hence, $SMV_i$ measures the similarity of node $i$'s
representations to the entire graph. For instance, a node in the
periphery or a leaf node usually has a large smoothness metric
value. Further, we can use $SMV_G$ to represent the smoothness
metric value of the whole graph $G$. Its mathematical expression
is defined as:
\begin{equation}
SMV_G = \frac{1}{n} \sum_{i \in V} SMV_i.
\end{equation}
Here, $SMV_G$ is negatively related to the overall smoothness of
nodes' representations in graph $G$.

\begin{figure}[t]
\centering
\includegraphics[width=0.75\columnwidth]{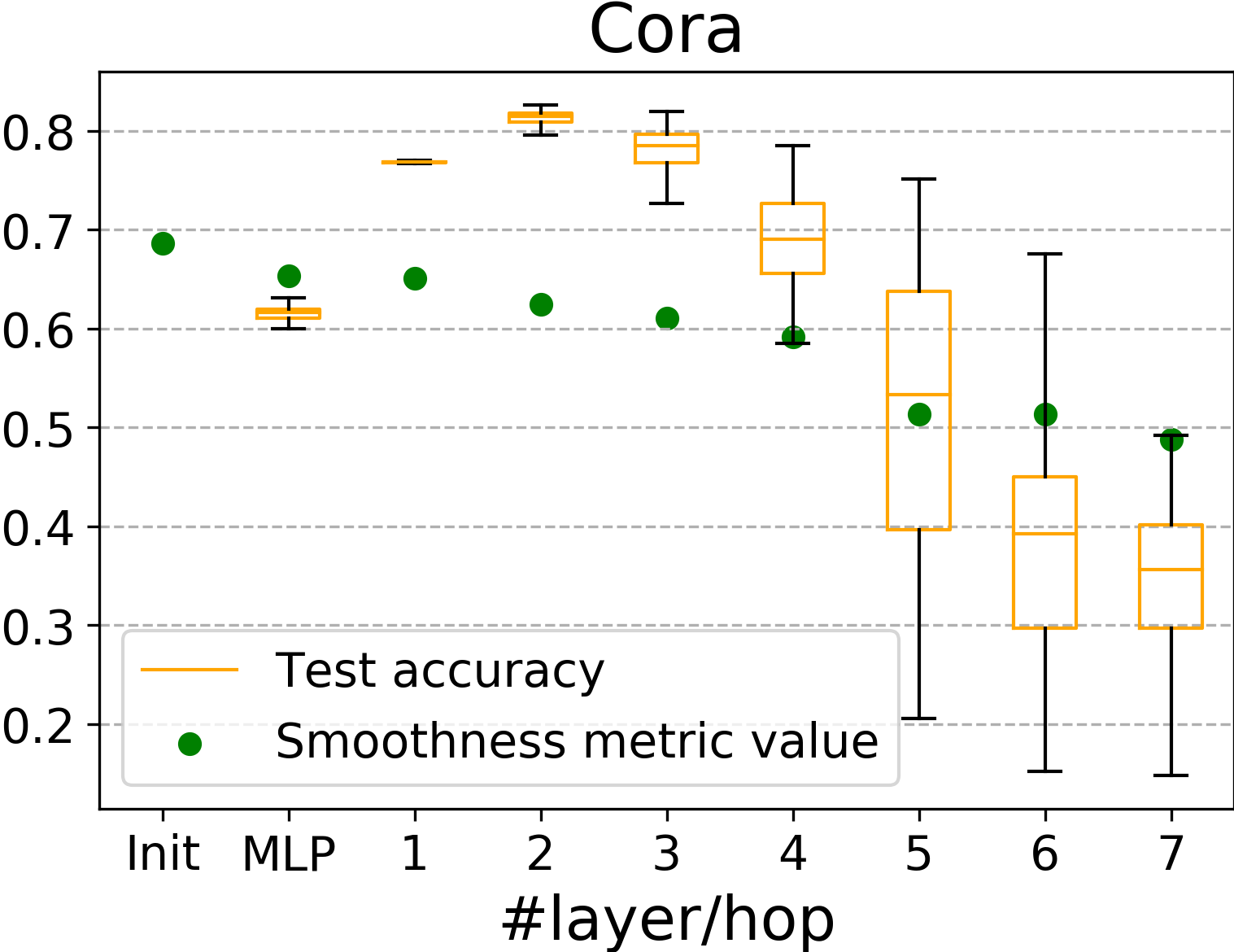}
\caption{Test accuracy and smoothness metric value of node representations with different numbers of GCN layers on Cora. "Init" means the smoothness metric value of the original data.}
\label{fig:gcn_degrade}
\end{figure}


\begin{figure*}
\centering
\subfigure{
\includegraphics[width=0.205\columnwidth]{./figures/cora_ori_tt.png}
}
\subfigure{
\includegraphics[width=0.205\columnwidth]{./figures/cora_mlp_tt.png}
}
\subfigure{
\includegraphics[width=0.205\columnwidth]{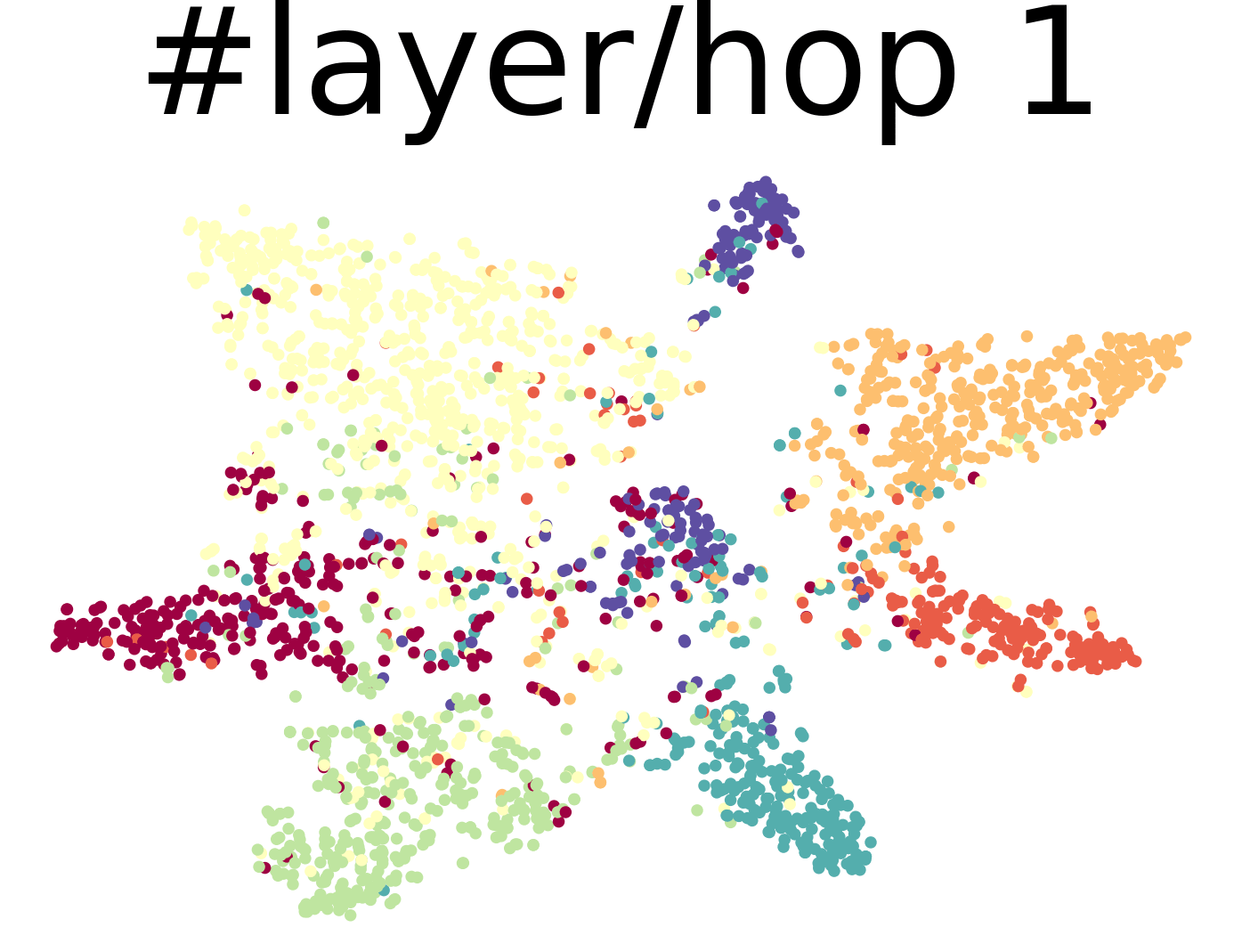}
}
\subfigure{
\includegraphics[width=0.205\columnwidth]{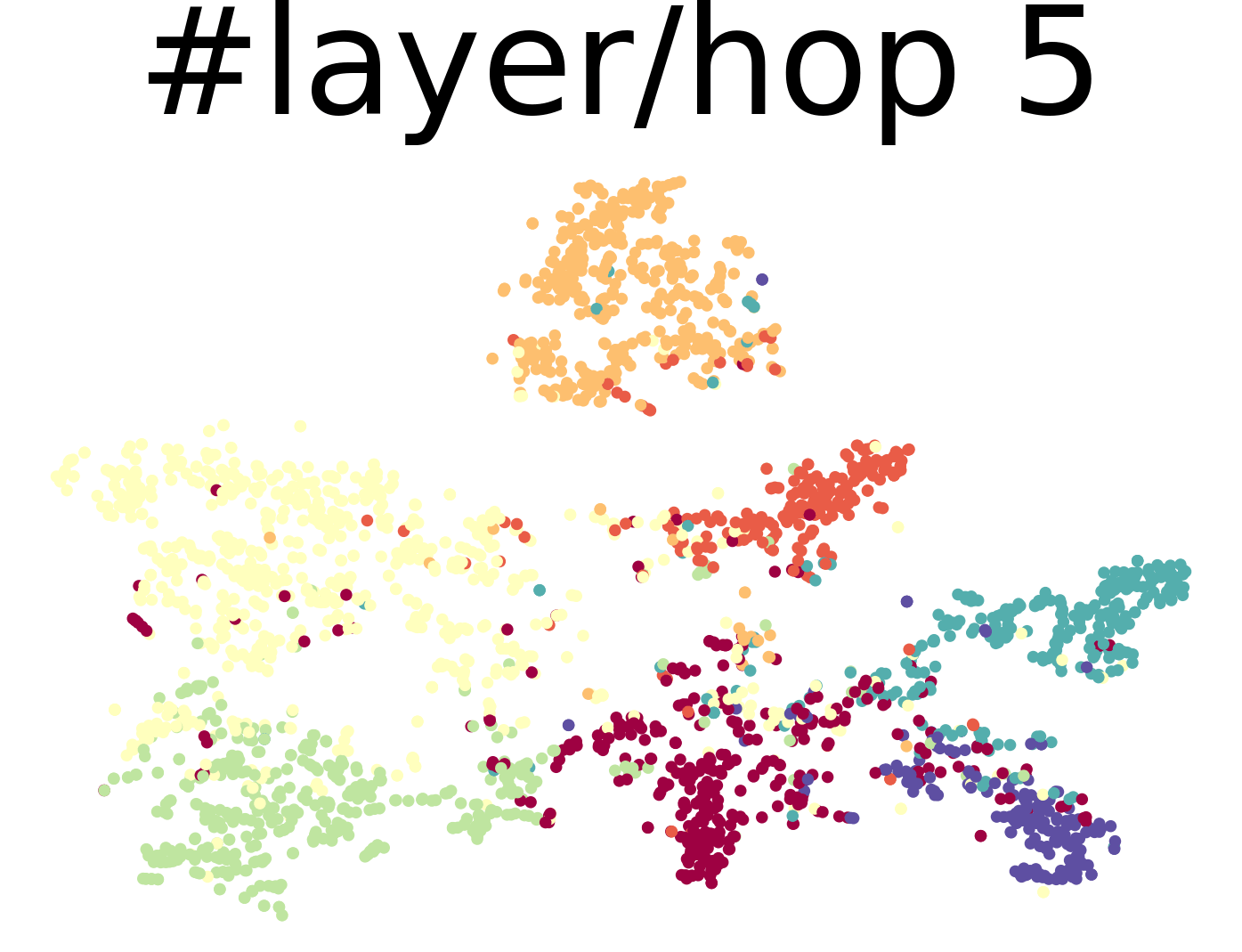}
}
\subfigure{
\includegraphics[width=0.205\columnwidth]{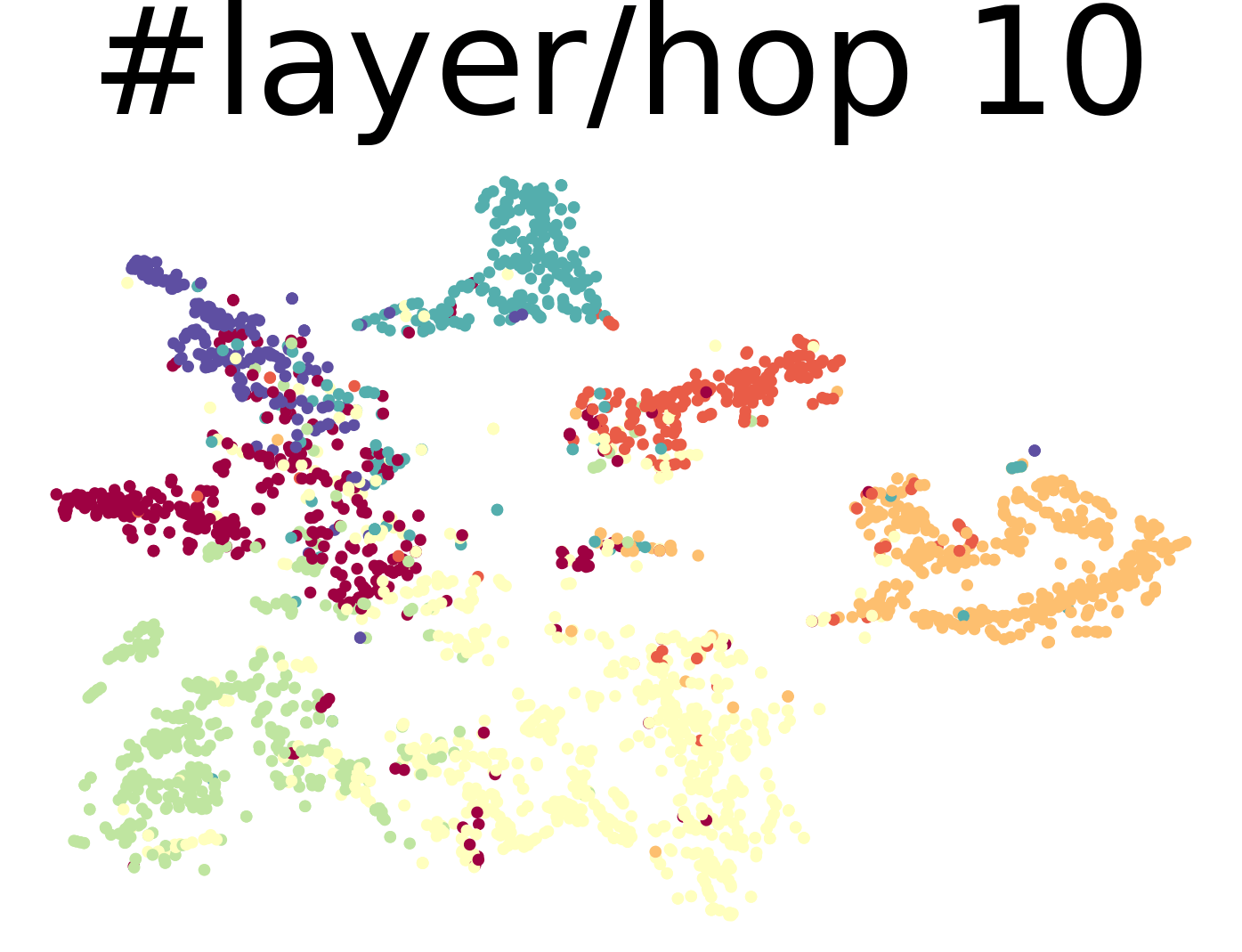}
}
\subfigure{
\includegraphics[width=0.205\columnwidth]{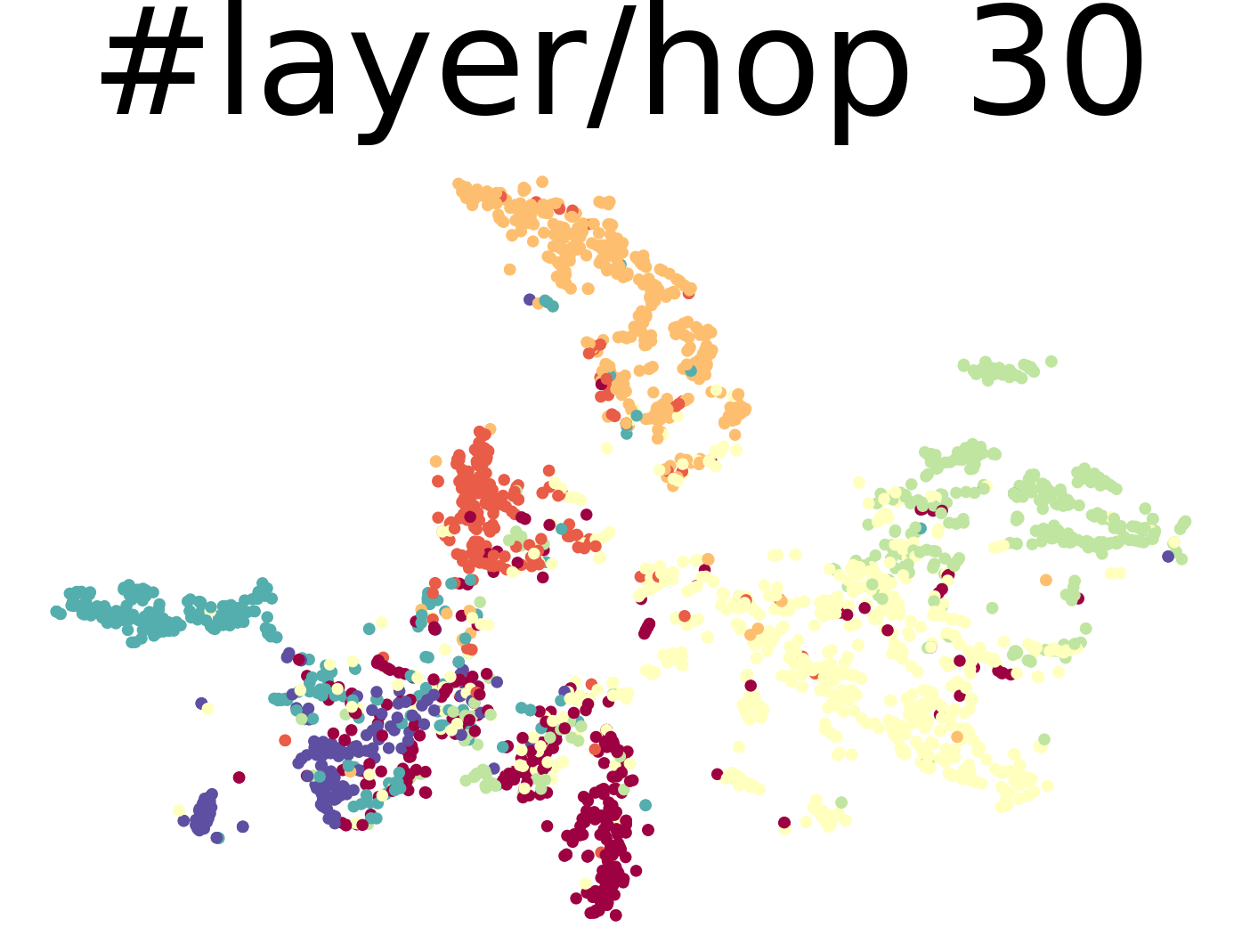}
}
\subfigure{
\includegraphics[width=0.205\columnwidth]{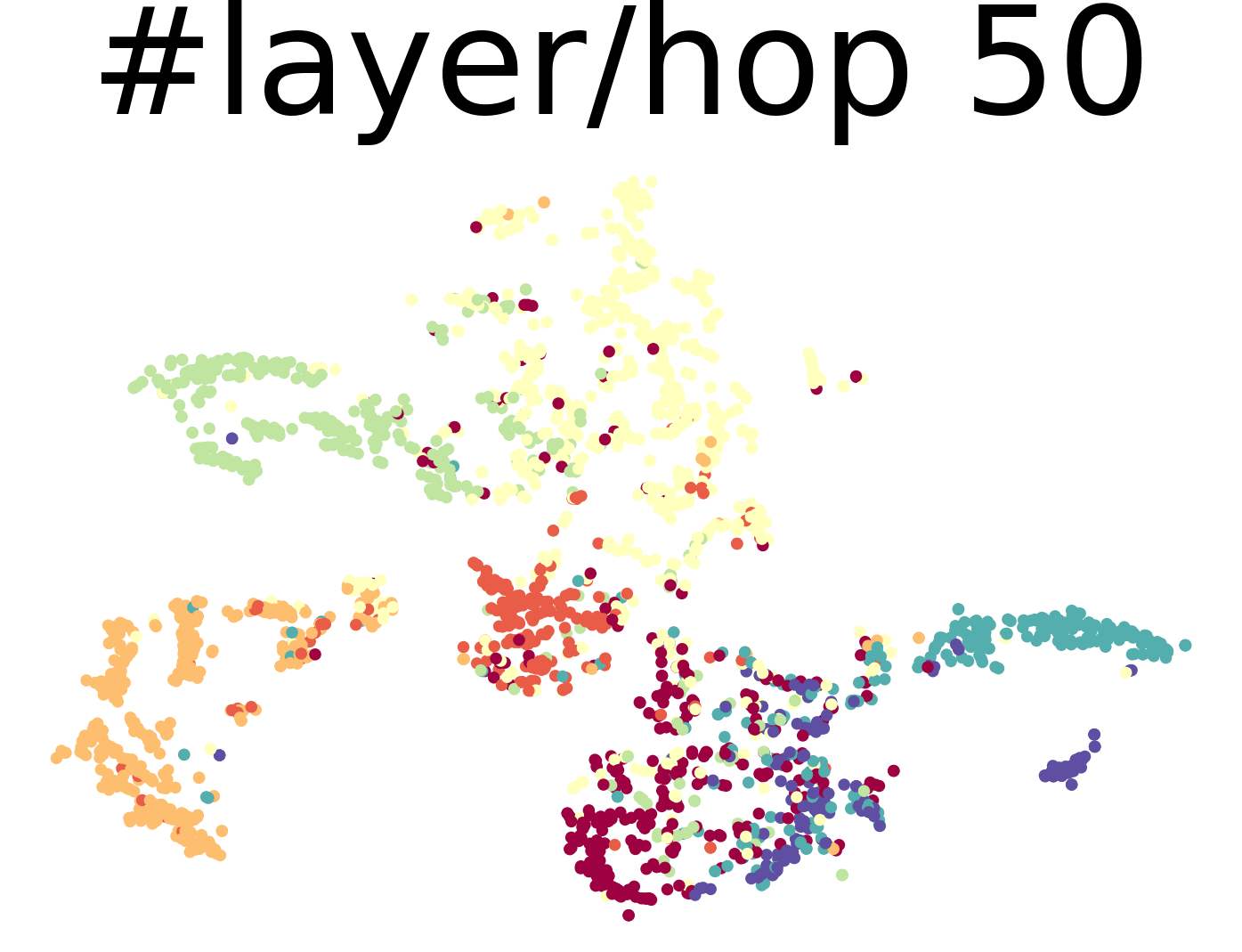}
}
\subfigure{
\includegraphics[width=0.205\columnwidth]{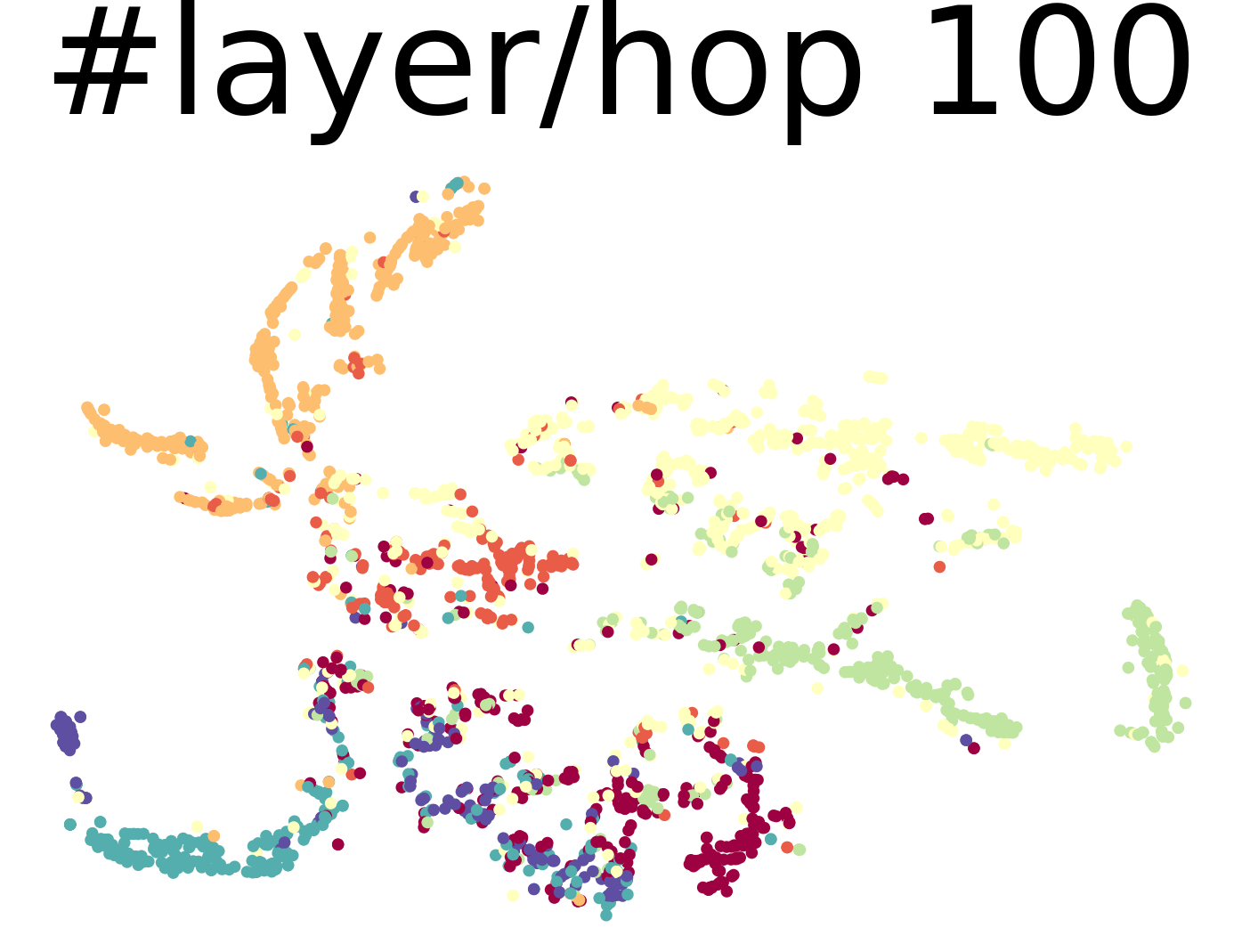}
}
\subfigure{
\includegraphics[width=0.205\columnwidth]{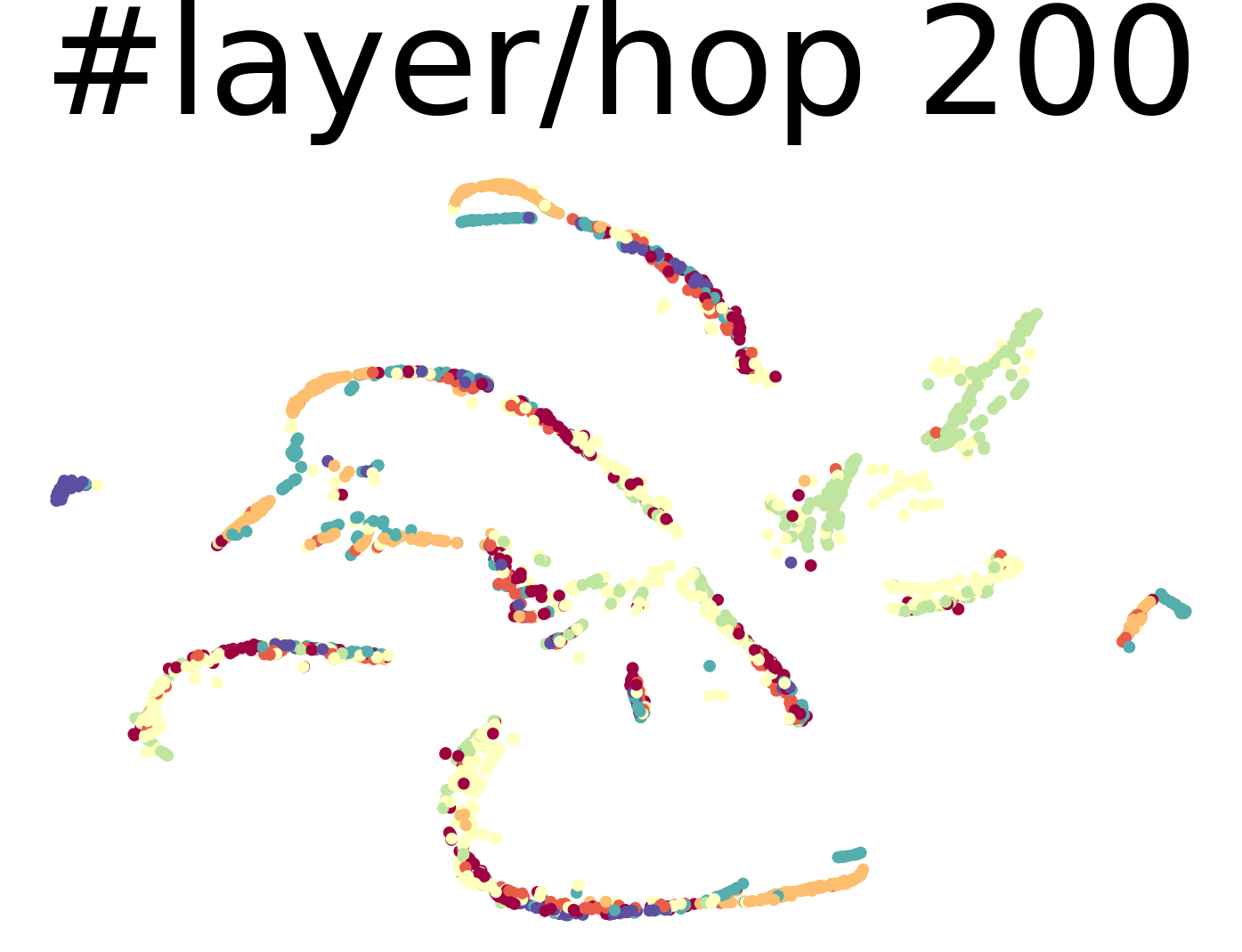}
}

\caption{t-SNE visualization of node representations derived by models as~Eq.(\ref{DetachGCN}) with different numbers of layers on Cora. Colors represent node classes.}
\label{fig:tsne2}
\end{figure*}

\begin{figure}
\centering
\includegraphics[width=0.75\columnwidth]{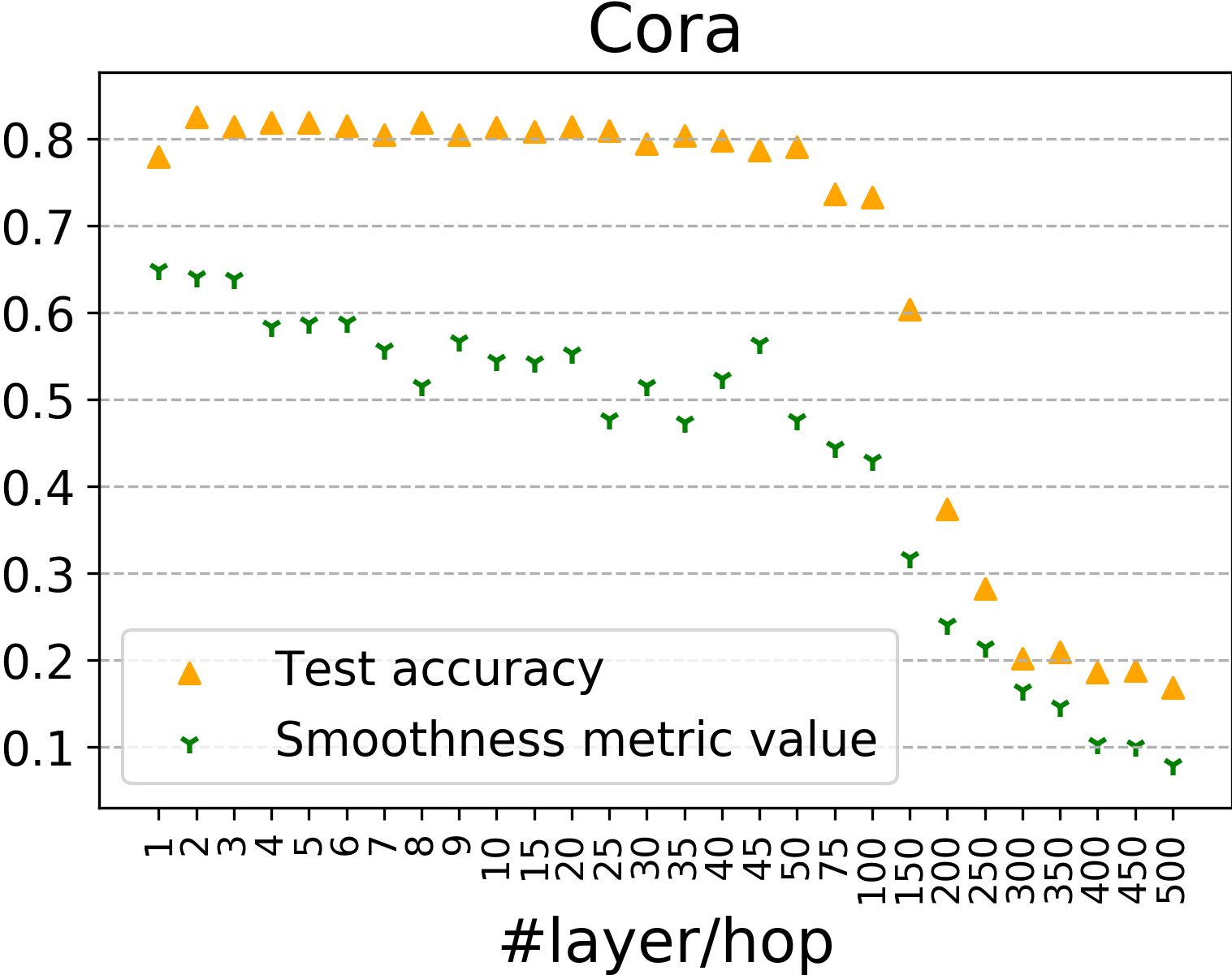}
\caption{Test accuracy and smoothness metric value of node representations with different numbers of layers adopted in models as~Eq.(\ref{DetachGCN}) on Cora.}
\label{fig:detachgcn}
\end{figure}

\subsection{Why Deeper GNNs Fail?}\label{Sec:why}
In this section, we utilize our proposed smoothness metric to
investigate the performance deterioration phenomenon in deep graph neural
networks. Here, we mainly use the GCN layer for analysis, but the
main results can be easily applied to other graph deep learning
methods. Besides using our proposed metric from the quantitative
perspective, we employ a data visualization technique
t-SNE~\cite{maaten2008visualizing}. t-SNE provides an
interpretable visualization, especially on high-dimensional data.
t-SNE is capable of capturing both the local structure and the
global structure like clusters in high-dimensional data, which is
consistent with the classification of nodes. Hence, we utilize
t-SNE to demonstrate the discriminative power of node
representations in the graph.

We develop a series of graph neural networks~(GNNs) with
different depths in terms of the number of GCN layers, and evaluate
them on three citation datasets; those are Cora, CiteSeer and
PubMed~\cite{sen2008collective}. In particular, we also include a
graph neural network with depth of 0, which is approximated with
a multi-layer perceptron network. A GNN with depth of 0 can be
viewed as aggregating information from a 0-hop neighborhood,
which only considers node features while with the graph structure
ignored. We conduct 100 runs for each model on each dataset, using the same data split scheme
as~\cite{kipf2016semi}.

The result on Cora is illustrated in
Figure~\ref{fig:gcn_degrade}. We provide results on other
datasets in Section~\ref{Sec:acc_gcn} in the appendix. We can
observe that test accuracy increases as the rise of the number of
layers in the beginning, but degrades dramatically from 3 layers.
Besides, from the t-SNE visualization on Cora in Figure
\ref{fig:tsne1} (t-SNE visualization results of other datasets
are provided in Appendix \ref{Sec:visual_citeseer_pubmed_gcn}.),
the discriminative power of the node representations derived by
different numbers of GCN layers has the similar trend. The node
representations generated by multiple GCN layers, like 6 layers,
are very difficult to be separated.

Several studies~\cite{li2018deeper,chen2019measuring} attribute this
performance degradation phenomenon to the over-smoothing issue.
However, we question this view for the following two reasons.
First, we hold that the over-smoothing issue only happens when
node representations propagate repeatedly for a large number of
iterations, especially for a graph with sparsely connected edges.
As shown in Table \ref{tab:dataset}, all these three citation
datasets have a small value of edge density, hence several
propagation iterations are not enough to make over-smoothing
happen, theoretically. Second, as shown in Figure
\ref{fig:gcn_degrade}, the smoothness metric value of graph node
representations has a slight downward trend as the number of
propagation iterations increases. According to
~\cite{li2018deeper}, the node representations suffering from the
over-smoothing issue will converge to the same value or be
proportional to the square root of the node degree, where the
corresponding smoothness metric value should be close to $0$,
computed by our quantitative metric. However, the metric value in
our experiments is relatively far from the ideal over-smoothing
situation.

In this work, we argue that it is the entanglement of transformation and propagation that significantly compromise the performance of
deep graph neural networks. Our argument is originated from the
following two intuitions. First, the entanglement of representation
transformation and propagation makes the number of parameters in
transformation intertwined with the receptive fields in
propagation. As illustrated in~Eq.(\ref{eq:gnn}), one hop
propagation requires a transformation function, thus leading to
a large number of parameters when considering a large receptive
field. Hence, it might be hard to train a deep GNN with a large number of parameters.
This can possibly explain why the performance of multiple GCN
layers in Figure \ref{fig:gcn_degrade} fluctuates greatly.
Second, representation propagation and transformation should be viewed
as two separate operations. Note that the class of a node can be
totally predictable by its initial features, which explains why
MLP, as shown in Figure \ref{fig:gcn_degrade} and
\ref{fig:tsne1}, performs well without using any graph structure
information. Propagation based on the graph structure can
help to ease the classification task by making node representations in
the same class to be similar, under the assumption that connected
nodes usually belong to the same class. For instance,
intuitively, the class of a document is completely determined by
its content (i.e. its feature derived by word embedding), instead
of the references relationships with other documents. Utilizing
its neighbors' features just eases the classification of
documents. Hence, representation transformation and propagation play
their distinct roles from feature and structure aspects,
respectively.

To support and verify our argument, we decouple the propagation and
transformation in~Eq.(\ref{GCN_EQ}), leading to the following
model:
\begin{equation}
\label{DetachGCN}
\begin{aligned}
\boldsymbol{Z} &= {\rm MLP}\left(\boldsymbol{X}\right)
\\
\boldsymbol{X}_{out} &= {\rm softmax} \left(\widehat{\boldsymbol{A}}^k\boldsymbol{Z}\right).
\end{aligned}
\end{equation}
$\boldsymbol{Z} \in
\mathbb{R}^{n \times c}$ denotes the new feature matrix
transformed from the original feature matrix by an MLP network, where $c$ is the number of classes. After the transformation, we apply a $k$-steps propagation to
derive the output feature matrix $\boldsymbol{X}_{out} \in
\mathbb{R}^{n \times c}$. A \rm{softmax} classifier is applied to
compute the classification probabilities. Notably, the separation
of transformation and propagation processes is also adopted in~\cite{klicpera2018predict} and ~\cite{wu2019simplifying} but for
the sake of reducing complexity. In this work, we analyze
this scheme systematically and reveal that it can help to build
deeper models without suffering from performance degradation,
which has not been prioritized by the community.

The test accuracy and smoothness metric value of representations
with different numbers of layers adopted in~Eq.(\ref{DetachGCN})
on Cora are illustrated in Figure~\ref{fig:detachgcn} (Results
for other datasets are shown in Appendix
\ref{Sec:detachgcn_app}). After resolving the entanglement of
feature transformation and propagation, deeper models is capable
of leveraging larger receptive fields without suffering from
performance degradation. We can observe that the over-smoothing
issue starts compromising the performance at an extremely large
receptive field, such as 75-hop on Cora. The smoothness metric
value decreases greatly after that, which is demonstrated by the
metric value of nearly 0. Besides, from the t-SNE visualization
in Figure \ref{fig:tsne2} (The t-SNE visualization results of
other datasets are provided in Appendix
\ref{Sec:visual_citeseer_pubmed_deachedgcn}), deep models with
large receptive fields, like 50-hop, still generating
distinguishable node representations, which is impressive
compared to the regular GCN model. In practice, we usually do not
need an extremely large receptive field because the highest
shortest path distance in a connected component usually is an
acceptable small number. Thus training signals can be propagated
to the entire graph with a small number of layers. This is
demonstrated by the fact that graph neural networks with 2 or 3
GCN layers usually perform competitively. However, deep models
with large receptive fields are necessary to incorporate more
information, especially with limited training nodes under a semi-supervised learning setting.

\subsection{Theoretical Analysis of Very Deep Models}\label{Sec:TH}

The empirical analysis in the previous section shows that
decoupling transformation and propagation can help to build much
deeper models which can leverage larger receptive fields to incorporate more information. In this
section, we provide a theoretical analysis of the above observation when building very deep graph neural networks, which aligns with the over-smoothing issue.
\cite{li2018deeper} and ~\cite{xu2018representation} study the
over-smoothing issue from the perspective of Laplacian smoothing
and nodes' influence distribution, with several simplified
assumptions like non-linear transformation and probability
approximations. After decoupling transformation from propagation, our theoretical analysis can serve as a more
rigorous and gentle description of the over-smoothing issue. In
this section, we strictly describe the over-smoothing issue for 2
typical propagation mechanisms.

$\widehat{\boldsymbol{A}}_\oplus=\widetilde{\boldsymbol{D}}^{-1}\widetilde{\boldsymbol{A}}$
and
$\widehat{\boldsymbol{A}}_\odot=\widetilde{\boldsymbol{D}}^{-\frac{1}{2}}\widetilde{\boldsymbol{A}}\widetilde{\boldsymbol{D}}^{-\frac{1}{2}}$,
where $\widetilde{\boldsymbol{A}}=\boldsymbol{A}+\boldsymbol{I}$,
are two frequently utilized propagation mechanisms. The
row-averaging normalization $\widehat{\boldsymbol{A}}_\oplus$ is
adopted in GraphSAGE~\cite{hamilton2017inductive} and
DGCNN~\cite{zhang2018end}. The symmetrical normalization scheme
$\widehat{\boldsymbol{A}}_\odot$ is applied in
GCN~\cite{kipf2016semi}. In the following, we describe the
over-smoothing issue by proving the convergence of
$\widehat{\boldsymbol{A}}_\oplus^k$ and
$\widehat{\boldsymbol{A}}_\odot^k$, respectively, when $k$ goes
to infinity.

Let $\boldsymbol{e}=[1,1,\cdots,1] \in \mathbb{R}^{1 \times n}$
be a row vector whose all entries are $1$. Function
$\Psi(\boldsymbol{x})=\frac{\boldsymbol{x}}{{\rm
sum}(\boldsymbol{x})}$ normalizes a vector to sum to $1$ and
function $\Phi(\boldsymbol{x})=\frac{\boldsymbol{x}}{\lVert
\boldsymbol{x} \rVert}$ normalizes a vector such that its
magnitude is $1$.

\begin{theorem}
\label{TH1}
Given a connected graph $G$, $\lim_{k\to \infty}
\widehat{\boldsymbol{A}}_\oplus^k=\boldsymbol{\Pi}_\oplus$,
where $\boldsymbol{\Pi}_\oplus$ is the matrix with all rows
are $\boldsymbol{\pi}_\oplus$ and
$\boldsymbol{\pi}_\oplus=\Psi(\boldsymbol{e}\widetilde{\boldsymbol{D}})$.
\end{theorem}

\begin{theorem}
\label{TH2}
Given a connected graph $G$, $\lim_{k\to \infty}
\widehat{\boldsymbol{A}}_\odot^k=\boldsymbol{\Pi}_\odot$,
where
$\boldsymbol{\Pi}_\odot=\Phi(\widetilde{\boldsymbol{D}}^{\frac{1}{2}}\boldsymbol{e}^T)(\Phi(\widetilde{\boldsymbol{D}}^{\frac{1}{2}}\boldsymbol{e}^T))^T$.
\end{theorem}

From the above two theorems, we can derive the exact convergence
value of $\widehat{\boldsymbol{A}}_\oplus^k$ and
$\widehat{\boldsymbol{A}}_\odot^k$, respectively, when $k$ goes
to infinity in an infinite deep model. Hence, applying infinite
layers to propagate information iteratively is equivalent to
utilizing $\boldsymbol{\Pi}_\oplus$ or $\boldsymbol{\Pi}_\odot$
to propagate features by one step. Rows of
$\boldsymbol{\Pi}_\oplus$ are the same and rows of
$\boldsymbol{\Pi}_\odot$ are proportional to the square root
value of the corresponding nodes' degrees. Therefore, rows of
$\boldsymbol{\Pi}_\oplus$ or $\boldsymbol{\Pi}_\odot$ are
linearly inseparable and utilizing them as propagation mechanism
will generate indistinguishable representations, thereby leading
to the over-smoothing issue.

To prove these two theorems, we first introduce the following two
lemmas. The proofs of these two lemmas can be found in Appendix~\ref{Sec:proof1} and ~\ref{Sec:proof2}.

\begin{lemma}
\label{lemma1}
Given a graph $G$, $\lambda$ is an eigenvalue of
$\widehat{\boldsymbol{A}}_\oplus$ with left eigenvector
$\boldsymbol{v}_l \in \mathbb{R}^{1 \times n}$ and right
eigenvector $\boldsymbol{v}_r \in \mathbb{R}^{n \times 1}$ if
and only if $\lambda$ is an eigenvalue of
$\widehat{\boldsymbol{A}}_\odot$ with left eigenvector
$\boldsymbol{v}_l\widetilde{\boldsymbol{D}}^{-\frac{1}{2}}
\in \mathbb{R}^{1 \times n}$ and right eigenvector
$\widetilde{\boldsymbol{D}}^{\frac{1}{2}}\boldsymbol{v}_r \in
\mathbb{R}^{n \times 1}$.
\end{lemma}

\begin{lemma}
\label{lemma2}
Given a connected graph $G$,
$\widehat{\boldsymbol{A}}_\oplus$ and
$\widehat{\boldsymbol{A}}_\odot$ always have an eigenvalue
$1$ with unique associated eigenvectors and all other
eigenvalues $\lambda$ satisfy $\lvert \lambda \rvert < 1$.
The left and right eigenvectors of
$\widehat{\boldsymbol{A}}_\oplus$  associated with eigenvalue
$1$ are $\boldsymbol{e}\widetilde{\boldsymbol{D}} \in
\mathbb{R}^{1 \times n}$ and $\boldsymbol{e}^T \in
\mathbb{R}^{n \times 1}$, respectively. For
$\widehat{\boldsymbol{A}}_\odot$, they are
$\boldsymbol{e}\widetilde{\boldsymbol{D}}^{\frac{1}{2}} \in
\mathbb{R}^{1 \times n}$ and
$\widetilde{\boldsymbol{D}}^{\frac{1}{2}}\boldsymbol{e}^T \in
\mathbb{R}^{n \times 1}$.
\end{lemma}

\begin{figure*}
\centering
\includegraphics[width=2.1\columnwidth]{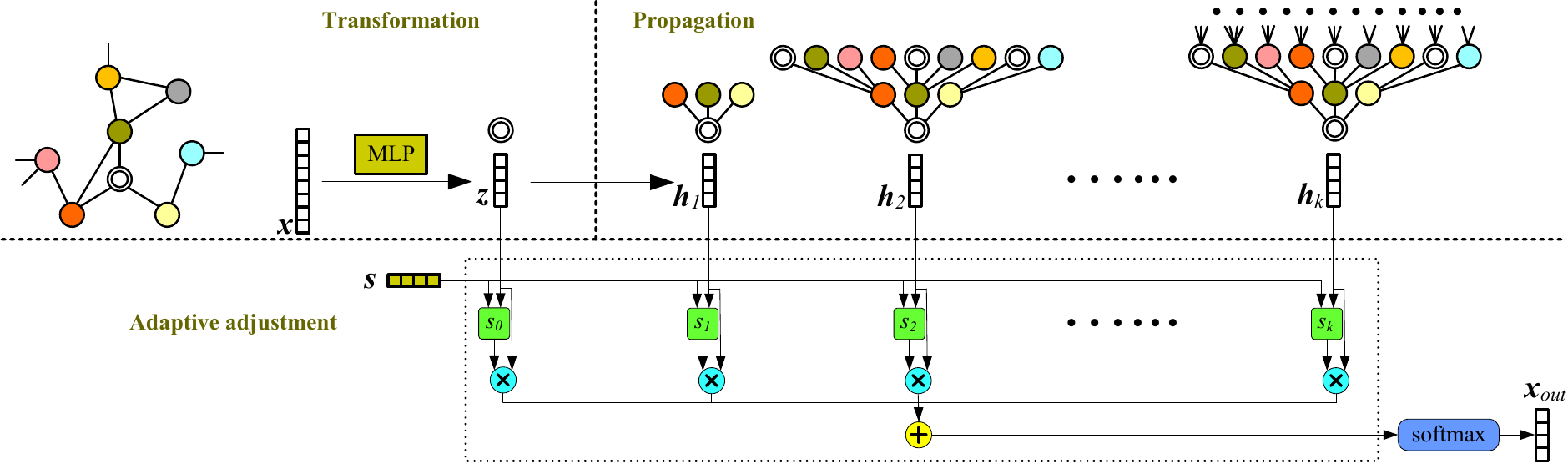}
\caption{An illustration of the proposed Deep Adaptive Graph Neural Network (DAGNN) . For clarity, we show the pipeline to generate the prediction for one node. Notation letters are consistent with~Eq.(\ref{EQ:DTCGN}) but bold lowercase versions are applied to denote representation vectors. $\boldsymbol{s}$ is the projection vector that computes retainment scores for representations generating from various receptive fields. $s_0$, $s_1$, $s_2$, and $s_k$ represent the retainment scores of $\boldsymbol{z}$, $\boldsymbol{h}_1$, $\boldsymbol{h}_2$, and $\boldsymbol{h}_k$, respectively.}
\label{fig:dtgcn}
\end{figure*}

\begin{proof}
(of Theorem \ref{TH1}) $\widehat{\boldsymbol{A}}_\oplus$ can
be viewed as a transition matrix because all entries are
nonnegative and each row sums to $1$. The graph $G$ can be
further regarded as a Markov chain, whose transition matrix
$\boldsymbol{P}$ is $\widehat{\boldsymbol{A}}_\oplus$. This
Markov chain is irreducible and aperiodic because the graph
$G$ is connected and self-loops are included in the
connectivity. If a Markov chain is irreducible and aperiodic,
then $\lim_{k\to \infty} \boldsymbol{P}^k=\boldsymbol{\Pi}$,
where $\boldsymbol{\Pi}$ is the matrix with all rows equal to
$\boldsymbol{\pi}$ and $\boldsymbol{\pi}$ can be computed by
$\boldsymbol{\pi}\boldsymbol{P}=\boldsymbol{\pi}$, s.t.
$\sum_i \boldsymbol{\pi}_i = 1$\cite{kumar2015stochastic}. It
is obvious that $\boldsymbol{\pi}$ is the unique left
eigenvector of $\boldsymbol{P}$ and is normalized such that
all entries sum to $1$.  Hence, $\lim_{k\to \infty}
\widehat{\boldsymbol{A}}_\oplus^k=\boldsymbol{\Pi}_\oplus$,
where $\boldsymbol{\Pi}_\oplus$ is the matrix with all rows
are $\boldsymbol{\pi}_\oplus$ and
$\boldsymbol{\pi}_\oplus=\Psi(\boldsymbol{e}\widetilde{\boldsymbol{D}})$
from Lemma \ref{lemma2}.
\end{proof}

\begin{proof}
(of Theorem \ref{TH2}) Although
$\widehat{\boldsymbol{A}}_\odot$ cannot be processed as a
transition matrix like $\widehat{\boldsymbol{A}}_\oplus$, it
is a symmetric matrix, which is diagonalizable. We have
$\widehat{\boldsymbol{A}}_\odot =
\boldsymbol{Q}\boldsymbol{\Lambda}\boldsymbol{Q}^T$,
where $\boldsymbol{Q}$ is an orthogonal matrix whose columns
are normalized eigenvectors of
$\widehat{\boldsymbol{A}}_\odot$ and $\boldsymbol{\Lambda}$
is the diagonal matrix whose diagonal entries are the
eigenvalues. Then the $k$-th power of
$\widehat{\boldsymbol{A}}_\odot$ can be computed by
\begin{equation}
\label{Eq:decom}
\widehat{\boldsymbol{A}}^k_\odot = \boldsymbol{Q}\boldsymbol{\Lambda}\boldsymbol{Q}^T\cdots\boldsymbol{Q}\boldsymbol{\Lambda}\boldsymbol{Q}^T=\boldsymbol{Q}\boldsymbol{\Lambda}^k\boldsymbol{Q}^T=\sum_{i=1}^{k} \lambda_i^n \boldsymbol{v}_i \boldsymbol{v}_i^T,
\end{equation}
where $\boldsymbol{v}_i$ is the normalized right eigenvector
associated with $\lambda_i$. From Lemma \ref{lemma2},
$\widehat{\boldsymbol{A}}_\odot$ always has an eigenvalue $1$
with unique associated eigenvectors and all other eigenvalues
$\lambda$ satisfy $\lvert \lambda \rvert < 1$. Hence,
$\lim_{k\to \infty}
\widehat{\boldsymbol{A}}_\odot^k=\Phi(\widetilde{\boldsymbol{D}}^{\frac{1}{2}}\boldsymbol{e}^T)(\Phi(\widetilde{\boldsymbol{D}}^{\frac{1}{2}}\boldsymbol{e}^T))^T$.
\end{proof}

These two theorems hold for connected graphs that are frequently
studied in graph neural networks. For a disconnected graph, these
theorems can also be applied to each of its connected components,
which means that applying these propagation mechanisms infinite times
will generate indistinguishable node representations in each
connected components.

The above theorems reveal that over-smoothing will make node
representations inseparable and provide the exact convergence value of
frequently used propagation mechanisms. Theoretically, we have
proved that the over-smoothing issue is inevitable in very deep models.
Further, the convergence speed is a more important factor that we
should consider in practice. Mathematically, according to~Eq.(\ref{Eq:decom}), the convergence speed depends on the other
eigenvalues except $1$ of the propagation matrix, especially the
second largest eigenvalue. Intuitively, the propagation matrix is
determined by the topology information of the corresponding graph.
This might be the reason for our observation in Section
\ref{Sec:why} that a sparsely connected graph suffers from the
over-smoothing only when extremely deep models are applied.

\section{Deep Adaptive Graph Neural Network}

In this section, we propose Deep Adaptive Graph Neural Network (DAGNN) based on the above insights. Our DAGNN contributes two
prominent advantages. First, it decouples the representation
transformation from propagation so that large receptive fields
can be applied without suffering from performance degradation,
which has been verified in Section \ref{Sec:why}. Second, it
utilizes an adaptive adjustment mechanism that can adaptively
balance the information from local and global neighborhoods for
each node, thus leading to more discriminative node
representations. The mathematical expression of DAGNN is defined
as
\begin{equation}
\begin{aligned}
&\boldsymbol{Z} = {\rm MLP}\left(\boldsymbol{X}\right) & & \in \mathbb{R}^{n \times c}
\\
&\boldsymbol{H}_\ell = \widehat{\boldsymbol{A}}^\ell\boldsymbol{Z}, \ell=1,2,\cdots,k & & \in \mathbb{R}^{n \times c}
\\
&\boldsymbol{H} = {\rm stack} \left(\boldsymbol{Z}, \boldsymbol{H}_1, \cdots, \boldsymbol{H}_k\right) & & \in \mathbb{R}^{n \times (k+1) \times c}
\\
&\boldsymbol{S} =\sigma\left(\boldsymbol{H}\boldsymbol{s}\right)  & & \in \mathbb{R}^{n \times (k+1) \times 1}
\\
&\widetilde{\boldsymbol{S}} = {\rm reshape}\left(\boldsymbol{S}\right) & & \in \mathbb{R}^{n \times 1 \times (k+1)}
\\
&\boldsymbol{X}_{out}= {\rm softmax}\left({\rm squeeze}\left(\widetilde{\boldsymbol{S}}\boldsymbol{H}\right)\right) & & \in \mathbb{R}^{n \times c},
\end{aligned}
\label{EQ:DTCGN}
\end{equation}
where $c$ is the number of node classes. $\boldsymbol{Z}\in
\mathbb{R}^{n \times c}$ is the feature matrix derived by
applying an MLP network to the original feature matrix. We utilize
the symmetrical normalization propagation mechanism
$\widehat{\boldsymbol{A}}=\widetilde{\boldsymbol{D}}^{-\frac{1}{2}}\widetilde{\boldsymbol{A}}\widetilde{\boldsymbol{D}}^{-\frac{1}{2}}$,
where $\widetilde{\boldsymbol{A}}=\boldsymbol{A}+\boldsymbol{I}$.
$k$ is a hyperparameter that indicates the depth of the model.
$\boldsymbol{s}\in \mathbb{R}^{c \times 1}$ is a trainable
projection vector. $\sigma(\cdot)$ is an activation function and
we apply \textit{sigmoid}. \textit{stack}, \textit{reshape} and
\textit{squeeze} are utilized to rearrange the data dimension so
that dimension can be matched during computation.

An illustration of our proposed DAGNN is provided in
Figure~\ref{fig:dtgcn}. There are three main steps in DAGNN:
transformation, propagation and adaptive adjustment. We first
utilize a shared MLP network for feature transformation.
Theoretically, MLP can approximate any measurable
function~\cite{hornik1989multilayer}. Obviously, $\boldsymbol{Z}$
only contains the information of individual nodes themselves with
no structure information included. After transformation, a
propagation mechanism $\widehat{\boldsymbol{A}}$ is applied to
gather neighborhood information. $\boldsymbol{H}_\ell$ denotes the
representations obtained by propagating information from nodes
that are $\ell$-hop away, hence $\boldsymbol{H}_\ell$ captures the
information from the subtree of height $\ell$ rooted at individual
nodes. As the depth $\ell$ increase, more global information is
included in $\boldsymbol{H}_\ell$ because the corresponding subtree
is deeper. However, it is difficult to determine a suitable $\ell$.
Small $\ell$ may fail to capture sufficient and essential
neighborhood information, while large $\ell$ may bring too much
global information and dilute the special local information.
Furthermore, each node has a different subtree structure rooted
at this node and the most suitable receptive field for each node
should be different. To this end, we include an adaptive
adjustment mechanism after the propagation. We utilize a
trainable projection vector $\boldsymbol{s}$  that is shared by
all nodes to generate retainment scores. These scores are used
for representations that carry information from various range of
neighborhood. These retainment scores measure how much
information of the corresponding representations derived by
different propagation layers should be retained to generate the
final representation for each node. Utilizing this adaptive adjustment mechanism, DAGNN can adaptively balance the information from local and global neighborhoods for each node. Obviously, the transformation
process and the adaptive adjustment process have trainable
parameters and there is no trainable parameter in the propagation
process, leading to a parameter-efficient model. Note that DAGNN
is trained end-to-end, which means that these three steps are
considered together when optimizing the network.

\begin{table*}[t]
\caption{Statistics of datasets. The edge density is computed by $\frac{2m}{n^2}$. Note that for a fair comparison with other baselines, we only consider the largest connected component in co-purchase graphs as ~\cite{shchur2018pitfalls}.}
\label{tab:dataset}
\resizebox{2.1\columnwidth}{!}{
\begin{tabular}{lcccccccc}
\toprule
\textbf{Dataset} &\textbf{\#Classes} &\textbf{\#Nodes} &\textbf{\#Edges} &\textbf{Edge Density} &\textbf{\#Features} &\textbf{\#Training Nodes} &\textbf{\#Validation Nodes} &\textbf{\#Test Nodes} \\
\midrule
Cora &7 &2708 &5278 &0.0014 &1433 &20 per class &500 &1000 \\
CiteSeer &6 &3327 &4552 &0.0008&3703 &20 per class &500 &1000 \\
PubMed &3 &19717 &44324 &0.0002 &500 &20 per class &500 &1000 \\
Coauthor CS &15 &18333 &81894 &0.0005&6805 &20 per class &30 per class &Rest nodes \\
Coauthor Physics &5 &34493 &247962 &0.0004&8415 &20 per class &30 per class &Rest nodes\\
Amazon Computers &10 &13381 &245778&0.0027 &767 &20 per class &30 per class &Rest nodes \\
Amazon Photo &8 &7487 &119043 &0.0042&745 &20 per class &30 per class &Rest nodes \\
\bottomrule
\end{tabular}
}
\end{table*}
\begin{table}[t]
\caption{Results on citation datasets with both fixed and random splits in terms of classification accuracy (in percent).}
\label{tab:result_citation}
\resizebox{\columnwidth}{!}{
\begin{tabular}{lcccccc}
\toprule
\multirow{2}{*}{\textbf{Models}} &\multicolumn{2}{c}{\textbf{Cora}} &\multicolumn{2}{c}{\textbf{CiteSeer}} &\multicolumn{2}{c}{\textbf{PubMed}}  \\
&Fixed &Random &Fixed &Random &Fixed &Random \\
\midrule
MLP &$61.6\pm0.6$ &$59.8\pm2.4$ &$61.0\pm1.0$ &$58.8\pm2.2$ &$74.2\pm0.7$ &$70.1\pm2.4$ \\
ChebNet &$80.5\pm1.1$ &$76.8\pm2.5$ &$69.6\pm1.4$ &$67.5\pm2.0$ &$78.1\pm0.6$ &$75.3\pm2.5$ \\
GCN &$81.3\pm0.8$ &$79.1\pm1.8$ &$71.1\pm0.7$ &$68.2\pm1.6$ &$78.8\pm0.6$ &$77.1\pm2.7$ \\
GAT &$83.1\pm0.4$ &$80.8\pm1.6$ &$70.8\pm0.5$ &$68.9\pm1.7$ &$79.1\pm0.4$ &$77.8\pm2.1$ \\
APPNP &$83.3\pm0.5$ &$81.9\pm1.4$ &$71.8\pm0.4$ &$69.8\pm1.7$ &$80.1\pm0.2$ &$79.5\pm2.2$ \\
SGC &$81.7\pm0.1$ &$80.4\pm1.8$ &$71.3\pm0.2$ &$68.7\pm2.1$ &$78.9\pm0.1$ &$76.8\pm2.6$ \\
\textbf{DAGNN (Ours)}  &$\textbf{84.4}\pm\textbf{0.5}$ &$\textbf{83.7}\pm\textbf{1.4}$ &$\textbf{73.3}\pm\textbf{0.6}$ &$\textbf{71.2}\pm\textbf{1.4}$ &$\textbf{80.5}\pm\textbf{0.5}$ &$\textbf{80.1}\pm\textbf{1.7}$ \\
\bottomrule
\end{tabular}
}
\end{table}

\newcommand{\tabincell}[2]{\begin{tabular}{@{}#1@{}}#2\end{tabular}}

\begin{table}[t]
\caption{Results on co-authorship and co-purchase datasets in terms of classification accuracy (in percent).}
\label{tab:result_co}
\resizebox{\columnwidth}{!}{
\begin{tabular}{lcccccc}
\toprule
\textbf{Models} &\textbf{\tabincell{c}{Coauthor \\ CS}}&\textbf{\tabincell{c}{Coauthor\\ Physics}}&\textbf{\tabincell{c}{Amazon \\Computers}}&\textbf{\tabincell{c}{Amazon\\ Photo}}   \\
\midrule
LogReg &$86.4\pm0.9$ &$86.7\pm1.5$&$64.1\pm5.7$ &$73.0\pm6.5$ \\
MLP &$88.3\pm0.7$ &$88.9\pm1.1$ &$44.9\pm5.8$ &$69.6\pm3.8$ \\
LabelProp &$73.6\pm3.9$ &$86.6\pm2.0$&$70.8\pm8.1$ &$72.6\pm11.1$ \\
LabelProp NL &$76.7\pm1.4$ &$86.8\pm1.4$&$75.0\pm2.9$ &$83.9\pm2.7$ \\
GCN &$91.1\pm0.5$ &$92.8\pm1.0$&$82.6\pm2.4$ &$91.2\pm1.2$ \\
GAT &$90.5\pm0.6$ &$92.5\pm0.9$&$78.0\pm19.0$ &$85.7\pm20.3$ \\
MoNet &$90.8\pm0.6$ &$92.5\pm0.9$&$83.5\pm2.2$ &$91.2\pm1.3$ \\
GraphSAGE-mean &$91.3\pm2.8$ &$93.0\pm0.8$&$82.4\pm1.8$ &$91.4\pm1.3$  \\
GraphSAGE-maxpool &$85.0\pm1.1$ &$90.3\pm1.2$&N/A &$90.4\pm1.3$ \\
GraphSAGE-meanpool &$89.6\pm0.9$ &$92.6\pm1.0$&$79.9\pm2.3$ &$90.7\pm1.6$ \\
\textbf{DAGNN (Ours)}  &$\textbf{92.8}\pm\textbf{0.9}$ &$\textbf{94.0}\pm\textbf{0.6}$&$\textbf{84.5}\pm\textbf{1.2}$ &$\textbf{92.0}\pm\textbf{0.8}$  \\
\bottomrule
\end{tabular}
}
\end{table}

Decoupling representation transformation from propagation and utilizing
the learnable retainment score to adaptively adjust the information from local and
global neighborhoods make DAGNN have the ability to generate
suitable representations for specific nodes from large and
adaptive receptive fields. Besides, removing the entanglement of
representation transformation and propagation, we can derive a large
neighborhood without introducing more trainable parameters. Also,
transforming representations to a low dimensional space at an early stage
makes DAGNN computationally efficient and memory-saving. In order
to compute $\widehat{\boldsymbol{A}}^\ell\boldsymbol{Z}$
efficiently, we choose to compute it sequentially from right to
left with time complexity $\mathcal{O}(n^2c)$, which saves the
computational cost compared to $\mathcal{O}(n^3)$ of calculating
$\widehat{\boldsymbol{A}}^\ell$ first.

Notably, there are no fully-connected layers utilized as a
classifier at the end of this model. In our DAGNN, the final
representations $\boldsymbol{X}_{out}$ are used as the final
prediction. Thus, the cross-entropy loss for all labeled examples
can be calculated as
\begin{equation}
\mathcal{L} = -\sum_{i \in V_L} \sum_{p=1}^{c} \boldsymbol{Y}_{[i,p]} \ln {\boldsymbol{X}_{out}}_{[i,p]},
\end{equation}
where $V_L$ is the set of labeled nodes and $\boldsymbol{Y} \in
\mathbb{R}^{n \times c}$ is the label indicator matrix. $c$ is
the number of classes.

\section{Experimental Studies}

In this section, we conduct extensive experiments on node
classification tasks to evaluate the superiority of our proposed
DAGNN. We begin by introducing datasets and experimental setup we
utilized. We then compare DAGNN with prior state-of-the-art
baselines to demonstrate the effectiveness of DAGNN. Also, we
deploy some performance studies to further verify the proposed
model.

\subsection{Datasets and Setup}

We conduct experiments on $7$ datasets based on citation,
co-authorship, or co-purchase graphs for semi-supervised node
classification tasks; those are Cora~\cite{sen2008collective},
CiteSeer~\cite{sen2008collective},
PubMed~\cite{sen2008collective}, Coauthor
CS~\cite{shchur2018pitfalls}, Coauthor
Physics~\cite{shchur2018pitfalls}, Amazon
Computers~\cite{shchur2018pitfalls}, and Amazon
Photo~\cite{shchur2018pitfalls}. The statistics of these datasets are summarized in Table~\ref{tab:dataset}. The detailed description of these datasets are provided in Appendix \ref{Sec:dataset}.

We implemented our proposed DAGNN and some necessary baselines
using Pytorch~\cite{paszke2017automatic} and Pytorch
Geometric~\cite{Fey/Lenssen/2019}, a library for deep learning on
irregularly structured data built upon Pytorch. We consider
the following baselines: Logistic Regression (LogReg), Multilayer
Perceptron (MLP), Label Propagation
(LabelProp)~\cite{chapelle2009semi}, Normalized Laplacian Label
Propagation (LabelProp NL)~\cite{chapelle2009semi},
ChebNet~\cite{defferrard2016convolutional}, Graph Convolutional
Network (GCN)~\cite{kipf2016semi}, Graph Attention
Network(GAT)~\cite{velivckovic2017graph}, Mixture Model Network
(MoNet)~\cite{monti2017geometric},
GraphSAGE~\cite{hamilton2017inductive},
APPNP~\cite{klicpera2018predict}, and
SGC~\cite{wu2019simplifying}. We aim to provide a rigorous
and fair comparison between different models on each dataset by
using the same dataset splits and training procedure. We tune hyperparameters for all models individually and some baselines even achieve better results than their original reports. For our DAGNN, we tune the following hyperparameters: (1) k $\in$ \{5, 10, 20\}, (2) weight decay $\in$ \{0, 2e-2, 5e-3, 5e-4, 5e-5\}, and (3) dropout rate $\in$ \{0.5, 0.8\}. Our code is publicly available \footnote{https://github.com/divelab/DeeperGNN}.

\begin{table*}[t]
\caption{Results with different training set sizes on Cora in terms of classification accuracy (in percent). Results in brackets are the improvements of DAGNN over GCN.}
\label{tab:train_size}
\resizebox{2.1\columnwidth}{!}{
\begin{tabular}{lccccccccccc}
\toprule
\textbf{\#Training nodes per class} &\textbf{1}&\textbf{2}&\textbf{3}&\textbf{4}&\textbf{5}&\textbf{10}&\textbf{20}&\textbf{30}&\textbf{40}&\textbf{50}&\textbf{100}\\
\midrule
MLP &30.3&35.0&38.3&40.8&44.7&53.0&59.8&63.0&64.8&65.4&64.0
\\
GCN &34.7&48.9&56.8&62.5&65.3&74.3&79.1&80.8&82.2&82.9&84.7
\\
GAT &45.3&58.8&66.6&68.4&70.7&77.0&80.8&82.6&83.4&84.0&86.1
\\
APPNP &44.7&58.7&66.3&71.2&74.1&79.0&81.9&83.2&83.8&84.3&85.4
\\
SGC &43.7&59.2&67.2&70.4&71.5&77.5&80.4&81.3&81.9&82.1&83.6
\\
\textbf{DAGNN (Ours)} &\textbf{58.4\scriptsize(23.7$\uparrow$)}&\textbf{67.7\scriptsize(18.8$\uparrow$)}&\textbf{72.4\scriptsize(15.6$\uparrow$)}&\textbf{75.5\scriptsize(13.0$\uparrow$)}&\textbf{76.7\scriptsize(11.4$\uparrow$)}&\textbf{80.8\scriptsize(6.5$\uparrow$)}&\textbf{83.7\scriptsize(4.6$\uparrow$)}&\textbf{84.5\scriptsize(3.7$\uparrow$)}&\textbf{85.6\scriptsize(3.4$\uparrow$)}&\textbf{86.0\scriptsize(3.1$\uparrow$)}&\textbf{87.1\scriptsize(2.4$\uparrow$)}
\\
\bottomrule
\end{tabular}
}
\end{table*}

\subsection{Overall Results}

The results on citation datasets are summarized in Table ~\ref{tab:result_citation}. To ensure a fair comparison, we use 20 labeled nodes per class as the training set, 500 nodes as the validation set, and 1000 nodes as the test set for all models. For each model, we conduct $100$ runs for the fixed training/validation/test split from ~\cite{kipf2016semi}, which is commonly used to evaluate performance by the community. Also, we conduct $100$ runs for each model on randomly training/validation/test splits, where we additionally ensure uniform class distribution on the train split as ~\cite{Fey/Lenssen/2019}. We compute the average test accuracy of $100$ runs. As shown in Table ~\ref{tab:result_citation}, our DAGNN model performs better than the representative baselines by significant margins. Also, the fact that DAGNN achieves state-of-the-art performance on random splits demonstrates the strong robustness of DAGNN. Quantitatively, for the randomly split data, the improvements of DAGNN over GCN are 4.6\%, 3.0\%, and 3.0\% on Cora, CiteSeer, and PubMed, respectively.

The results on co-authorship and co-purchase datasets are
summarized in Table ~\ref{tab:result_co}. We utilize 20 labeled
nodes per class as the training set, 30 nodes per class as the
validation set, and the rest as the test set. The results of
baselines are obtained from ~\cite{shchur2018pitfalls}. For DAGNN,
we conduct $100$ runs for randomly training/validation/test
splits as ~\cite{shchur2018pitfalls} to ensure a fair comparison
with baselines. Our DAGNN model achieves better performance over
the current state-of-the-art models by significant margins of
1.5\%, 1.0\%, 1.0\%, and 0.6\% on the Coauthor CS, Coauthor
Physics, Amazon Computers, and Amazon Photo, respectively. Note
that DAGNN reduces the error rate by $11\%$ on average.

In summary, our DAGNN achieves superior performance on all these
seven datasets, which significantly demonstrates the
effectiveness of our proposed model. These results verify the
superiority of learning node representations from large and
adaptive receptive fields, which is achieved by decoupling
transformation from propagation and utilizing an adaptive
adjustment mechanism in DAGNN.

\subsection{Training Set Sizes}

The number of training samples is usually limited in the real
world graph. Hence, it is necessary to explore how models perform
with different training set sizes. To further demonstrate the
advantage that our DAGNN is capable of capturing information from
large and adaptive receptive fields, we conduct experiments with
different training set sizes for several representative
baselines. MLP only utilizes node features to learn
representations. GCN and GAT include the structure information by
propagation representations through edges, however, only limited
receptive fields can be taken into consideration and it is shown
in Section \ref{Sec:why} that performance degrades when stacking
multiple GCN layers to enable large receptive fields. APPNP, SGC,
and our DAGNN all have the ability to deploy a large receptive
field. Note that for a fair comparison, we set the depth in APPNP, SGC, and DAGNN as 10, where information in the 10-hop
neighborhood can be included. For each model, we conduct $100$
runs on randomly training/validation/test splits for every
training set size on Cora dataset. The results are present in
Table \ref{tab:train_size} where our improvements over GCN are
also highlighted. Our DAGNN achieves the best performance under
different training set sizes. Notably, The superiority of our
DAGNN can be demonstrated more obviously when fewer training nodes
are used. The improvement of DAGNN over GCN increases greatly when
the number of training nodes decreases. Extremely, only utilizing
one training node per class, DAGNN achieves an overwhelming result
over GCN by a significant margin of 23.7\%. These considerable
improvements are mainly attributed to the advantage that DAGNN can
incorporate information from large receptive fields by removing
the entanglement of representation transformation and
propagation. Reachable large receptive fields are beneficial for
propagating training signals to distant nodes, which is very
essential when the number of training nodes is limited. APPNP and SGC also can gather information from a large neighborhood, but
they perform not as good as DAGNN. This performance gap is mainly
caused by the adaptive adjustment of DAGNN, which can adjust the
information from different receptive fields for each node
adaptively.

\begin{figure}
\centering
\includegraphics[width=\columnwidth]{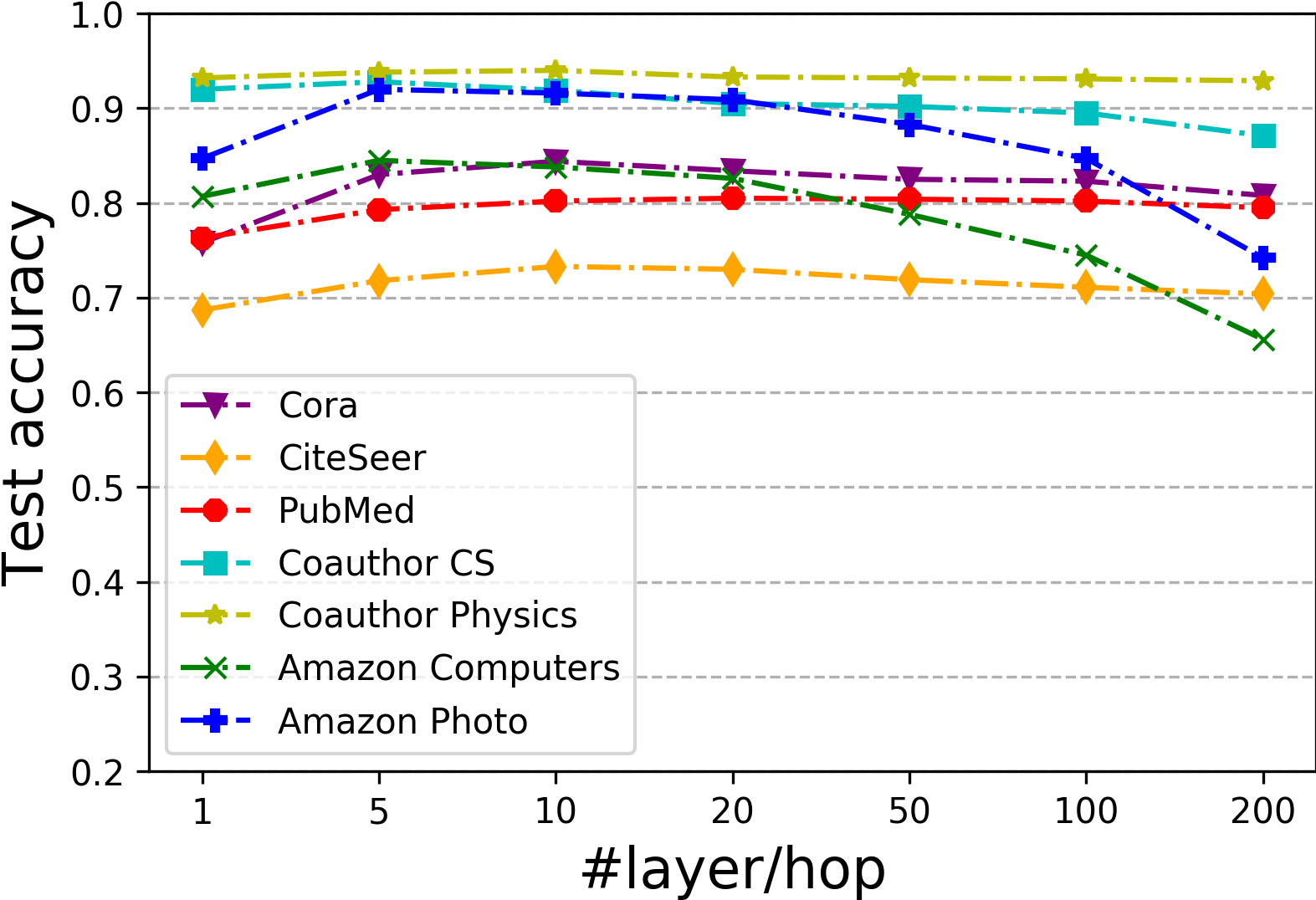}
\caption{Results of DAGNN with different depths.}
\label{krise}
\end{figure}

\subsection{Model Depths}
In order to investigate when over-smoothing happens in our DAGNN,
we conduct experiments for DAGNN with different depths. For each
dataset, we choose different hyperparameter $k$ in DAGNN, which
means the $k$-hop neighborhood is visible by each node, and
conduct $100$ runs for each setting. The results are illustrated
in Figure \ref{krise}. For citation and co-authorship datasets,
very deep models with large numbers of propagation iterations can
be applied with keeping stable or slightly decreasing
performance, which can be attributed to the design that we
decouple the transformation from propagation and utilize adaptive
receptive fields to learn node representations in DAGNN. Note that
performances on co-purchase datasets decrease obviously with the
increment of depth. This should be resulted by their larger value
of edge density than other datasets, as shown in Table
\ref{tab:dataset}. Intuitively, when nodes are more densely
connected, their representations will become indistinguishable by
applying a smaller number of propagation iterations, which aligns
with our assumption in Section \ref{Sec:TH} that further
connection exists between the graph topology information and convergence
speed.

\section{Conclusion}

In this paper, we consider the performance deterioration problem existed in current deep
graph neural networks and develop new insights towards deeper graph
neural networks. We first provide a systematical analysis on this issue
and argue that the key factor that compromises the network
performance is the entanglement of representation transformation and
propagation. We propose to decouple these two operations and show
that deep graph neural networks without this entanglement can
leverage large receptive fields without suffering from
performance deterioration. Further, we provide a theoretically analysis of the above strategy when building very deep models, which can
serve as a rigorous and gentle description of the over-smoothing
issue. Utilizing our insights, DAGNN is proposed to conduct node
representation learning with the ability to capture information
from large and adaptive receptive fields. According to our
comprehensive experiments, our DAGNN achieves a better performance
than current state-of-the-art models by significant margins,
especially when training samples are limited, which demonstrates
its superiority.

\begin{acks}
    This work was supported in part by National Science Foundation grants DBI-1922969, IIS-1908166, and IIS-1908198.
\end{acks}

\bibliographystyle{ACM-Reference-Format}
\balance
\bibliography{oversmooth}


\begin{thebibliography}{38}


\ifx \showCODEN    \undefined \def \showCODEN     #1{\unskip}     \fi
\ifx \showDOI      \undefined \def \showDOI       #1{#1}\fi
\ifx \showISBNx    \undefined \def \showISBNx     #1{\unskip}     \fi
\ifx \showISBNxiii \undefined \def \showISBNxiii  #1{\unskip}     \fi
\ifx \showISSN     \undefined \def \showISSN      #1{\unskip}     \fi
\ifx \showLCCN     \undefined \def \showLCCN      #1{\unskip}     \fi
\ifx \shownote     \undefined \def \shownote      #1{#1}          \fi
\ifx \showarticletitle \undefined \def \showarticletitle #1{#1}   \fi
\ifx \showURL      \undefined \def \showURL       {\relax}        \fi
\providecommand\bibfield[2]{#2}
\providecommand\bibinfo[2]{#2}
\providecommand\natexlab[1]{#1}
\providecommand\showeprint[2][]{arXiv:#2}

\bibitem[\protect\citeauthoryear{Cai and Ji}{Cai and Ji}{2020}]%
        {cai2020multi}
\bibfield{author}{\bibinfo{person}{Lei Cai} {and} \bibinfo{person}{Shuiwang
  Ji}.} \bibinfo{year}{2020}\natexlab{}.
\newblock \showarticletitle{A Multi-Scale Approach for Graph Link Prediction}.
  In \bibinfo{booktitle}{\emph{Thirty-Four AAAI Conference on Artificial
  Intelligence}}.
\newblock


\bibitem[\protect\citeauthoryear{Chapelle, Scholkopf, and Zien}{Chapelle
  et~al\mbox{.}}{2009}]%
        {chapelle2009semi}
\bibfield{author}{\bibinfo{person}{Olivier Chapelle}, \bibinfo{person}{Bernhard
  Scholkopf}, {and} \bibinfo{person}{Alexander Zien}.}
  \bibinfo{year}{2009}\natexlab{}.
\newblock \showarticletitle{Semi-supervised learning (chapelle, o. et al.,
  eds.; 2006)[book reviews]}.
\newblock \bibinfo{journal}{\emph{IEEE Transactions on Neural Networks}}
  \bibinfo{volume}{20}, \bibinfo{number}{3} (\bibinfo{year}{2009}),
  \bibinfo{pages}{542--542}.
\newblock


\bibitem[\protect\citeauthoryear{Chen, Lin, Li, Li, Zhou, and Sun}{Chen
  et~al\mbox{.}}{2020}]%
        {chen2019measuring}
\bibfield{author}{\bibinfo{person}{Deli Chen}, \bibinfo{person}{Yankai Lin},
  \bibinfo{person}{Wei Li}, \bibinfo{person}{Peng Li}, \bibinfo{person}{Jie
  Zhou}, {and} \bibinfo{person}{Xu Sun}.} \bibinfo{year}{2020}\natexlab{}.
\newblock \showarticletitle{Measuring and Relieving the Over-smoothing Problem
  for Graph Neural Networks from the Topological View}. In
  \bibinfo{booktitle}{\emph{Thirty-Four AAAI Conference on Artificial
  Intelligence}}.
\newblock


\bibitem[\protect\citeauthoryear{Defferrard, Bresson, and
  Vandergheynst}{Defferrard et~al\mbox{.}}{2016}]%
        {defferrard2016convolutional}
\bibfield{author}{\bibinfo{person}{Micha{\"e}l Defferrard},
  \bibinfo{person}{Xavier Bresson}, {and} \bibinfo{person}{Pierre
  Vandergheynst}.} \bibinfo{year}{2016}\natexlab{}.
\newblock \showarticletitle{Convolutional neural networks on graphs with fast
  localized spectral filtering}. In \bibinfo{booktitle}{\emph{Advances in
  neural information processing systems}}. \bibinfo{pages}{3844--3852}.
\newblock


\bibitem[\protect\citeauthoryear{Fey and Lenssen}{Fey and Lenssen}{2019}]%
        {Fey/Lenssen/2019}
\bibfield{author}{\bibinfo{person}{Matthias Fey} {and} \bibinfo{person}{Jan~E.
  Lenssen}.} \bibinfo{year}{2019}\natexlab{}.
\newblock \showarticletitle{Fast Graph Representation Learning with {PyTorch
  Geometric}}. In \bibinfo{booktitle}{\emph{ICLR Workshop on Representation
  Learning on Graphs and Manifolds}}.
\newblock


\bibitem[\protect\citeauthoryear{Gao and Ji}{Gao and Ji}{2019}]%
        {gao2019graph}
\bibfield{author}{\bibinfo{person}{Hongyang Gao} {and}
  \bibinfo{person}{Shuiwang Ji}.} \bibinfo{year}{2019}\natexlab{}.
\newblock \showarticletitle{Graph U-Nets}. In
  \bibinfo{booktitle}{\emph{International Conference on Machine Learning}}.
  \bibinfo{pages}{2083--2092}.
\newblock


\bibitem[\protect\citeauthoryear{Gao, Wang, and Ji}{Gao et~al\mbox{.}}{2018}]%
        {gao2018large}
\bibfield{author}{\bibinfo{person}{Hongyang Gao}, \bibinfo{person}{Zhengyang
  Wang}, {and} \bibinfo{person}{Shuiwang Ji}.} \bibinfo{year}{2018}\natexlab{}.
\newblock \showarticletitle{Large-scale learnable graph convolutional
  networks}. In \bibinfo{booktitle}{\emph{Proceedings of the 24th ACM SIGKDD
  International Conference on Knowledge Discovery \& Data Mining}}.
  \bibinfo{pages}{1416--1424}.
\newblock


\bibitem[\protect\citeauthoryear{Gilmer, Schoenholz, Riley, Vinyals, and
  Dahl}{Gilmer et~al\mbox{.}}{2017}]%
        {gilmer2017neural}
\bibfield{author}{\bibinfo{person}{Justin Gilmer}, \bibinfo{person}{Samuel~S
  Schoenholz}, \bibinfo{person}{Patrick~F Riley}, \bibinfo{person}{Oriol
  Vinyals}, {and} \bibinfo{person}{George~E Dahl}.}
  \bibinfo{year}{2017}\natexlab{}.
\newblock \showarticletitle{Neural message passing for quantum chemistry}. In
  \bibinfo{booktitle}{\emph{International Conference on Machine Learning}}.
  \bibinfo{pages}{1263--1272}.
\newblock


\bibitem[\protect\citeauthoryear{Hamilton, Ying, and Leskovec}{Hamilton
  et~al\mbox{.}}{2017}]%
        {hamilton2017inductive}
\bibfield{author}{\bibinfo{person}{Will Hamilton}, \bibinfo{person}{Zhitao
  Ying}, {and} \bibinfo{person}{Jure Leskovec}.}
  \bibinfo{year}{2017}\natexlab{}.
\newblock \showarticletitle{Inductive representation learning on large graphs}.
  In \bibinfo{booktitle}{\emph{Advances in Neural Information Processing
  Systems}}. \bibinfo{pages}{1024--1034}.
\newblock


\bibitem[\protect\citeauthoryear{Hornik, Stinchcombe, White,
  et~al\mbox{.}}{Hornik et~al\mbox{.}}{1989}]%
        {hornik1989multilayer}
\bibfield{author}{\bibinfo{person}{Kurt Hornik}, \bibinfo{person}{Maxwell
  Stinchcombe}, \bibinfo{person}{Halbert White}, {et~al\mbox{.}}}
  \bibinfo{year}{1989}\natexlab{}.
\newblock \showarticletitle{Multilayer feedforward networks are universal
  approximators.}
\newblock \bibinfo{journal}{\emph{IEEE Transactions on Neural Networks}}
  \bibinfo{volume}{2}, \bibinfo{number}{5} (\bibinfo{year}{1989}),
  \bibinfo{pages}{359--366}.
\newblock


\bibitem[\protect\citeauthoryear{Kipf and Welling}{Kipf and Welling}{2017}]%
        {kipf2016semi}
\bibfield{author}{\bibinfo{person}{Thomas~N Kipf} {and} \bibinfo{person}{Max
  Welling}.} \bibinfo{year}{2017}\natexlab{}.
\newblock \showarticletitle{Semi-supervised classification with graph
  convolutional networks}. In \bibinfo{booktitle}{\emph{International
  Conference on Learning Representations}}.
\newblock


\bibitem[\protect\citeauthoryear{Klicpera, Bojchevski, and
  G{\"u}nnemann}{Klicpera et~al\mbox{.}}{2019}]%
        {klicpera2018predict}
\bibfield{author}{\bibinfo{person}{Johannes Klicpera},
  \bibinfo{person}{Aleksandar Bojchevski}, {and} \bibinfo{person}{Stephan
  G{\"u}nnemann}.} \bibinfo{year}{2019}\natexlab{}.
\newblock \showarticletitle{Predict then propagate: Graph neural networks meet
  personalized pagerank}. In \bibinfo{booktitle}{\emph{International Conference
  on Learning Representations}}.
\newblock


\bibitem[\protect\citeauthoryear{Kumar and Varaiya}{Kumar and Varaiya}{2015}]%
        {kumar2015stochastic}
\bibfield{author}{\bibinfo{person}{Panqanamala~Ramana Kumar} {and}
  \bibinfo{person}{Pravin Varaiya}.} \bibinfo{year}{2015}\natexlab{}.
\newblock \bibinfo{booktitle}{\emph{Stochastic systems: Estimation,
  identification, and adaptive control}}. Vol.~\bibinfo{volume}{75}.
\newblock \bibinfo{publisher}{SIAM}.
\newblock


\bibitem[\protect\citeauthoryear{Lee, Lee, and Kang}{Lee et~al\mbox{.}}{2019}]%
        {lee2019self}
\bibfield{author}{\bibinfo{person}{Junhyun Lee}, \bibinfo{person}{Inyeop Lee},
  {and} \bibinfo{person}{Jaewoo Kang}.} \bibinfo{year}{2019}\natexlab{}.
\newblock \showarticletitle{Self-Attention Graph Pooling}. In
  \bibinfo{booktitle}{\emph{International Conference on Machine Learning}}.
  \bibinfo{pages}{3734--3743}.
\newblock


\bibitem[\protect\citeauthoryear{Li, Han, and Wu}{Li et~al\mbox{.}}{2018}]%
        {li2018deeper}
\bibfield{author}{\bibinfo{person}{Qimai Li}, \bibinfo{person}{Zhichao Han},
  {and} \bibinfo{person}{Xiao-Ming Wu}.} \bibinfo{year}{2018}\natexlab{}.
\newblock \showarticletitle{Deeper insights into graph convolutional networks
  for semi-supervised learning}. In \bibinfo{booktitle}{\emph{Thirty-Second
  AAAI Conference on Artificial Intelligence}}.
\newblock


\bibitem[\protect\citeauthoryear{Liu, Wang, and Ji}{Liu et~al\mbox{.}}{2020}]%
        {liu2020non}
\bibfield{author}{\bibinfo{person}{Meng Liu}, \bibinfo{person}{Zhengyang Wang},
  {and} \bibinfo{person}{Shuiwang Ji}.} \bibinfo{year}{2020}\natexlab{}.
\newblock \showarticletitle{Non-Local Graph Neural Networks}.
\newblock \bibinfo{journal}{\emph{arXiv preprint arXiv:2005.14612}}
  (\bibinfo{year}{2020}).
\newblock


\bibitem[\protect\citeauthoryear{Lov{\'a}sz et~al\mbox{.}}{Lov{\'a}sz
  et~al\mbox{.}}{1993}]%
        {lovasz1993random}
\bibfield{author}{\bibinfo{person}{L{\'a}szl{\'o} Lov{\'a}sz} {et~al\mbox{.}}}
  \bibinfo{year}{1993}\natexlab{}.
\newblock \showarticletitle{Random walks on graphs: A survey}.
\newblock \bibinfo{journal}{\emph{Combinatorics, Paul erdos is eighty}}
  \bibinfo{volume}{2}, \bibinfo{number}{1} (\bibinfo{year}{1993}),
  \bibinfo{pages}{1--46}.
\newblock


\bibitem[\protect\citeauthoryear{Ma, Wang, Aggarwal, and Tang}{Ma
  et~al\mbox{.}}{2019}]%
        {ma2019graph}
\bibfield{author}{\bibinfo{person}{Yao Ma}, \bibinfo{person}{Suhang Wang},
  \bibinfo{person}{Charu~C Aggarwal}, {and} \bibinfo{person}{Jiliang Tang}.}
  \bibinfo{year}{2019}\natexlab{}.
\newblock \showarticletitle{Graph convolutional networks with eigenpooling}. In
  \bibinfo{booktitle}{\emph{Proceedings of the 25th ACM SIGKDD International
  Conference on Knowledge Discovery \& Data Mining}}.
  \bibinfo{pages}{723--731}.
\newblock


\bibitem[\protect\citeauthoryear{Maaten and Hinton}{Maaten and Hinton}{2008}]%
        {maaten2008visualizing}
\bibfield{author}{\bibinfo{person}{Laurens van~der Maaten} {and}
  \bibinfo{person}{Geoffrey Hinton}.} \bibinfo{year}{2008}\natexlab{}.
\newblock \showarticletitle{Visualizing data using t-SNE}.
\newblock \bibinfo{journal}{\emph{Journal of machine learning research}}
  \bibinfo{volume}{9}, \bibinfo{number}{Nov} (\bibinfo{year}{2008}),
  \bibinfo{pages}{2579--2605}.
\newblock


\bibitem[\protect\citeauthoryear{McAuley, Targett, Shi, and Van
  Den~Hengel}{McAuley et~al\mbox{.}}{2015}]%
        {mcauley2015image}
\bibfield{author}{\bibinfo{person}{Julian McAuley},
  \bibinfo{person}{Christopher Targett}, \bibinfo{person}{Qinfeng Shi}, {and}
  \bibinfo{person}{Anton Van Den~Hengel}.} \bibinfo{year}{2015}\natexlab{}.
\newblock \showarticletitle{Image-based recommendations on styles and
  substitutes}. In \bibinfo{booktitle}{\emph{Proceedings of the 38th
  International ACM SIGIR Conference on Research and Development in Information
  Retrieval}}. \bibinfo{pages}{43--52}.
\newblock


\bibitem[\protect\citeauthoryear{Monti, Boscaini, Masci, Rodola, Svoboda, and
  Bronstein}{Monti et~al\mbox{.}}{2017}]%
        {monti2017geometric}
\bibfield{author}{\bibinfo{person}{Federico Monti}, \bibinfo{person}{Davide
  Boscaini}, \bibinfo{person}{Jonathan Masci}, \bibinfo{person}{Emanuele
  Rodola}, \bibinfo{person}{Jan Svoboda}, {and} \bibinfo{person}{Michael~M
  Bronstein}.} \bibinfo{year}{2017}\natexlab{}.
\newblock \showarticletitle{Geometric deep learning on graphs and manifolds
  using mixture model cnns}. In \bibinfo{booktitle}{\emph{Proceedings of the
  IEEE Conference on Computer Vision and Pattern Recognition}}.
  \bibinfo{pages}{5115--5124}.
\newblock


\bibitem[\protect\citeauthoryear{Nair and Hinton}{Nair and Hinton}{2010}]%
        {nair2010rectified}
\bibfield{author}{\bibinfo{person}{Vinod Nair} {and}
  \bibinfo{person}{Geoffrey~E Hinton}.} \bibinfo{year}{2010}\natexlab{}.
\newblock \showarticletitle{Rectified linear units improve restricted boltzmann
  machines}. In \bibinfo{booktitle}{\emph{International Conference on Machine
  Learning}}. \bibinfo{pages}{807--814}.
\newblock


\bibitem[\protect\citeauthoryear{Page, Brin, Motwani, and Winograd}{Page
  et~al\mbox{.}}{1999}]%
        {page1999pagerank}
\bibfield{author}{\bibinfo{person}{Lawrence Page}, \bibinfo{person}{Sergey
  Brin}, \bibinfo{person}{Rajeev Motwani}, {and} \bibinfo{person}{Terry
  Winograd}.} \bibinfo{year}{1999}\natexlab{}.
\newblock \bibinfo{booktitle}{\emph{The PageRank citation ranking: Bringing
  order to the web.}}
\newblock \bibinfo{type}{{T}echnical {R}eport}. \bibinfo{institution}{Stanford
  InfoLab}.
\newblock


\bibitem[\protect\citeauthoryear{Paszke, Gross, Chintala, Chanan, Yang, DeVito,
  Lin, Desmaison, Antiga, and Lerer}{Paszke et~al\mbox{.}}{2017}]%
        {paszke2017automatic}
\bibfield{author}{\bibinfo{person}{Adam Paszke}, \bibinfo{person}{Sam Gross},
  \bibinfo{person}{Soumith Chintala}, \bibinfo{person}{Gregory Chanan},
  \bibinfo{person}{Edward Yang}, \bibinfo{person}{Zachary DeVito},
  \bibinfo{person}{Zeming Lin}, \bibinfo{person}{Alban Desmaison},
  \bibinfo{person}{Luca Antiga}, {and} \bibinfo{person}{Adam Lerer}.}
  \bibinfo{year}{2017}\natexlab{}.
\newblock \showarticletitle{Automatic differentiation in pytorch}. In
  \bibinfo{booktitle}{\emph{Proceedings of Neural Information Processing
  Systems Autodiff Workshop}}.
\newblock


\bibitem[\protect\citeauthoryear{Pei, Wei, Chang, Lei, and Yang}{Pei
  et~al\mbox{.}}{2020}]%
        {pei2020geom}
\bibfield{author}{\bibinfo{person}{Hongbin Pei}, \bibinfo{person}{Bingzhe Wei},
  \bibinfo{person}{Kevin Chen-Chuan Chang}, \bibinfo{person}{Yu Lei}, {and}
  \bibinfo{person}{Bo Yang}.} \bibinfo{year}{2020}\natexlab{}.
\newblock \showarticletitle{Geom-gcn: Geometric graph convolutional networks}.
  In \bibinfo{booktitle}{\emph{International Conference on Learning
  Representations}}.
\newblock


\bibitem[\protect\citeauthoryear{Sen, Namata, Bilgic, Getoor, Galligher, and
  Eliassi-Rad}{Sen et~al\mbox{.}}{2008}]%
        {sen2008collective}
\bibfield{author}{\bibinfo{person}{Prithviraj Sen}, \bibinfo{person}{Galileo
  Namata}, \bibinfo{person}{Mustafa Bilgic}, \bibinfo{person}{Lise Getoor},
  \bibinfo{person}{Brian Galligher}, {and} \bibinfo{person}{Tina Eliassi-Rad}.}
  \bibinfo{year}{2008}\natexlab{}.
\newblock \showarticletitle{Collective classification in network data}.
\newblock \bibinfo{journal}{\emph{AI magazine}} \bibinfo{volume}{29},
  \bibinfo{number}{3} (\bibinfo{year}{2008}), \bibinfo{pages}{93--93}.
\newblock


\bibitem[\protect\citeauthoryear{Seneta}{Seneta}{2006}]%
        {seneta2006non}
\bibfield{author}{\bibinfo{person}{Eugene Seneta}.}
  \bibinfo{year}{2006}\natexlab{}.
\newblock \bibinfo{booktitle}{\emph{Non-negative matrices and Markov chains}}.
\newblock \bibinfo{publisher}{Springer Science \& Business Media}.
\newblock


\bibitem[\protect\citeauthoryear{Shchur, Mumme, Bojchevski, and
  G{\"u}nnemann}{Shchur et~al\mbox{.}}{2018}]%
        {shchur2018pitfalls}
\bibfield{author}{\bibinfo{person}{Oleksandr Shchur},
  \bibinfo{person}{Maximilian Mumme}, \bibinfo{person}{Aleksandar Bojchevski},
  {and} \bibinfo{person}{Stephan G{\"u}nnemann}.}
  \bibinfo{year}{2018}\natexlab{}.
\newblock \showarticletitle{Pitfalls of graph neural network evaluation}.
\newblock \bibinfo{journal}{\emph{arXiv preprint arXiv:1811.05868}}
  (\bibinfo{year}{2018}).
\newblock


\bibitem[\protect\citeauthoryear{Taubin}{Taubin}{1995}]%
        {taubin1995signal}
\bibfield{author}{\bibinfo{person}{Gabriel Taubin}.}
  \bibinfo{year}{1995}\natexlab{}.
\newblock \showarticletitle{A signal processing approach to fair surface
  design}. In \bibinfo{booktitle}{\emph{Proceedings of the 22nd annual
  conference on Computer graphics and interactive techniques}}. ACM,
  \bibinfo{pages}{351--358}.
\newblock


\bibitem[\protect\citeauthoryear{Veli{\v{c}}kovi{\'c}, Cucurull, Casanova,
  Romero, Lio, and Bengio}{Veli{\v{c}}kovi{\'c} et~al\mbox{.}}{2018}]%
        {velivckovic2017graph}
\bibfield{author}{\bibinfo{person}{Petar Veli{\v{c}}kovi{\'c}},
  \bibinfo{person}{Guillem Cucurull}, \bibinfo{person}{Arantxa Casanova},
  \bibinfo{person}{Adriana Romero}, \bibinfo{person}{Pietro Lio}, {and}
  \bibinfo{person}{Yoshua Bengio}.} \bibinfo{year}{2018}\natexlab{}.
\newblock \showarticletitle{Graph attention networks}. In
  \bibinfo{booktitle}{\emph{International Conference on Learning
  Representation}}.
\newblock


\bibitem[\protect\citeauthoryear{Wu, Souza, Zhang, Fifty, Yu, and
  Weinberger}{Wu et~al\mbox{.}}{2019}]%
        {wu2019simplifying}
\bibfield{author}{\bibinfo{person}{Felix Wu}, \bibinfo{person}{Amauri Souza},
  \bibinfo{person}{Tianyi Zhang}, \bibinfo{person}{Christopher Fifty},
  \bibinfo{person}{Tao Yu}, {and} \bibinfo{person}{Kilian Weinberger}.}
  \bibinfo{year}{2019}\natexlab{}.
\newblock \showarticletitle{Simplifying Graph Convolutional Networks}. In
  \bibinfo{booktitle}{\emph{International Conference on Machine Learning}}.
  \bibinfo{pages}{6861--6871}.
\newblock


\bibitem[\protect\citeauthoryear{Xu, Hu, Leskovec, and Jegelka}{Xu
  et~al\mbox{.}}{2019}]%
        {xu2018powerful}
\bibfield{author}{\bibinfo{person}{Keyulu Xu}, \bibinfo{person}{Weihua Hu},
  \bibinfo{person}{Jure Leskovec}, {and} \bibinfo{person}{Stefanie Jegelka}.}
  \bibinfo{year}{2019}\natexlab{}.
\newblock \showarticletitle{How powerful are graph neural networks?}. In
  \bibinfo{booktitle}{\emph{International Conference on Learning
  Representations}}.
\newblock


\bibitem[\protect\citeauthoryear{Xu, Li, Tian, Sonobe, Kawarabayashi, and
  Jegelka}{Xu et~al\mbox{.}}{2018}]%
        {xu2018representation}
\bibfield{author}{\bibinfo{person}{Keyulu Xu}, \bibinfo{person}{Chengtao Li},
  \bibinfo{person}{Yonglong Tian}, \bibinfo{person}{Tomohiro Sonobe},
  \bibinfo{person}{Ken-ichi Kawarabayashi}, {and} \bibinfo{person}{Stefanie
  Jegelka}.} \bibinfo{year}{2018}\natexlab{}.
\newblock \showarticletitle{Representation Learning on Graphs with Jumping
  Knowledge Networks}. In \bibinfo{booktitle}{\emph{International Conference on
  Machine Learning}}. \bibinfo{pages}{5449--5458}.
\newblock


\bibitem[\protect\citeauthoryear{Ying, You, Morris, Ren, Hamilton, and
  Leskovec}{Ying et~al\mbox{.}}{2018}]%
        {ying2018hierarchical}
\bibfield{author}{\bibinfo{person}{Zhitao Ying}, \bibinfo{person}{Jiaxuan You},
  \bibinfo{person}{Christopher Morris}, \bibinfo{person}{Xiang Ren},
  \bibinfo{person}{Will Hamilton}, {and} \bibinfo{person}{Jure Leskovec}.}
  \bibinfo{year}{2018}\natexlab{}.
\newblock \showarticletitle{Hierarchical graph representation learning with
  differentiable pooling}. In \bibinfo{booktitle}{\emph{Advances in neural
  information processing systems}}. \bibinfo{pages}{4800--4810}.
\newblock


\bibitem[\protect\citeauthoryear{Yuan and Ji}{Yuan and Ji}{2020}]%
        {Yuan2020StructPool:}
\bibfield{author}{\bibinfo{person}{Hao Yuan} {and} \bibinfo{person}{Shuiwang
  Ji}.} \bibinfo{year}{2020}\natexlab{}.
\newblock \showarticletitle{StructPool: Structured Graph Pooling via
  Conditional Random Fields}. In \bibinfo{booktitle}{\emph{International
  Conference on Learning Representations}}.
\newblock


\bibitem[\protect\citeauthoryear{Zhang and Chen}{Zhang and Chen}{2017}]%
        {zhang2017weisfeiler}
\bibfield{author}{\bibinfo{person}{Muhan Zhang} {and} \bibinfo{person}{Yixin
  Chen}.} \bibinfo{year}{2017}\natexlab{}.
\newblock \showarticletitle{Weisfeiler-lehman neural machine for link
  prediction}. In \bibinfo{booktitle}{\emph{Proceedings of the 23rd ACM SIGKDD
  International Conference on Knowledge Discovery \& Data Mining}}.
  \bibinfo{pages}{575--583}.
\newblock


\bibitem[\protect\citeauthoryear{Zhang and Chen}{Zhang and Chen}{2018}]%
        {zhang2018link}
\bibfield{author}{\bibinfo{person}{Muhan Zhang} {and} \bibinfo{person}{Yixin
  Chen}.} \bibinfo{year}{2018}\natexlab{}.
\newblock \showarticletitle{Link prediction based on graph neural networks}. In
  \bibinfo{booktitle}{\emph{Advances in Neural Information Processing
  Systems}}. \bibinfo{pages}{5165--5175}.
\newblock


\bibitem[\protect\citeauthoryear{Zhang, Cui, Neumann, and Chen}{Zhang
  et~al\mbox{.}}{2018}]%
        {zhang2018end}
\bibfield{author}{\bibinfo{person}{Muhan Zhang}, \bibinfo{person}{Zhicheng
  Cui}, \bibinfo{person}{Marion Neumann}, {and} \bibinfo{person}{Yixin Chen}.}
  \bibinfo{year}{2018}\natexlab{}.
\newblock \showarticletitle{An end-to-end deep learning architecture for graph
  classification}. In \bibinfo{booktitle}{\emph{Thirty-Second AAAI Conference
  on Artificial Intelligence}}.
\newblock


\end{thebibliography}

\appendix

\clearpage

\section{Appendix}
In this section, we provide necessary information for reproducing our insights and experimental results. These include the quantitative results and qualitative visualization on more datasets that can further support our insights, the proofs of lemmas, and the detailed description of datasets.

\subsection{Test Accuracy and Smoothness Metric Value of GCNs}\label{Sec:acc_gcn}

\begin{figure}[h]
\centering
\subfigure{
\includegraphics[width=0.45\columnwidth]{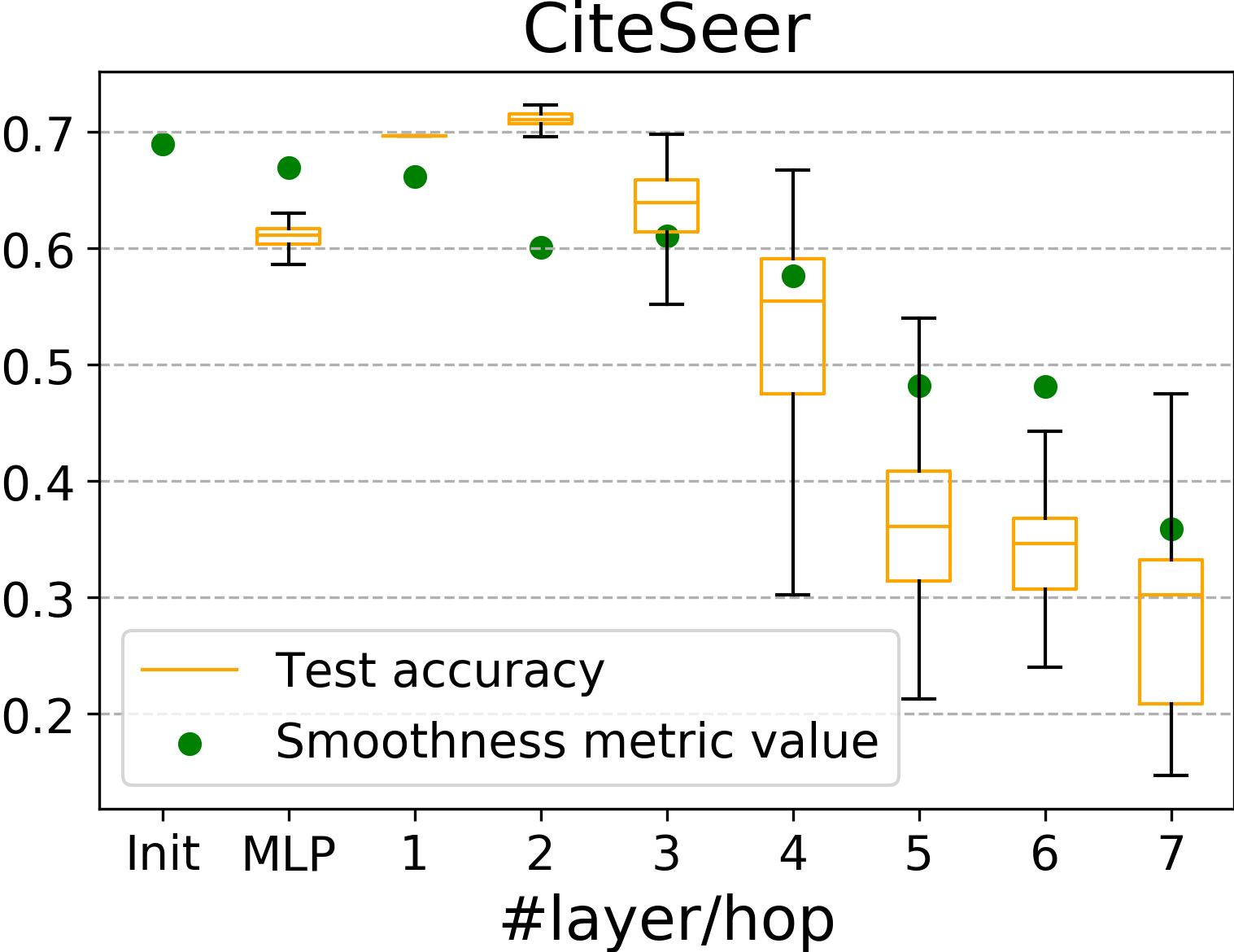}
}
\subfigure{
\includegraphics[width=0.45\columnwidth]{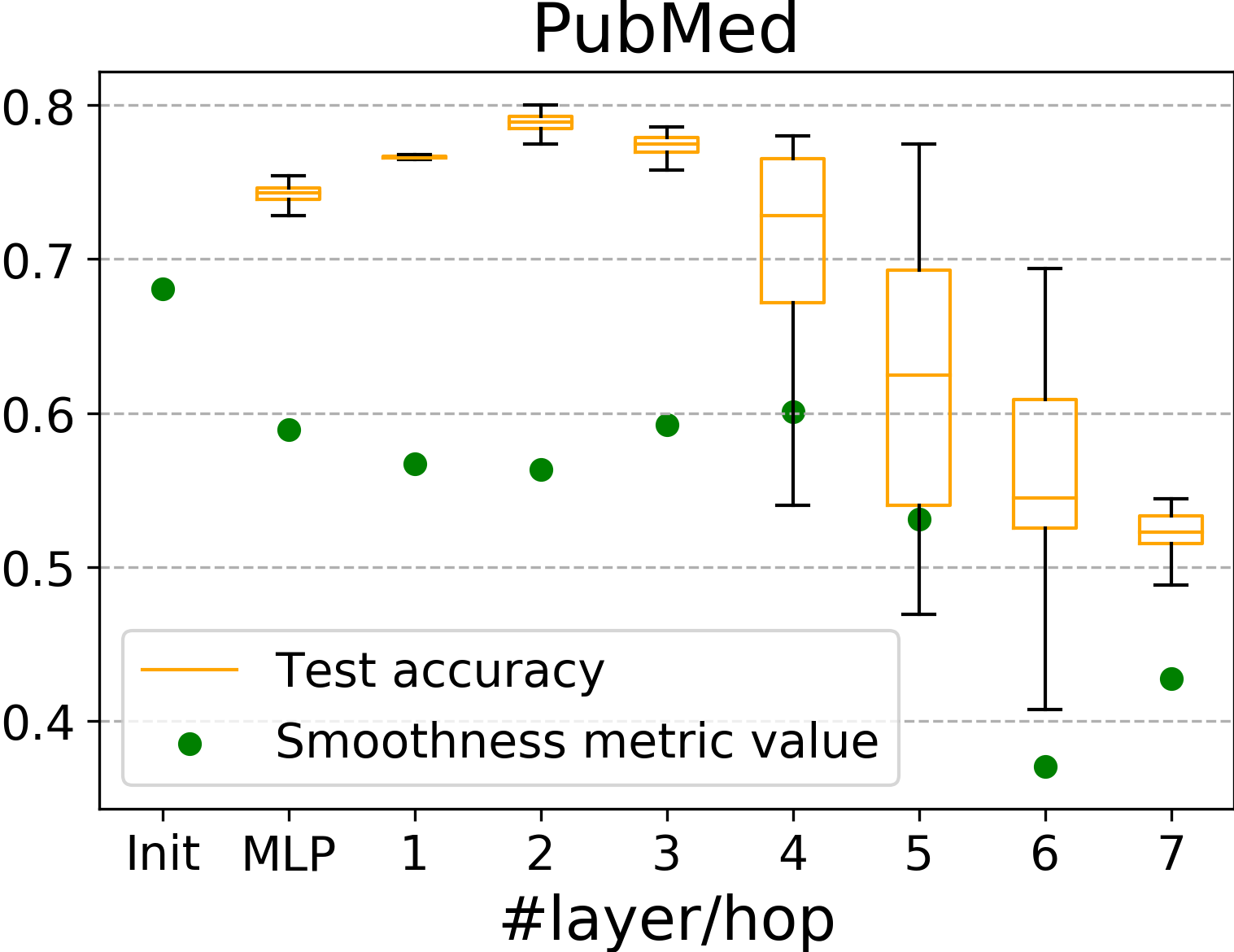}
}
\caption{Test accuracy and smoothness metric value of node representations with different numbers of GCN layers on CiteSeer and PubMed. "Init" means the smoothness metric value of the original data.}
\label{fig:gcn_degrade_citeseer_pubmed}
\end{figure}

Test accuracy and smoothness metric value of node representations with different numbers of GCN layers are shown in Figure \ref{fig:gcn_degrade_citeseer_pubmed} for CiteSeer and PubMed. They have the same trends as we discussed in Section \ref{Sec:why}.

\subsection{Test Accuracy and Smoothness Metric Value of Models as~Eq.(\ref{DetachGCN})}\label{Sec:detachgcn_app}
\begin{figure}[h]
\centering
\subfigure{
\includegraphics[width=0.45\columnwidth]{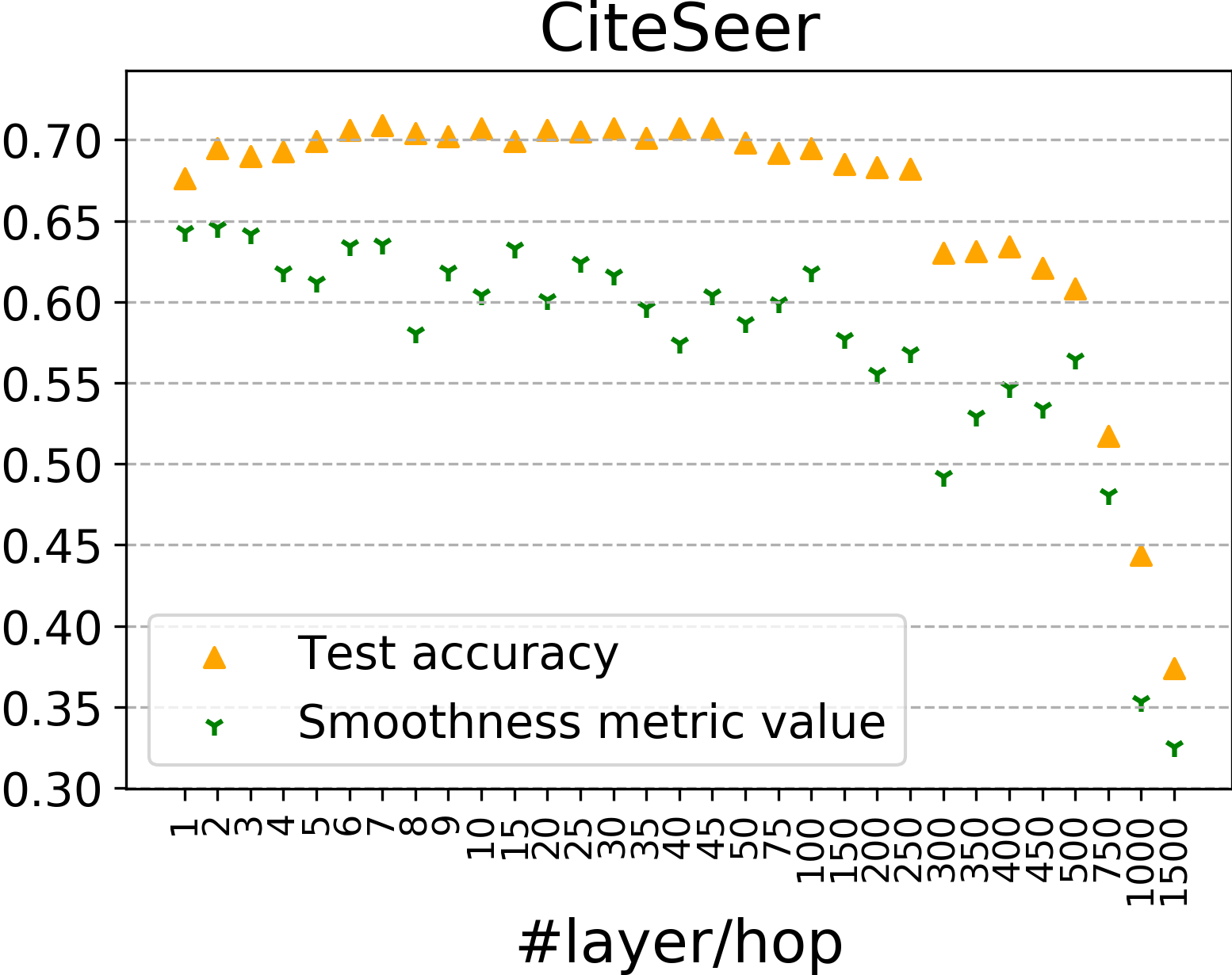}
}
\subfigure{
\includegraphics[width=0.45\columnwidth]{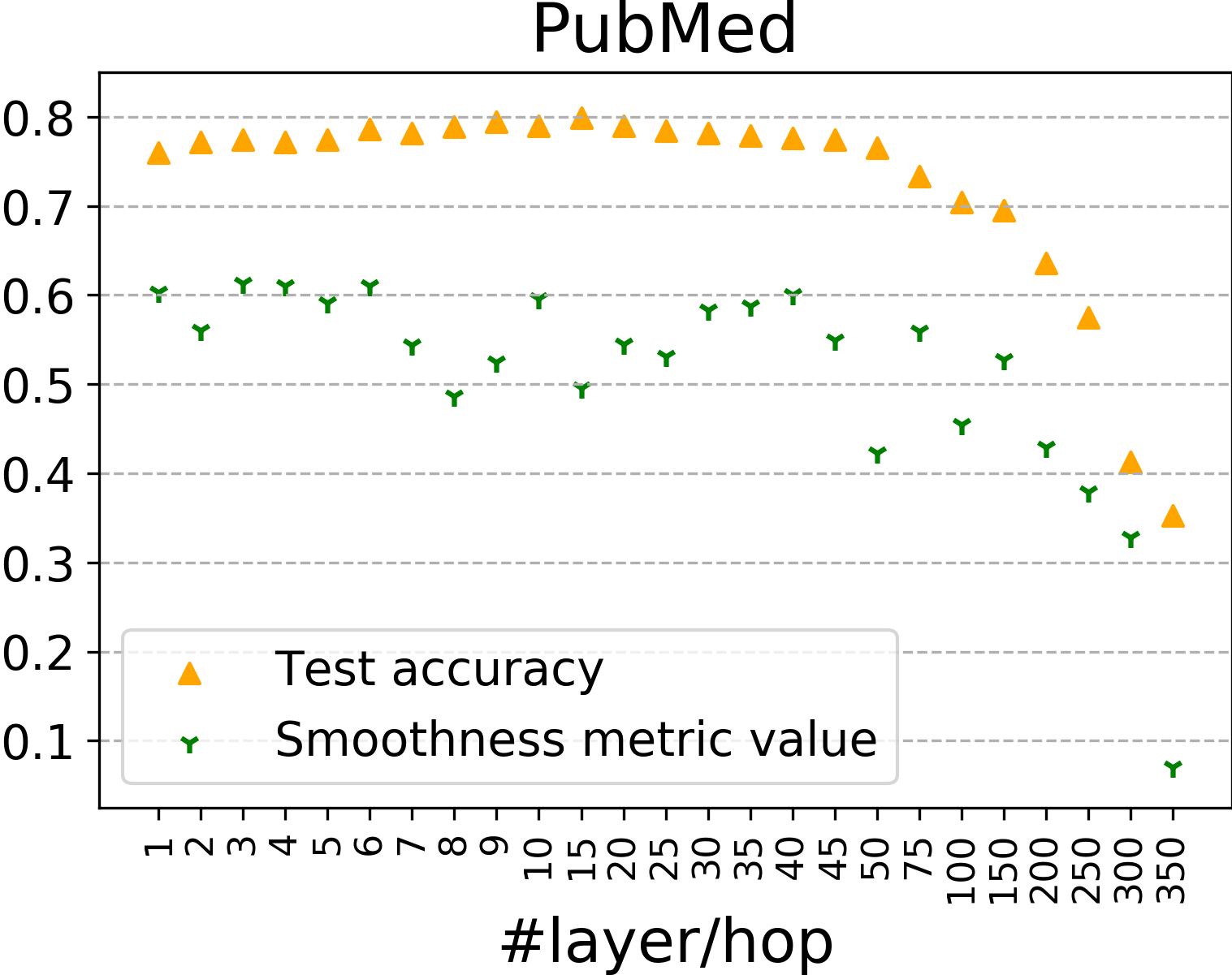}
}
\caption{Test accuracy and smoothness metric value of node representations with different numbers of layers adopted in models as~Eq.(\ref{DetachGCN}) on CiterSeer and PubMed.}
\label{fig:detachgcn_citeseer_pubmed}
\end{figure}

Test accuracy and smoothness metric value of node representations with different numbers of layers adopted in models as~Eq.(\ref{DetachGCN}) are shown in Figure \ref{fig:detachgcn_citeseer_pubmed} for CiteSeer and PubMed. It is illustrated that after decoupling transformation from propagation, we can apply deeper models without suffering from performance degradation.

\subsection{Visualization of Representations Derived by GCNs}\label{Sec:visual_citeseer_pubmed_gcn}

\begin{figure*}
\centering
\subfigure{
\includegraphics[width=0.205\columnwidth]{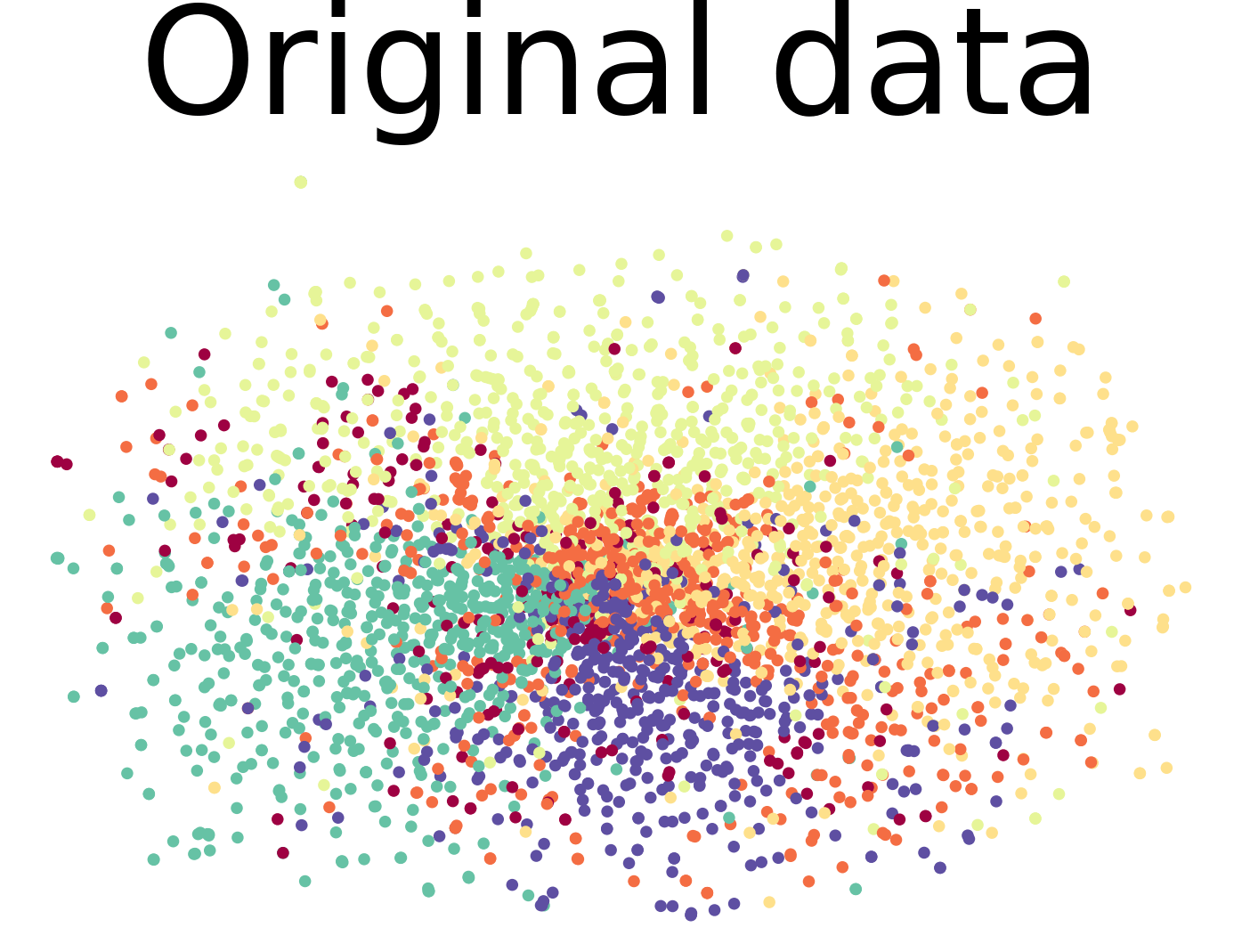}
}
\subfigure{
\includegraphics[width=0.205\columnwidth]{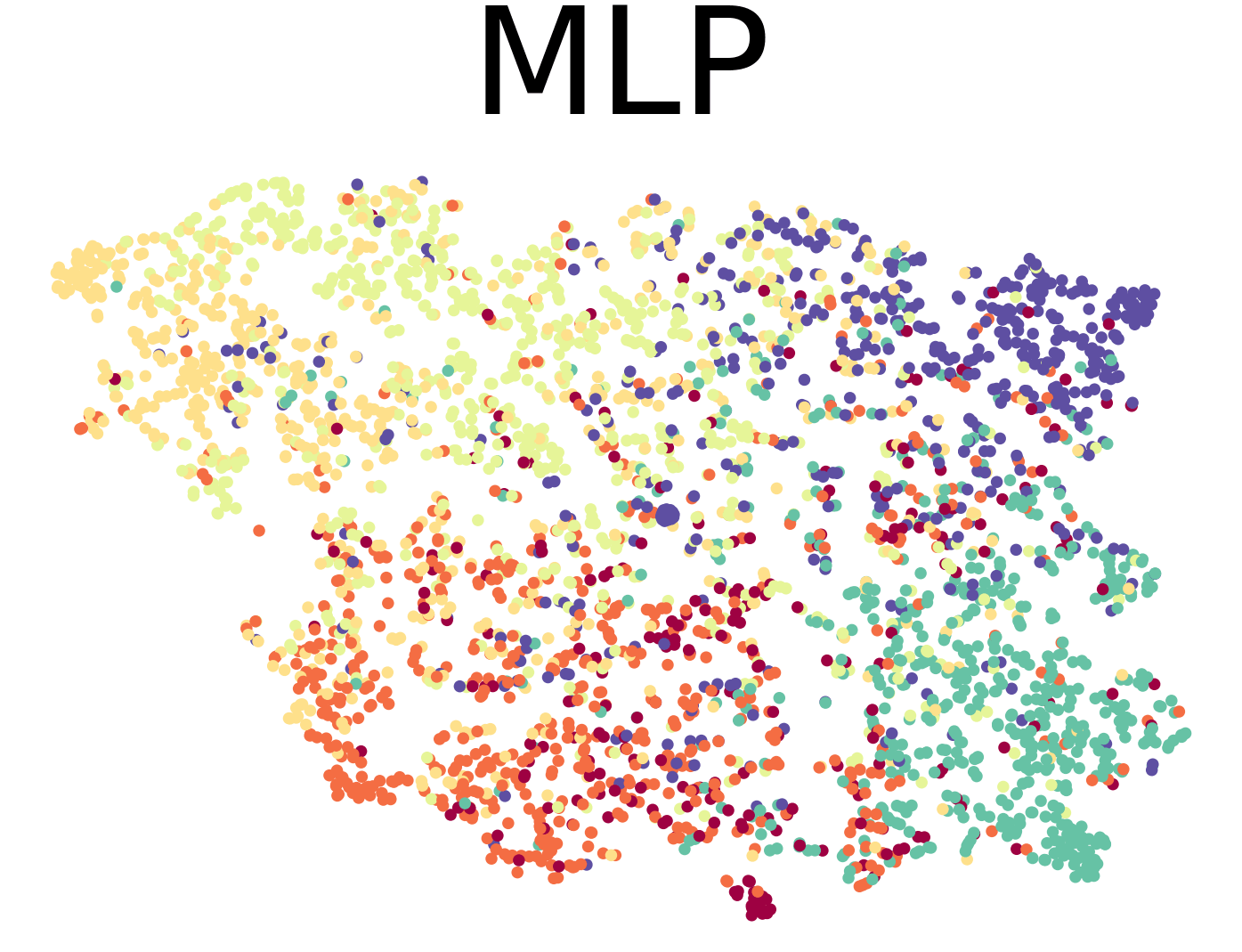}
}
\subfigure{
\includegraphics[width=0.205\columnwidth]{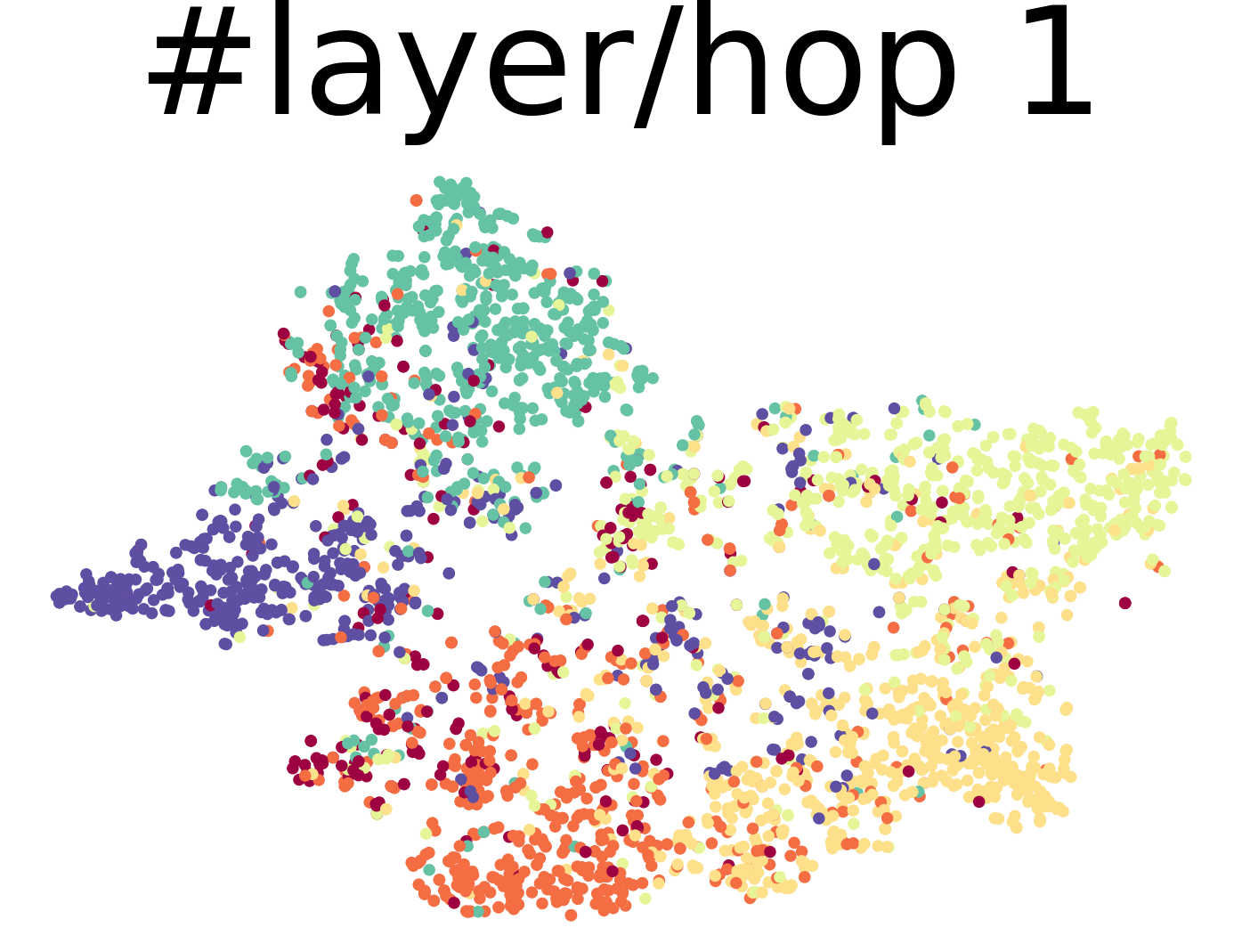}
}
\subfigure{
\includegraphics[width=0.205\columnwidth]{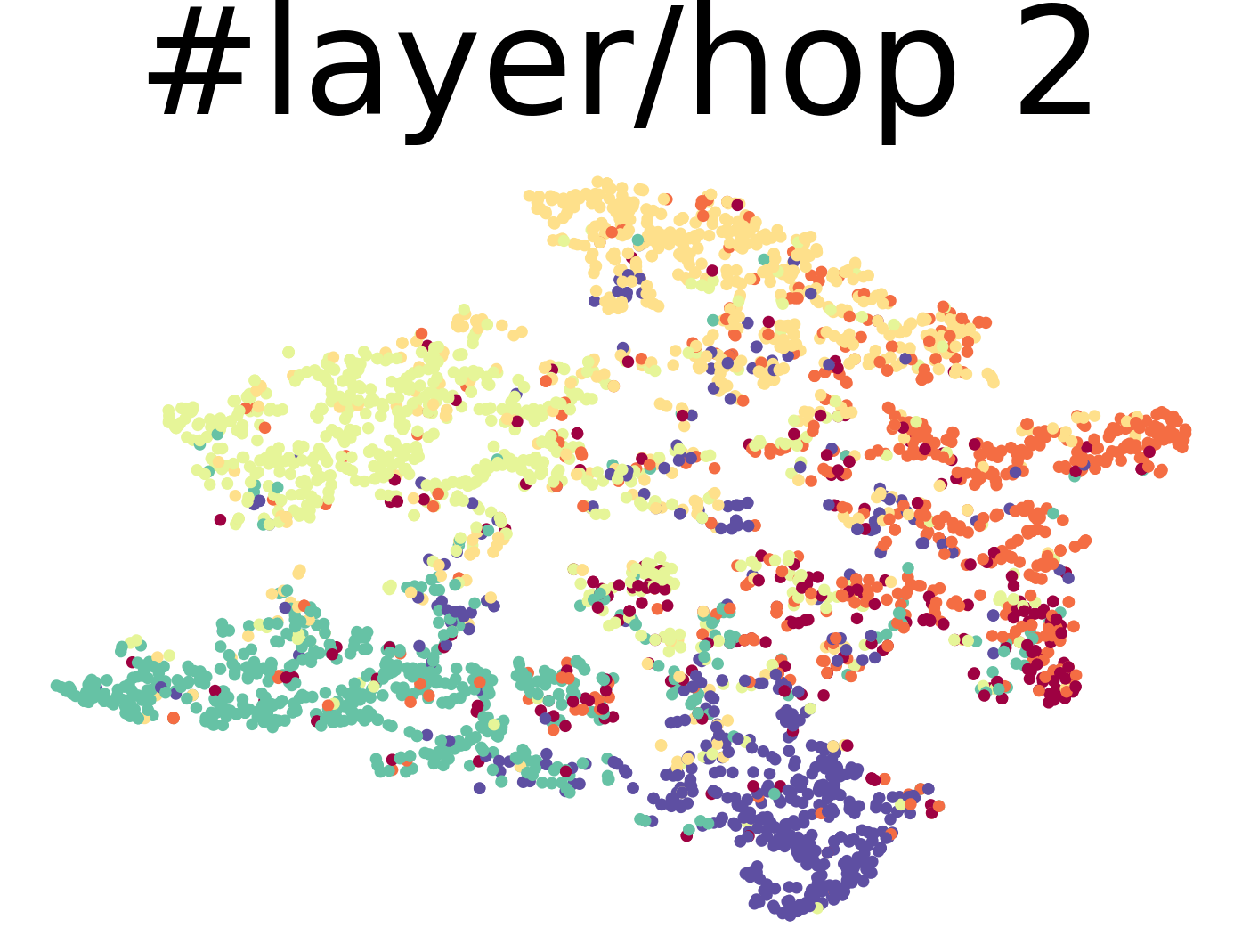}
}
\subfigure{
\includegraphics[width=0.205\columnwidth]{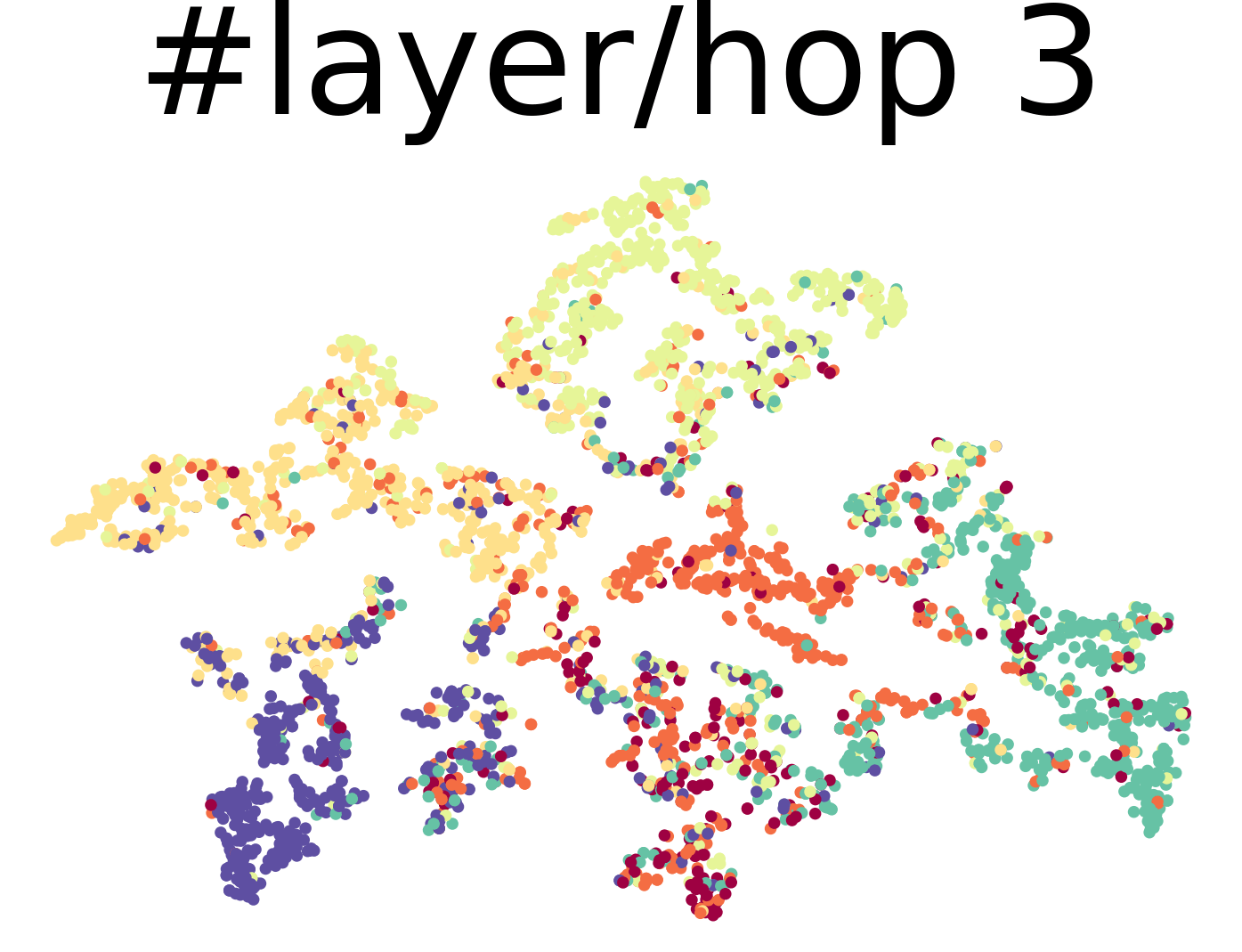}
}
\subfigure{
\includegraphics[width=0.205\columnwidth]{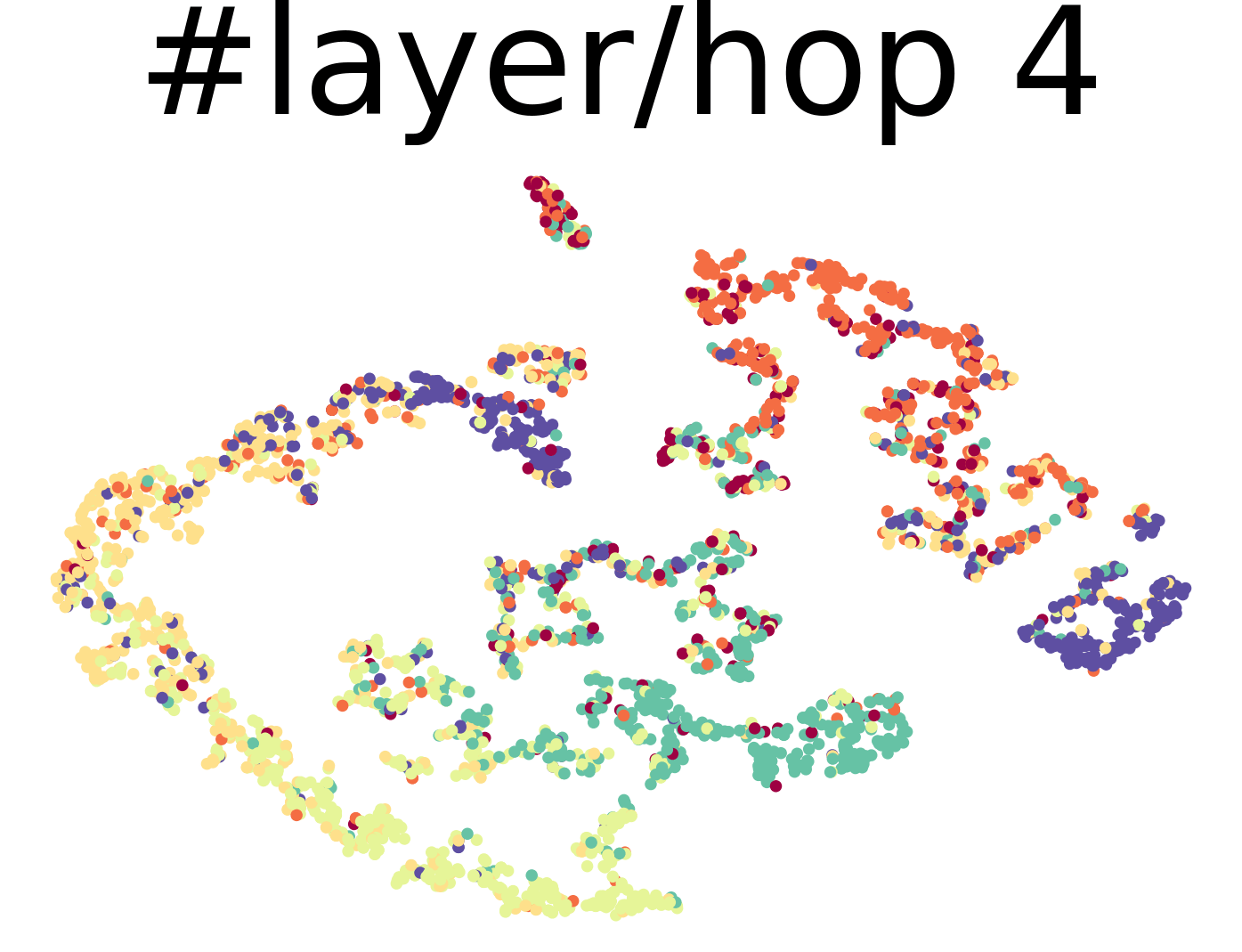}
}
\subfigure{
\includegraphics[width=0.205\columnwidth]{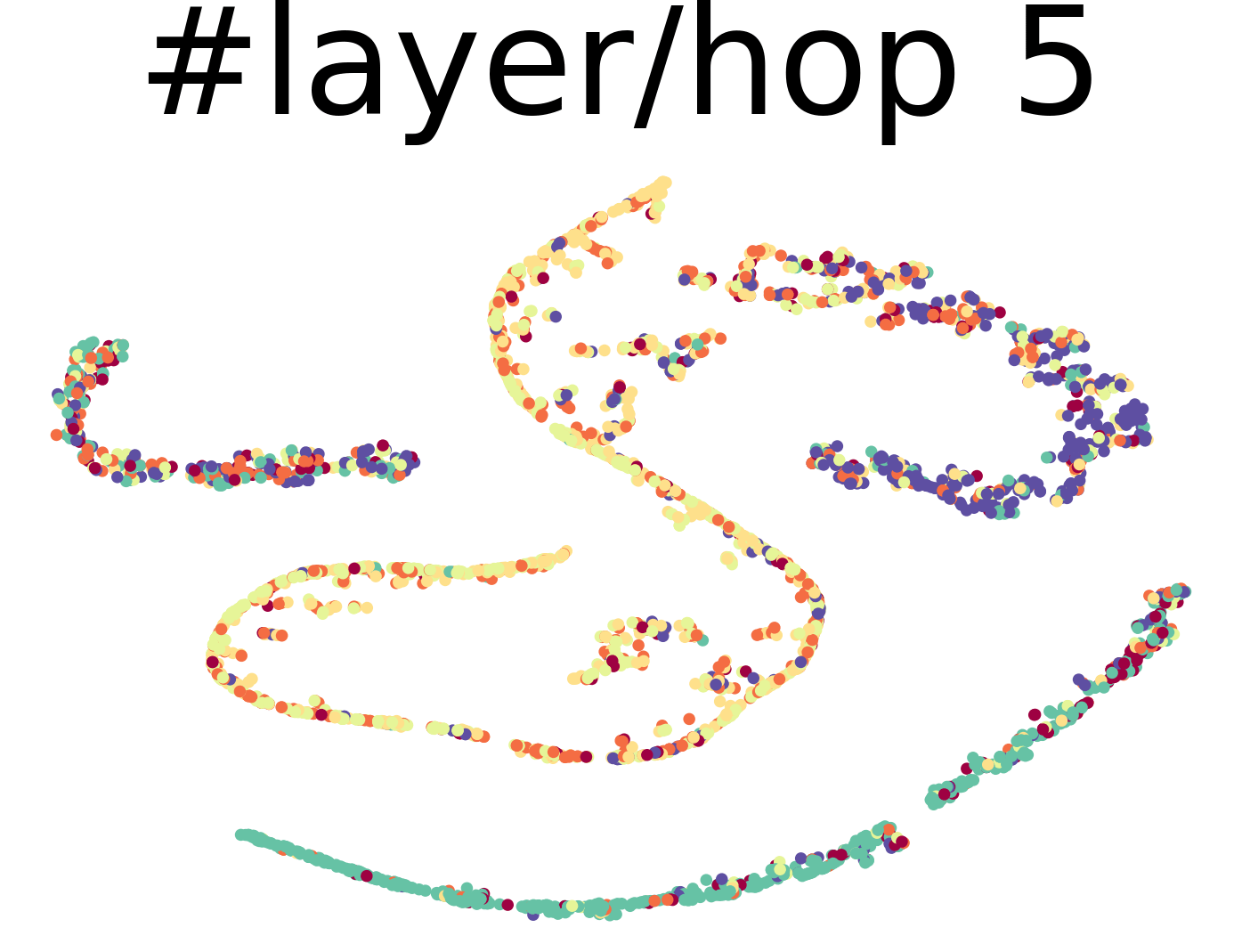}
}
\subfigure{
\includegraphics[width=0.205\columnwidth]{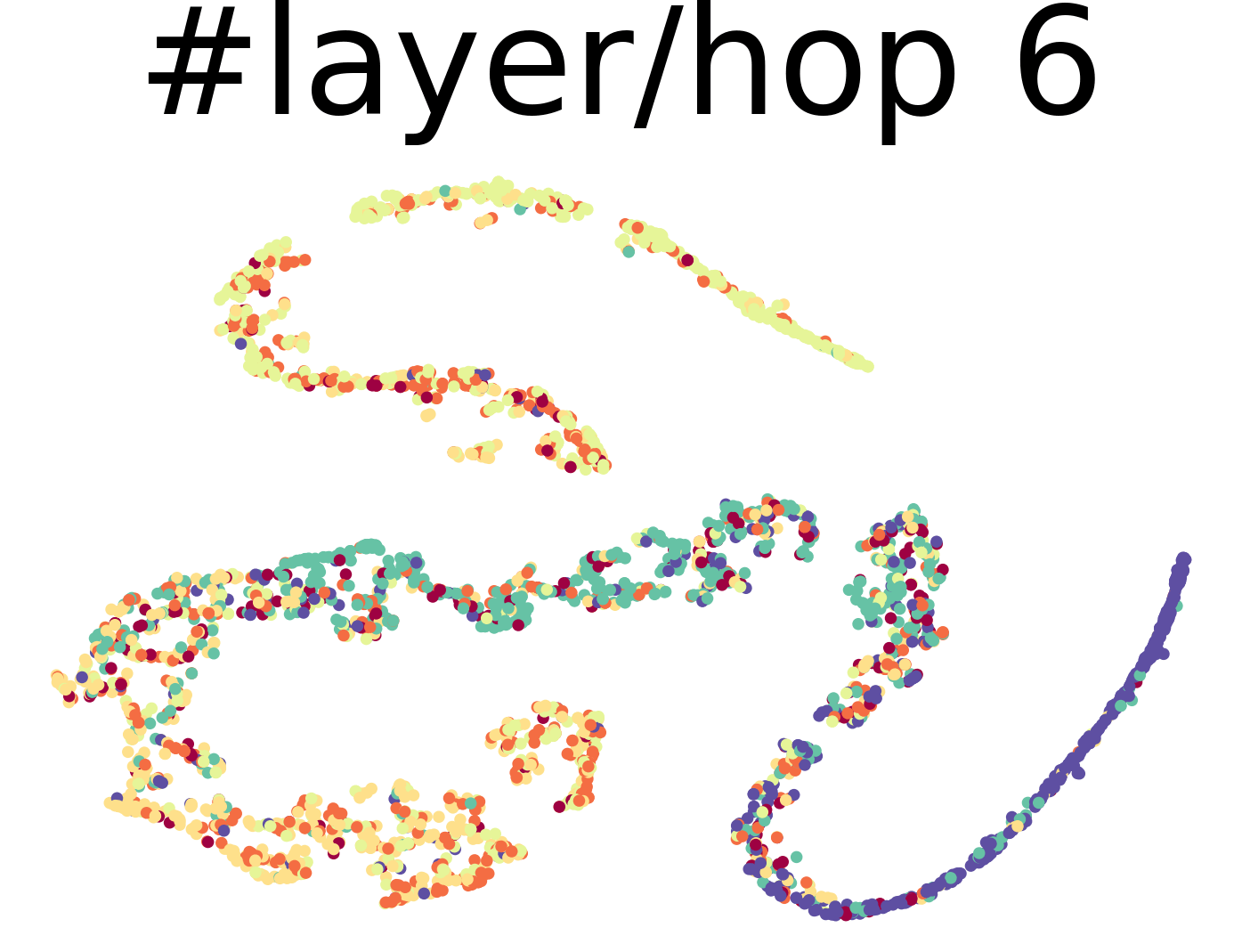}
}
\subfigure{
\includegraphics[width=0.205\columnwidth]{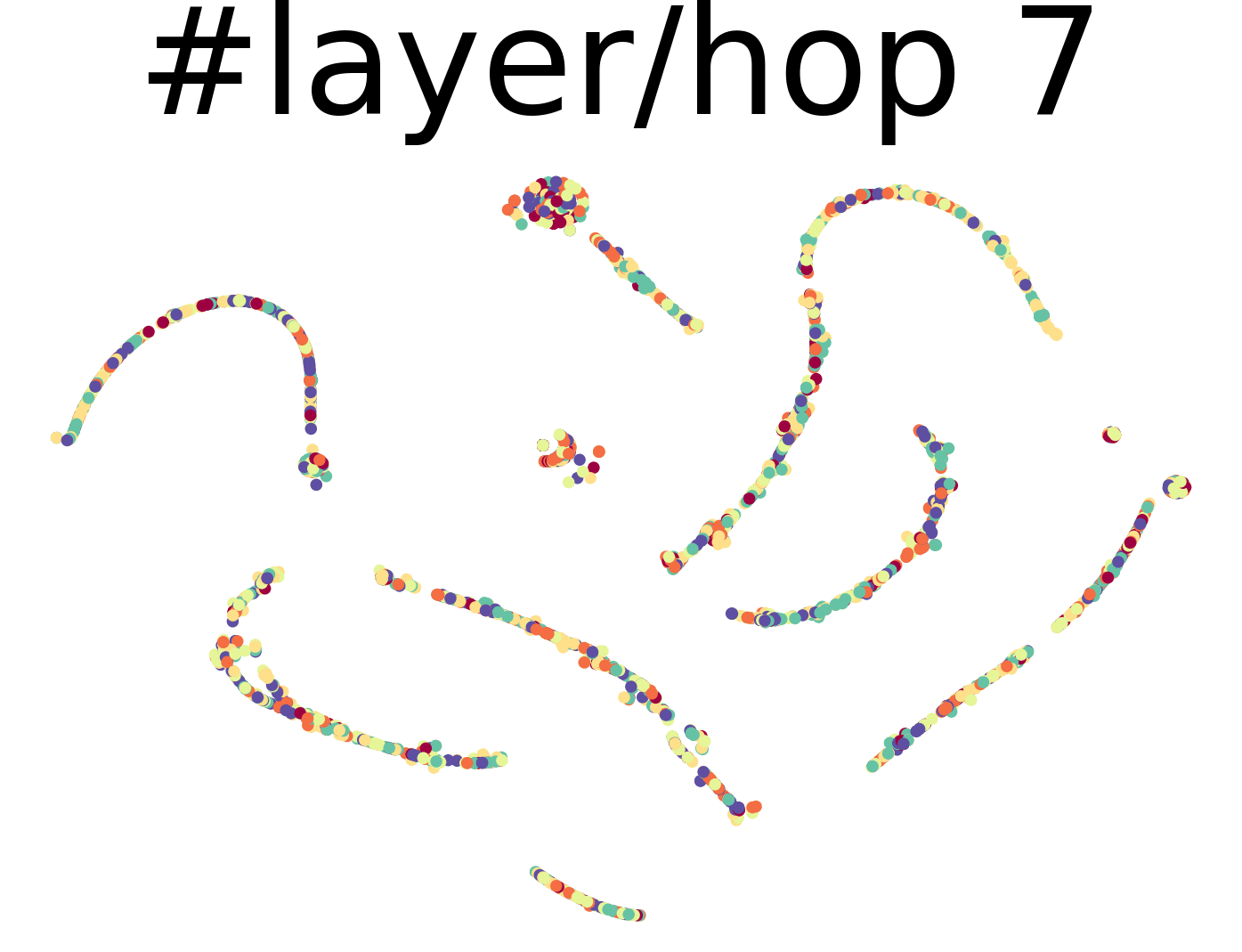}
}
\caption{t-SNE visualization of node representations derived by different numbers of GCN layers on CiteSeer. Colors represent node classes.}
\label{fig:tsne1_citeseer}
\end{figure*}

\begin{figure*}
\centering
\subfigure{
\includegraphics[width=0.205\columnwidth]{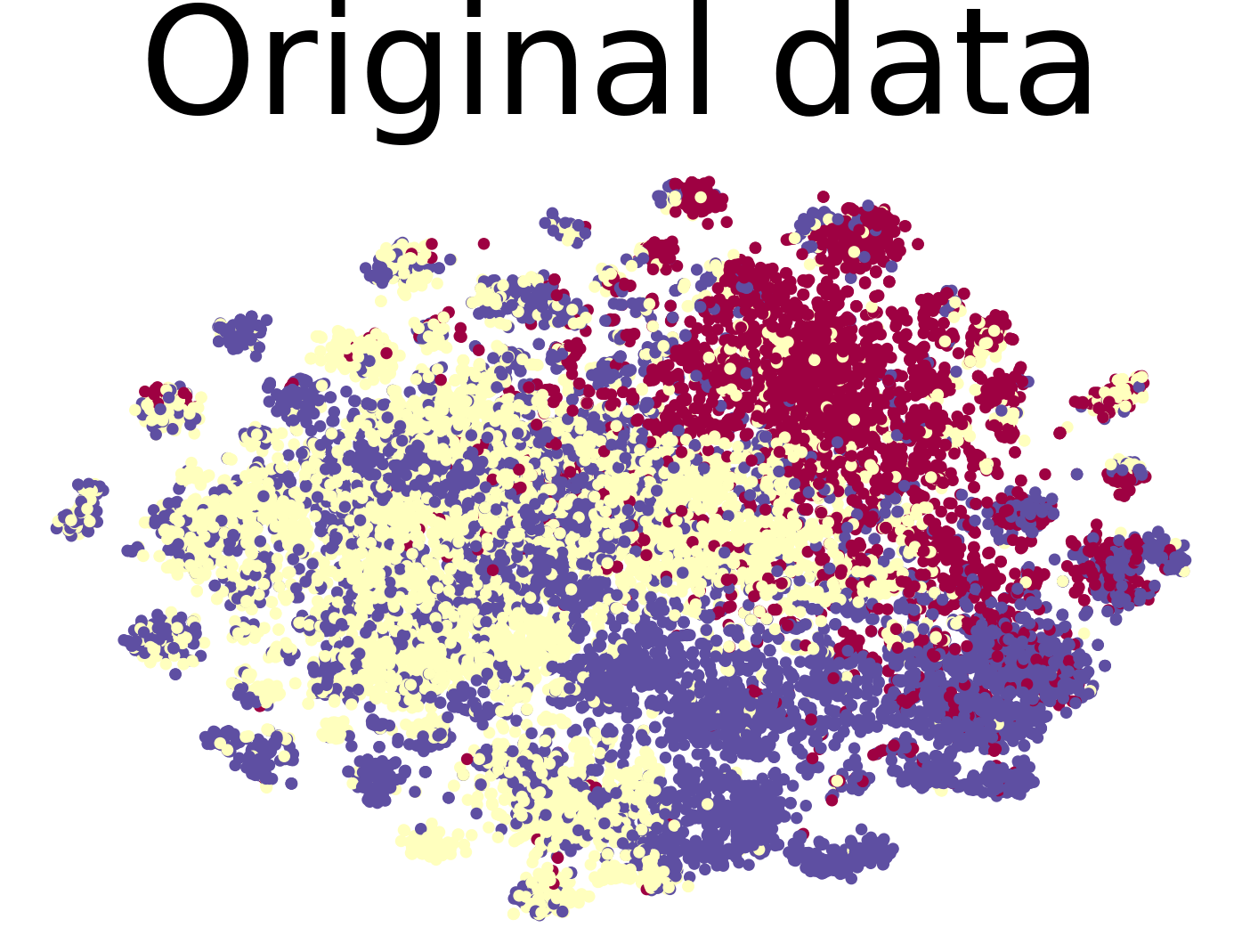}
}
\subfigure{
\includegraphics[width=0.205\columnwidth]{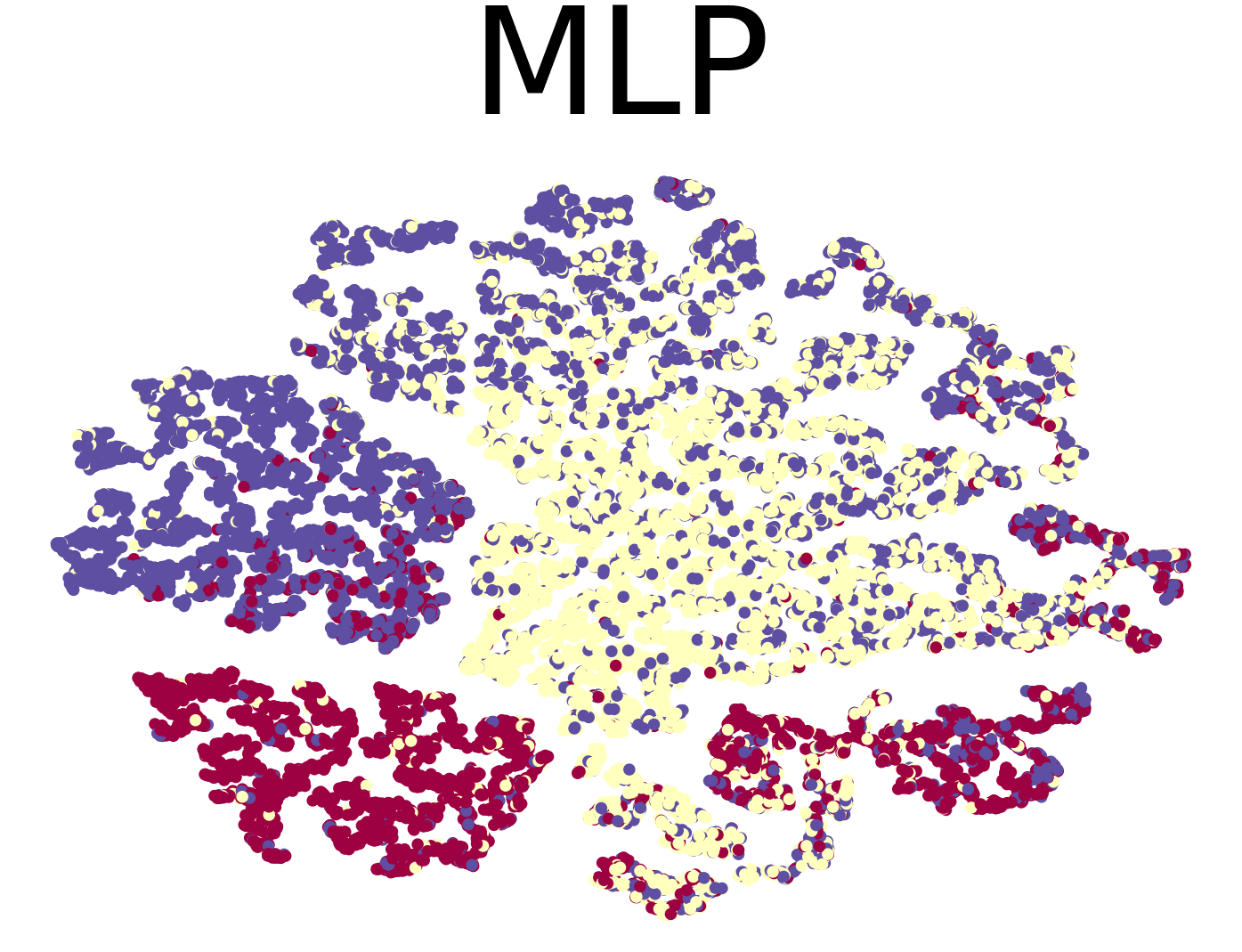}
}
\subfigure{
\includegraphics[width=0.205\columnwidth]{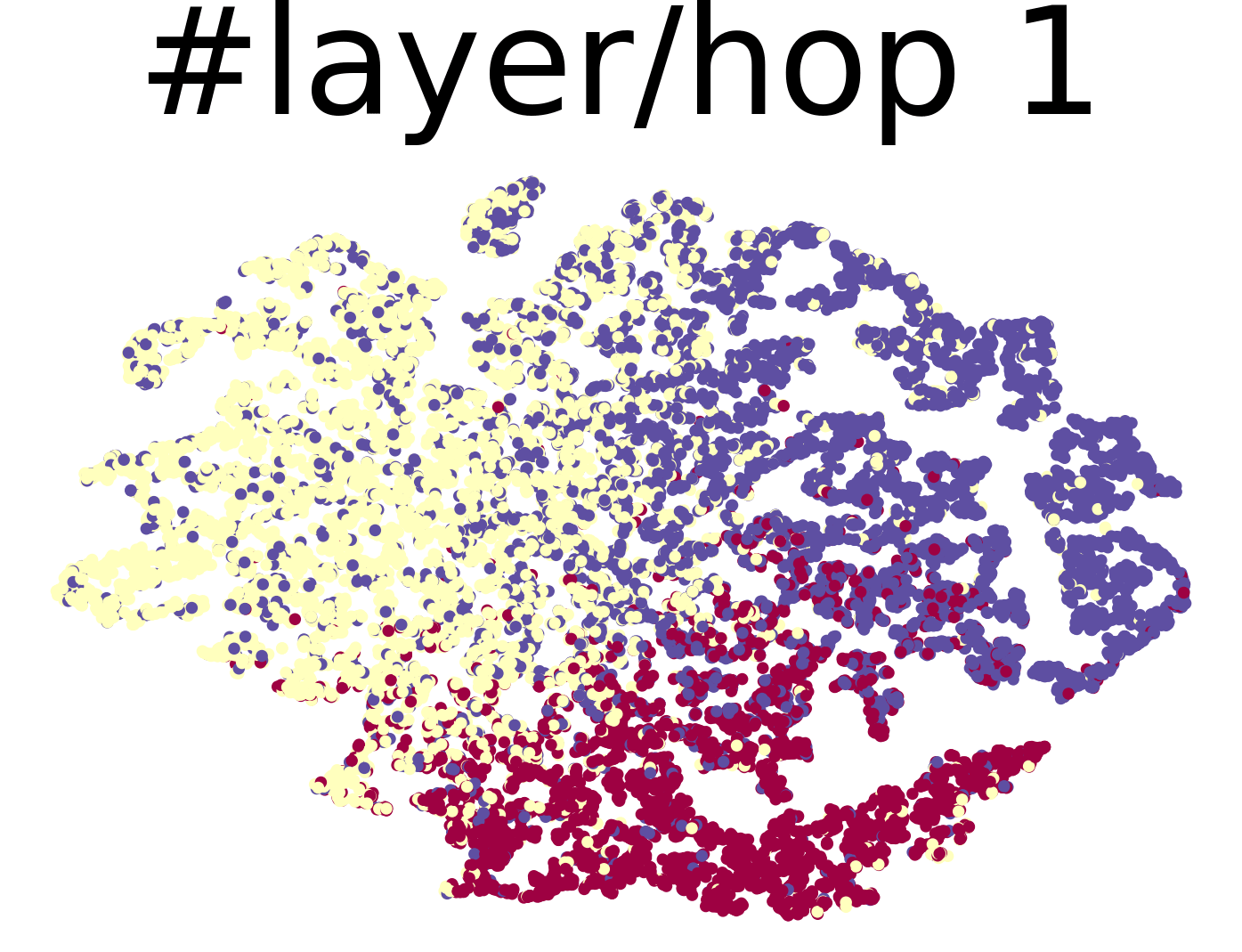}
}
\subfigure{
\includegraphics[width=0.205\columnwidth]{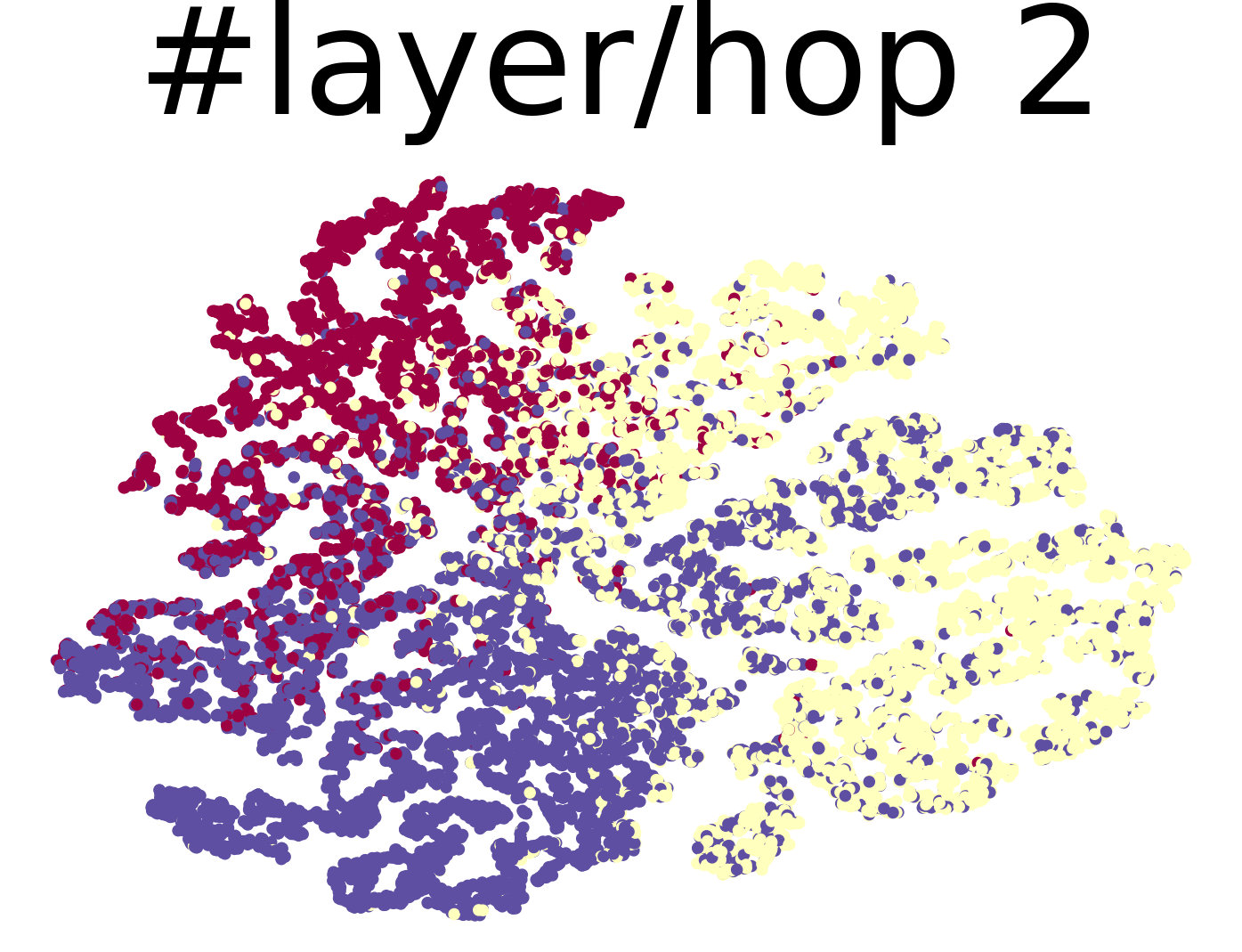}
}
\subfigure{
\includegraphics[width=0.205\columnwidth]{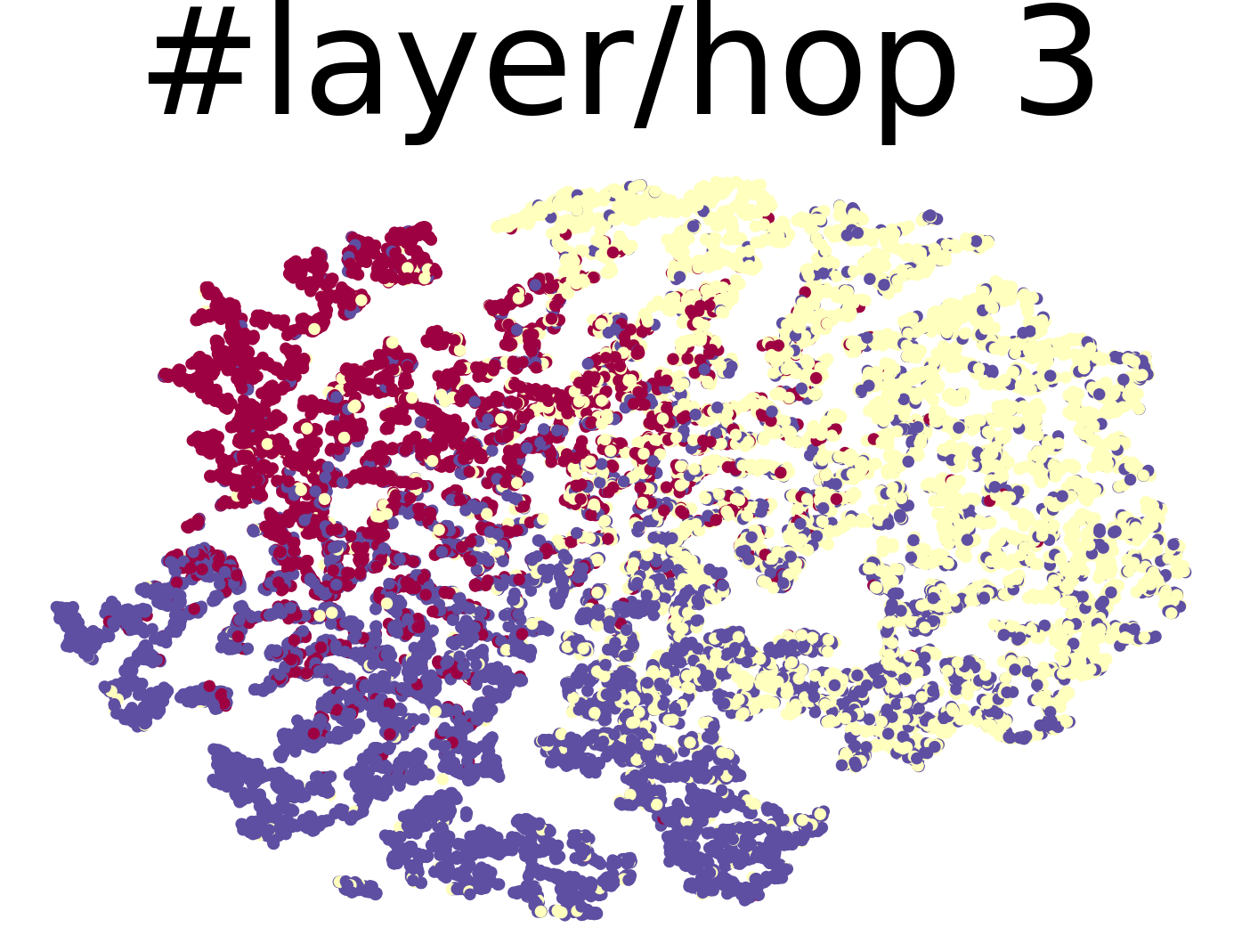}
}
\subfigure{
\includegraphics[width=0.205\columnwidth]{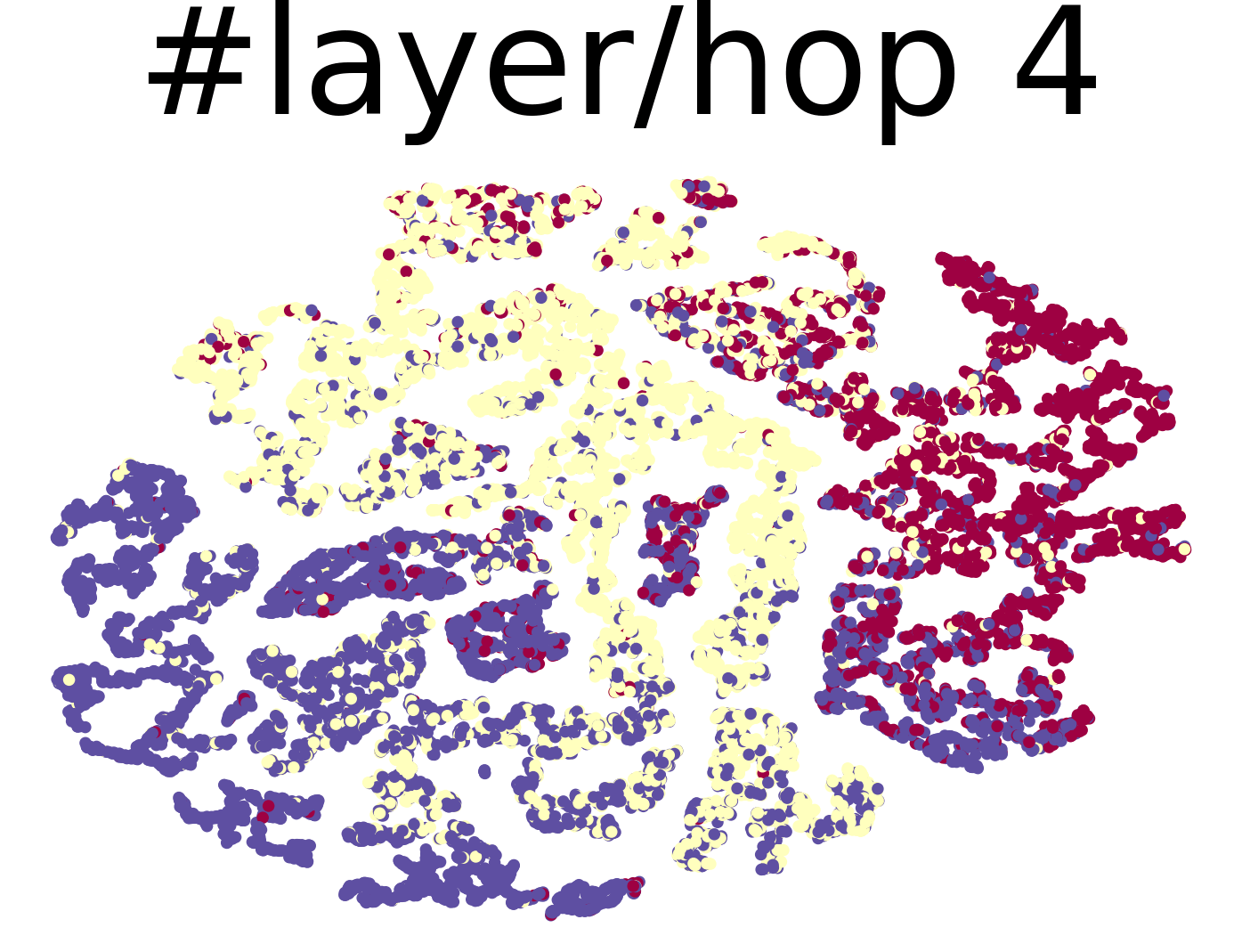}
}
\subfigure{
\includegraphics[width=0.205\columnwidth]{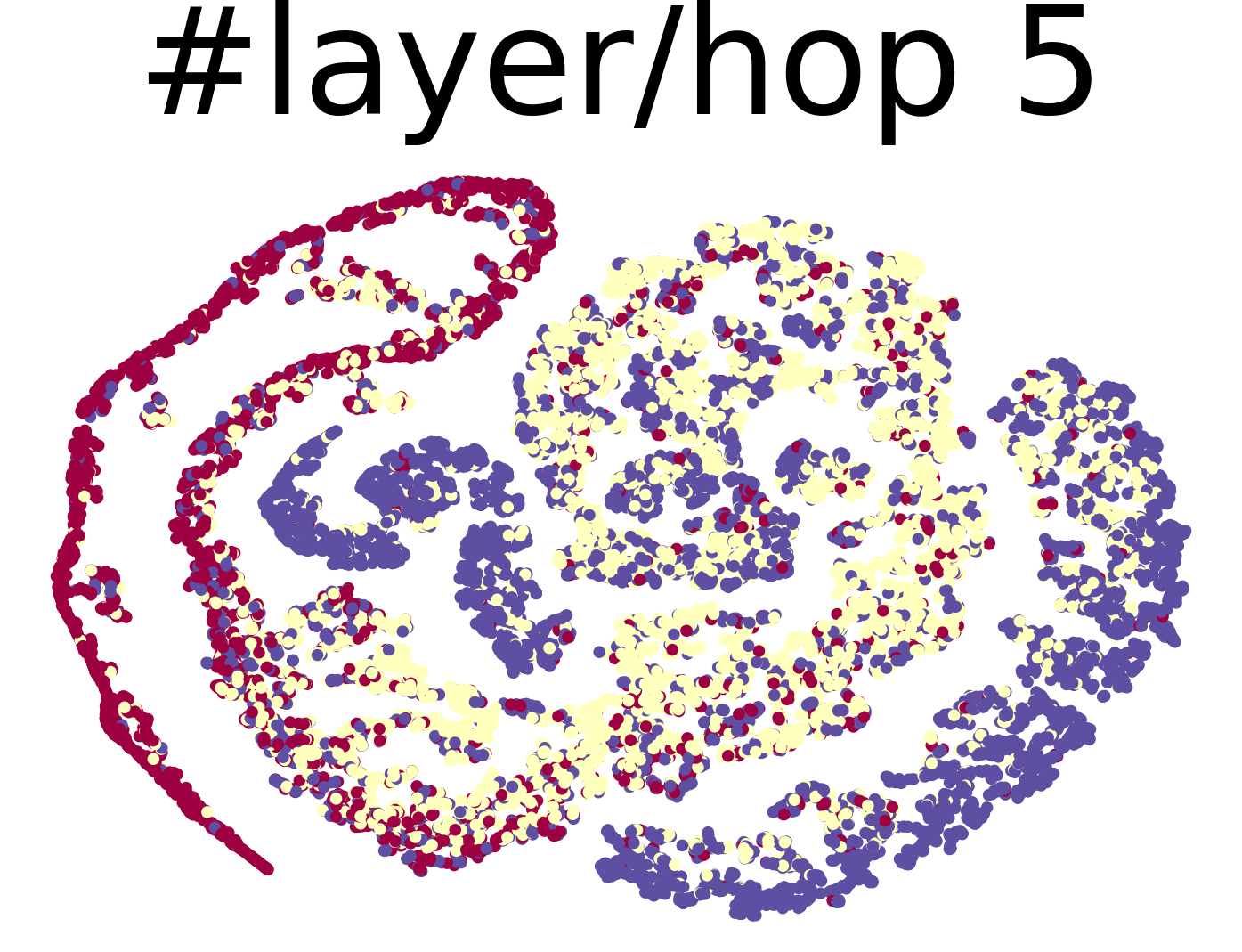}
}
\subfigure{
\includegraphics[width=0.205\columnwidth]{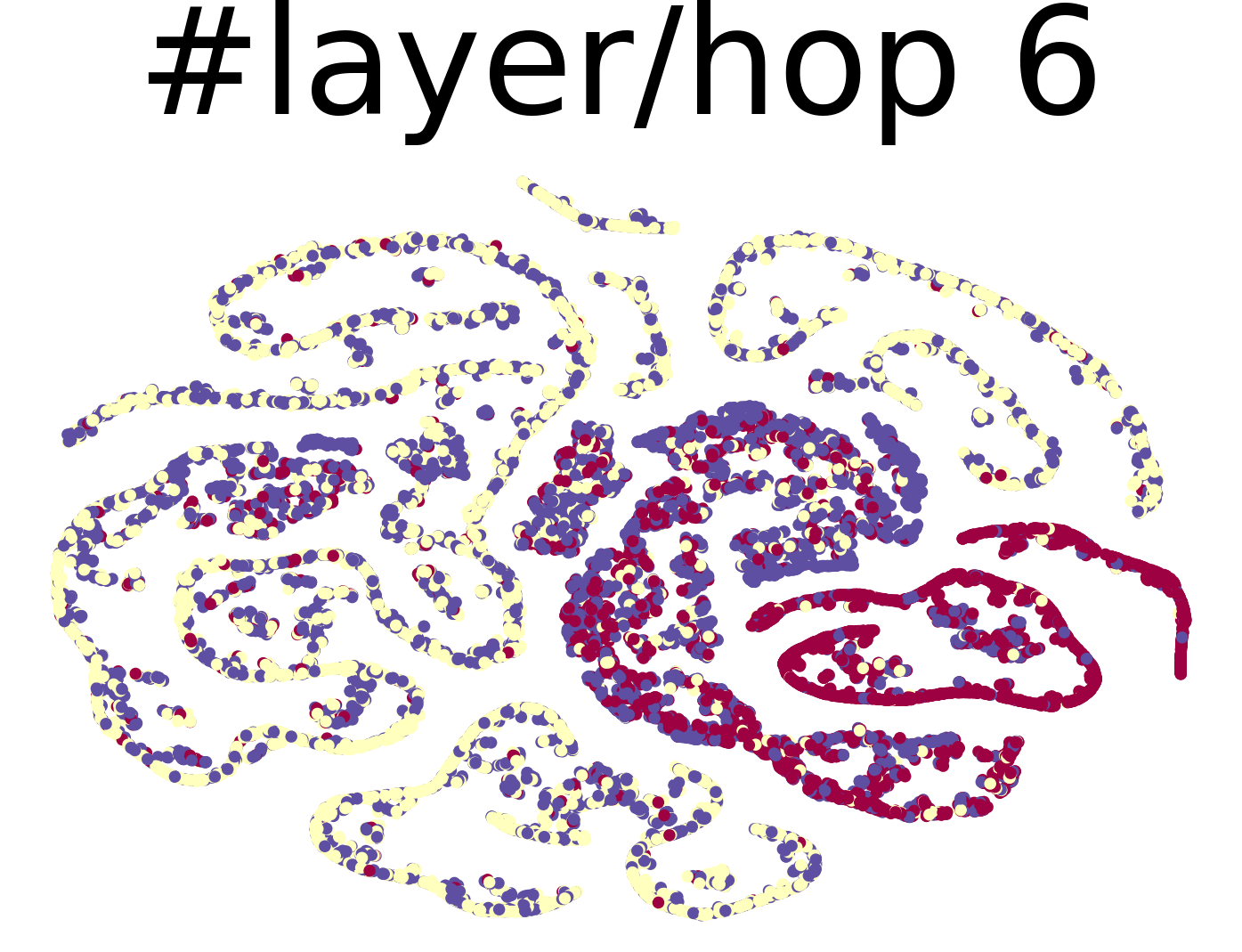}
}
\subfigure{
\includegraphics[width=0.205\columnwidth]{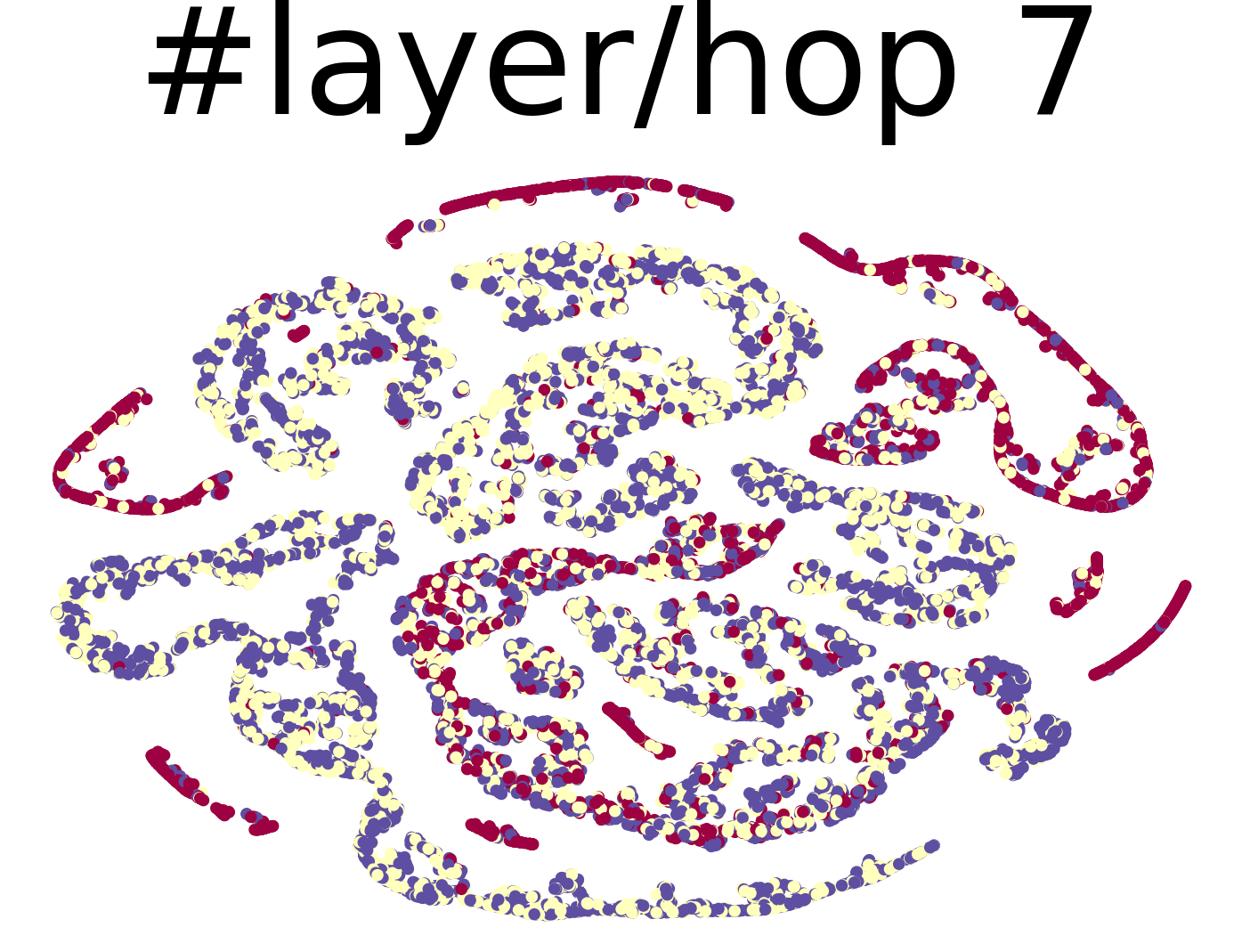}
}
\caption{t-SNE visualization of node representations derived by different numbers of GCN layers on PubMed. Colors represent node classes.}
\label{fig:tsne1_pubmed}
\end{figure*}

The t-SNE visualization of node representations derived by different numbers of GCN layers are shown in Figure \ref{fig:tsne1_citeseer} and \ref{fig:tsne1_pubmed} for CiteSeer and PubMed, respectively. The node representations become indistinguishable when several layers are deployed.

%
%
%

%
%
%
%

\subsection{Visualization of Representations Derived by Models as~Eq.(\ref{DetachGCN})}\label{Sec:visual_citeseer_pubmed_deachedgcn}
\begin{figure*}
\centering
\subfigure{
\includegraphics[width=0.205\columnwidth]{./figures/citeseer_ori_tt.png}
}
\subfigure{
\includegraphics[width=0.205\columnwidth]{./figures/citeseer_mlp_tt.png}
}
\subfigure{
\includegraphics[width=0.205\columnwidth]{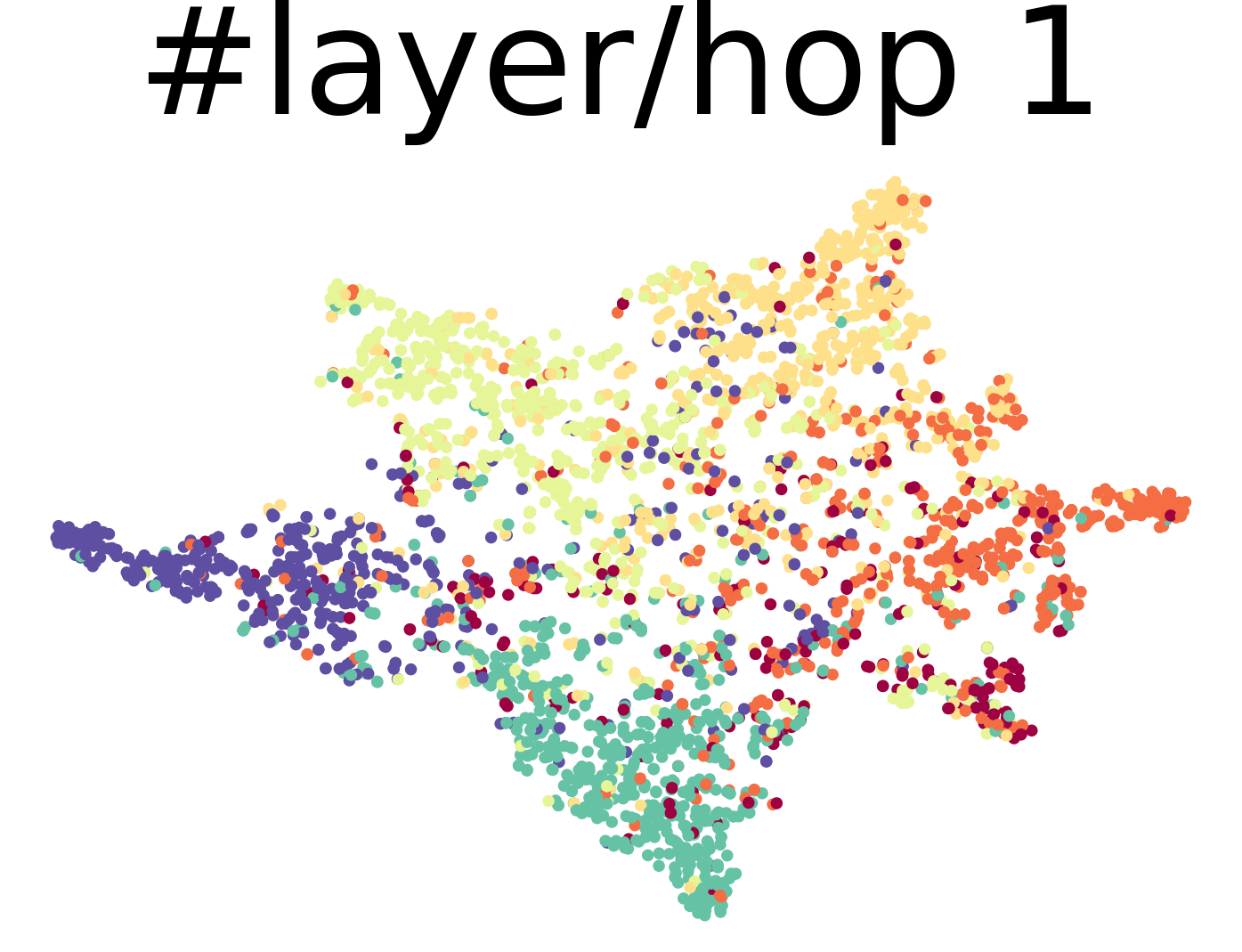}
}
\subfigure{
\includegraphics[width=0.205\columnwidth]{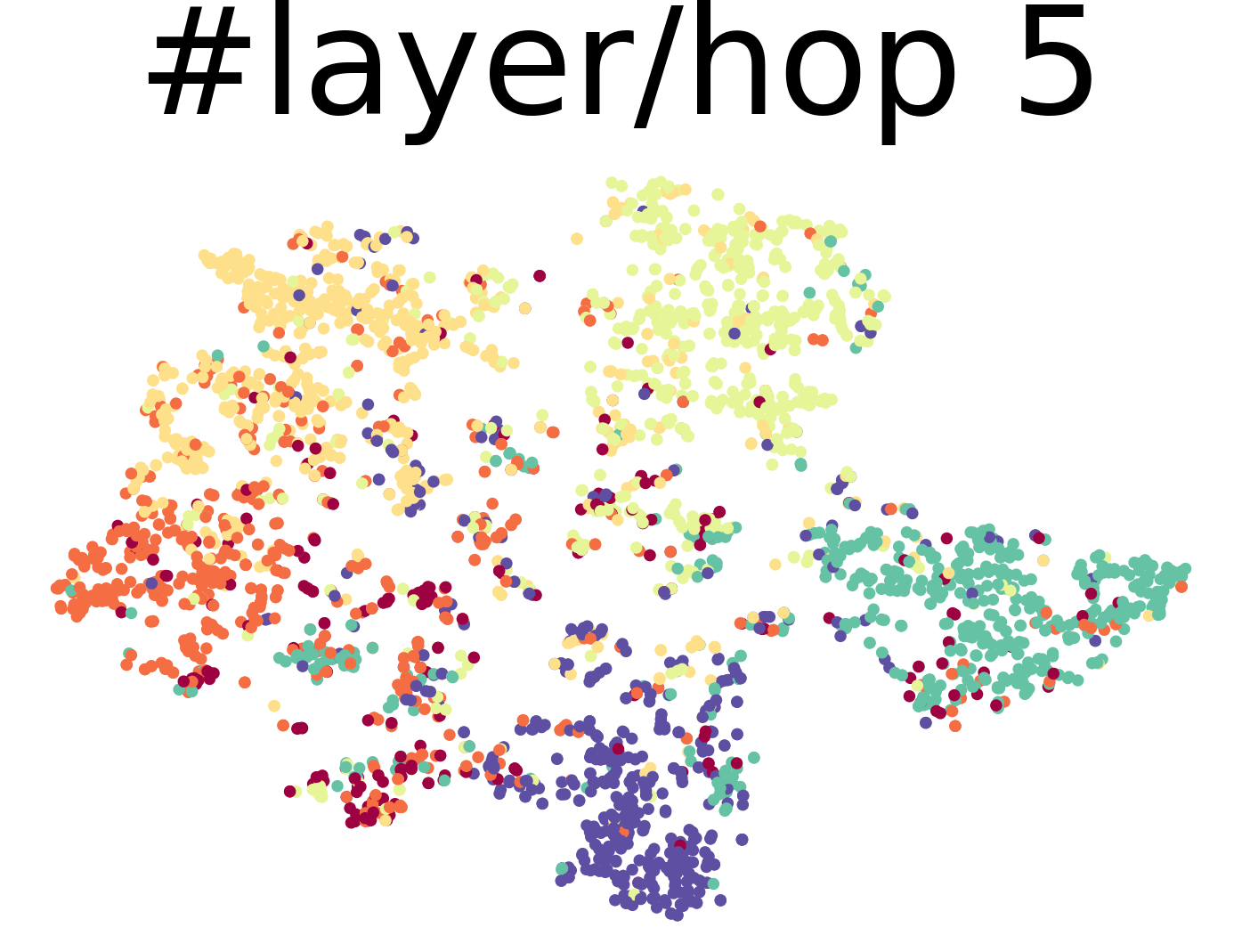}
}
\subfigure{
\includegraphics[width=0.205\columnwidth]{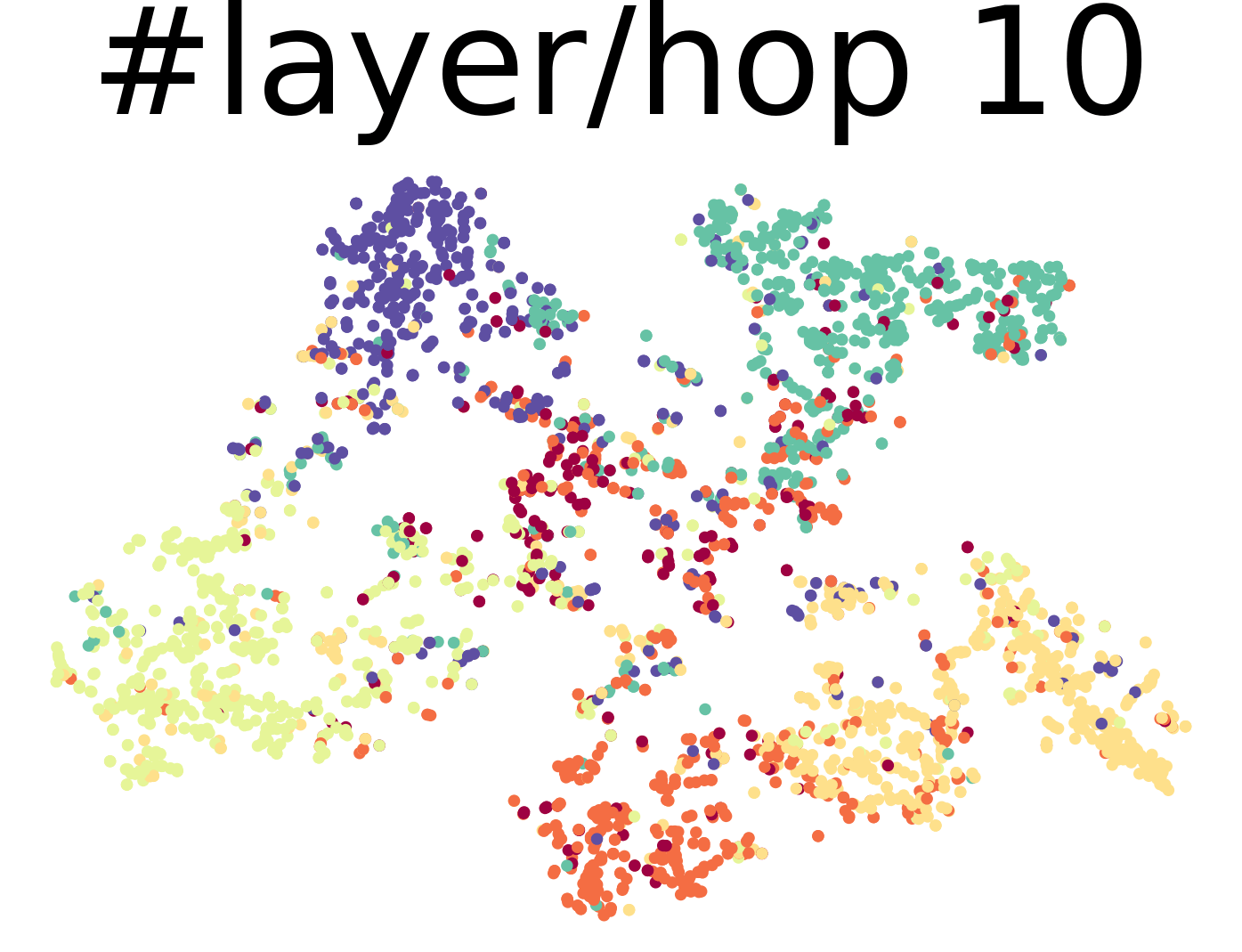}
}
\subfigure{
\includegraphics[width=0.205\columnwidth]{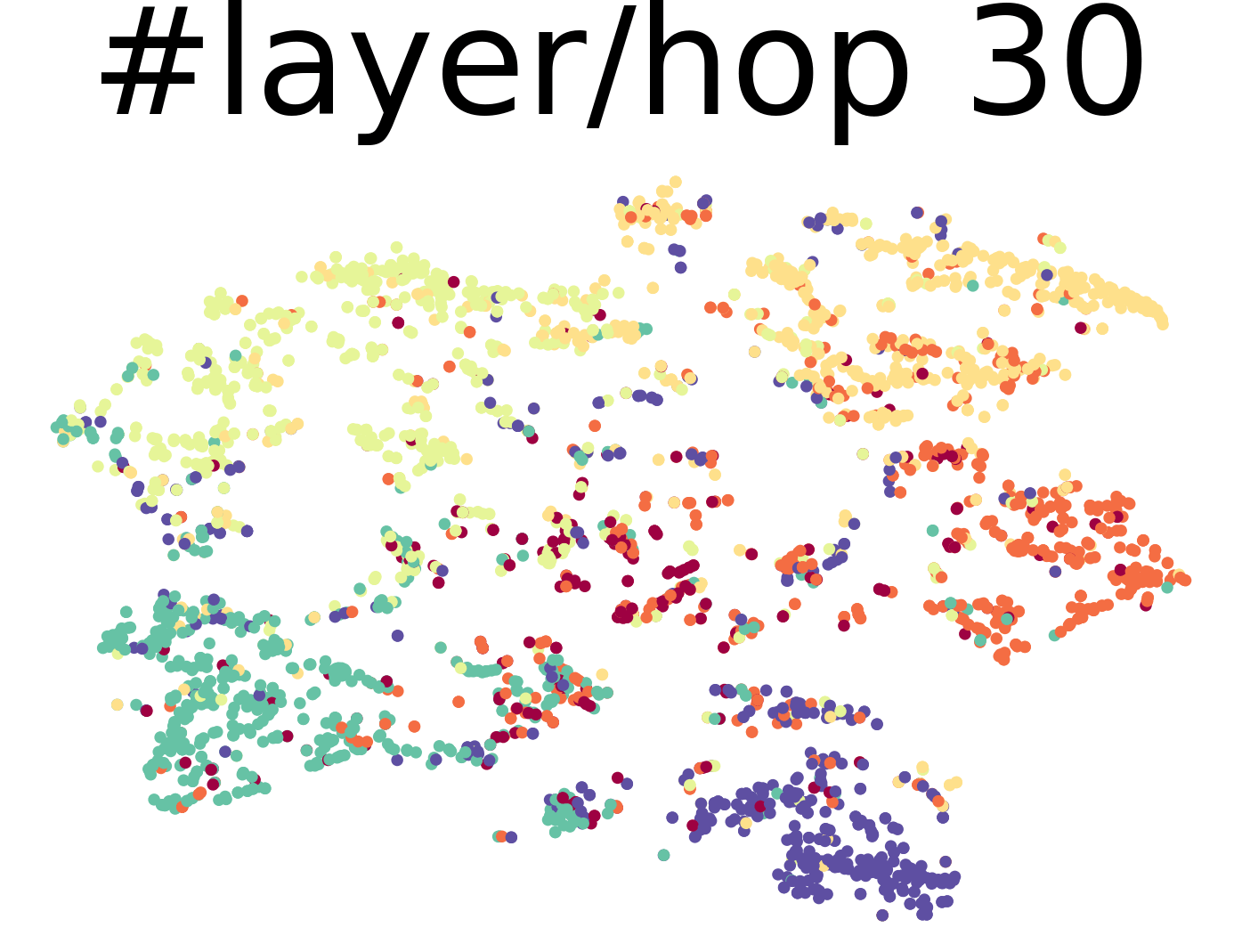}
}
\subfigure{
\includegraphics[width=0.205\columnwidth]{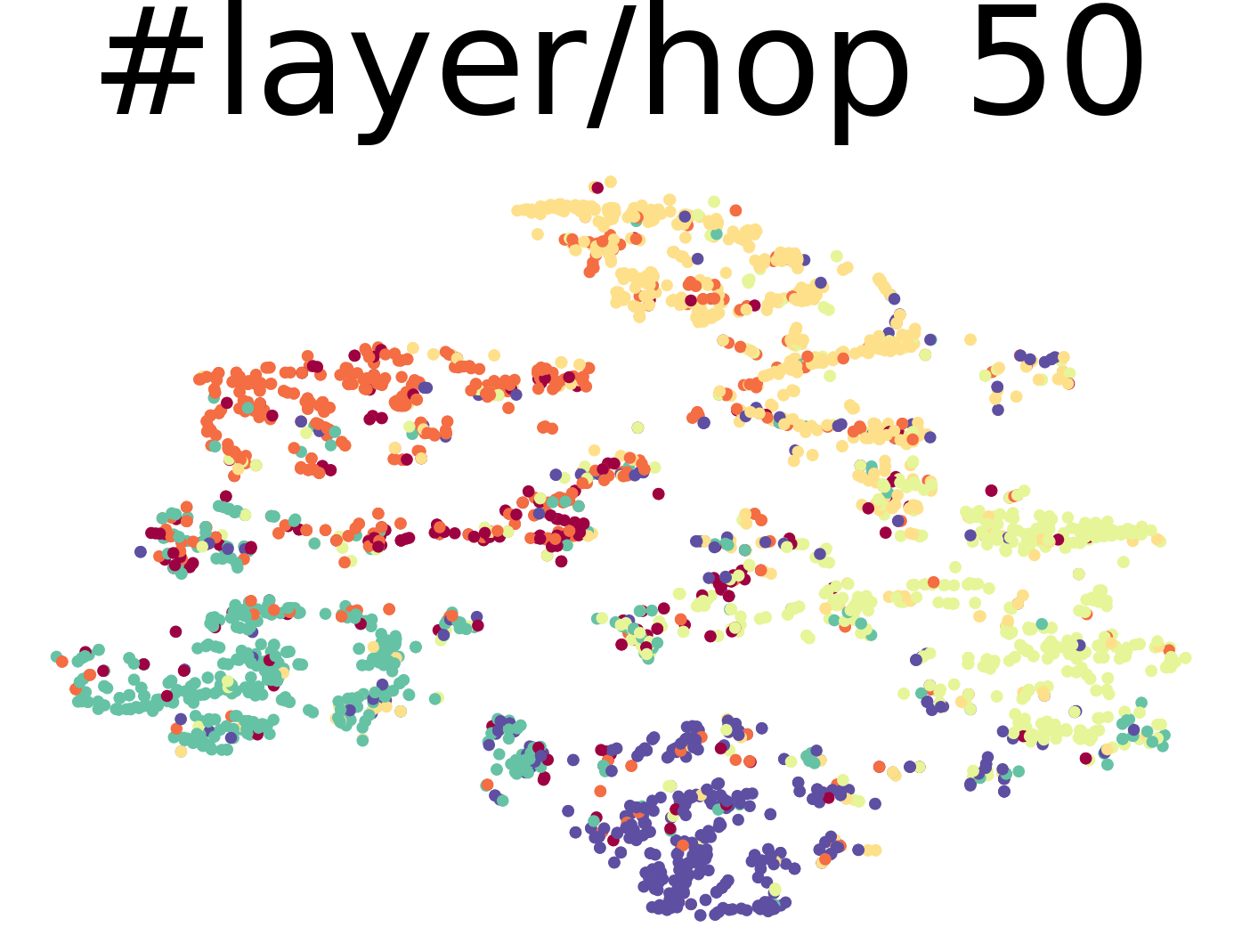}
}
\subfigure{
\includegraphics[width=0.205\columnwidth]{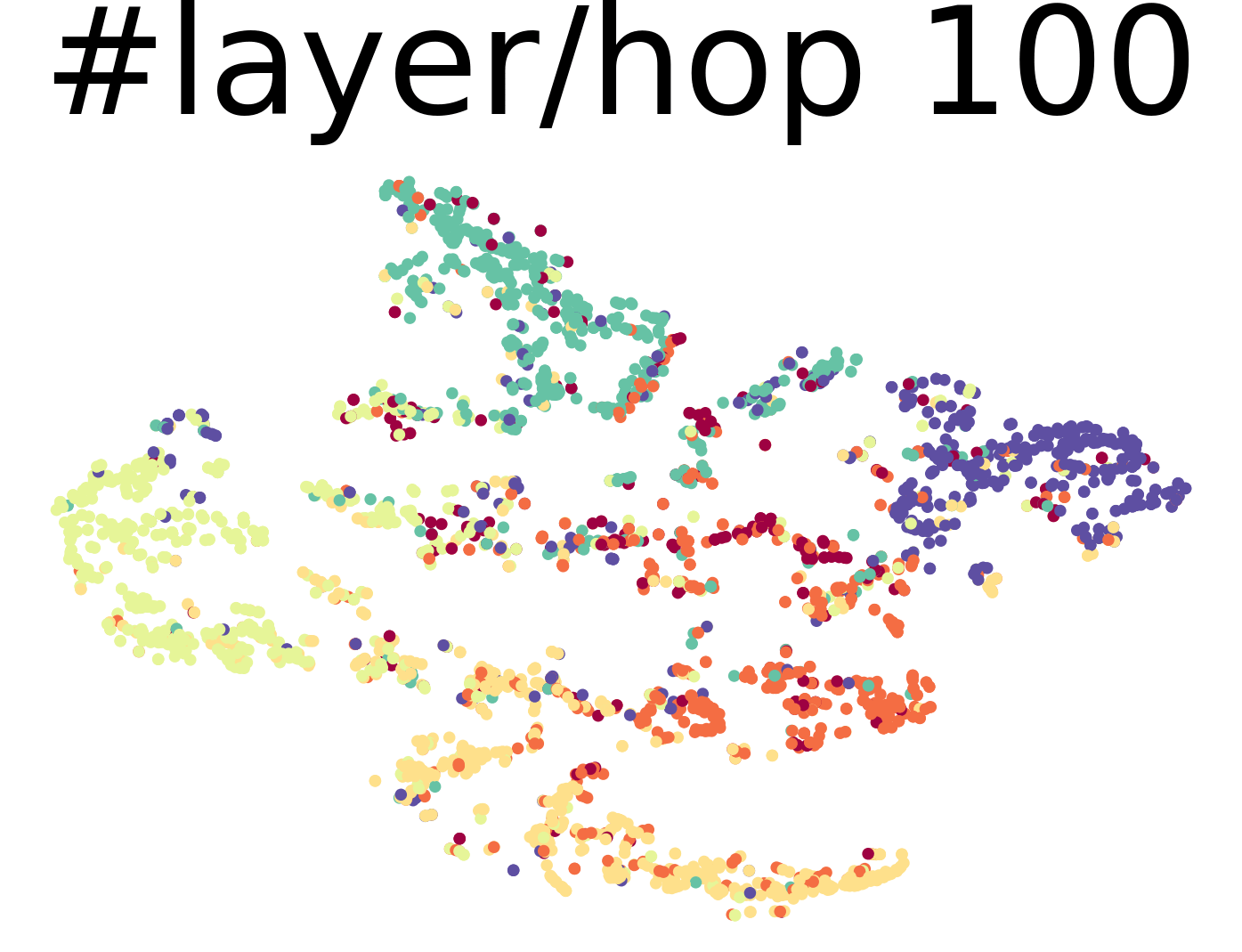}
}
\subfigure{
\includegraphics[width=0.205\columnwidth]{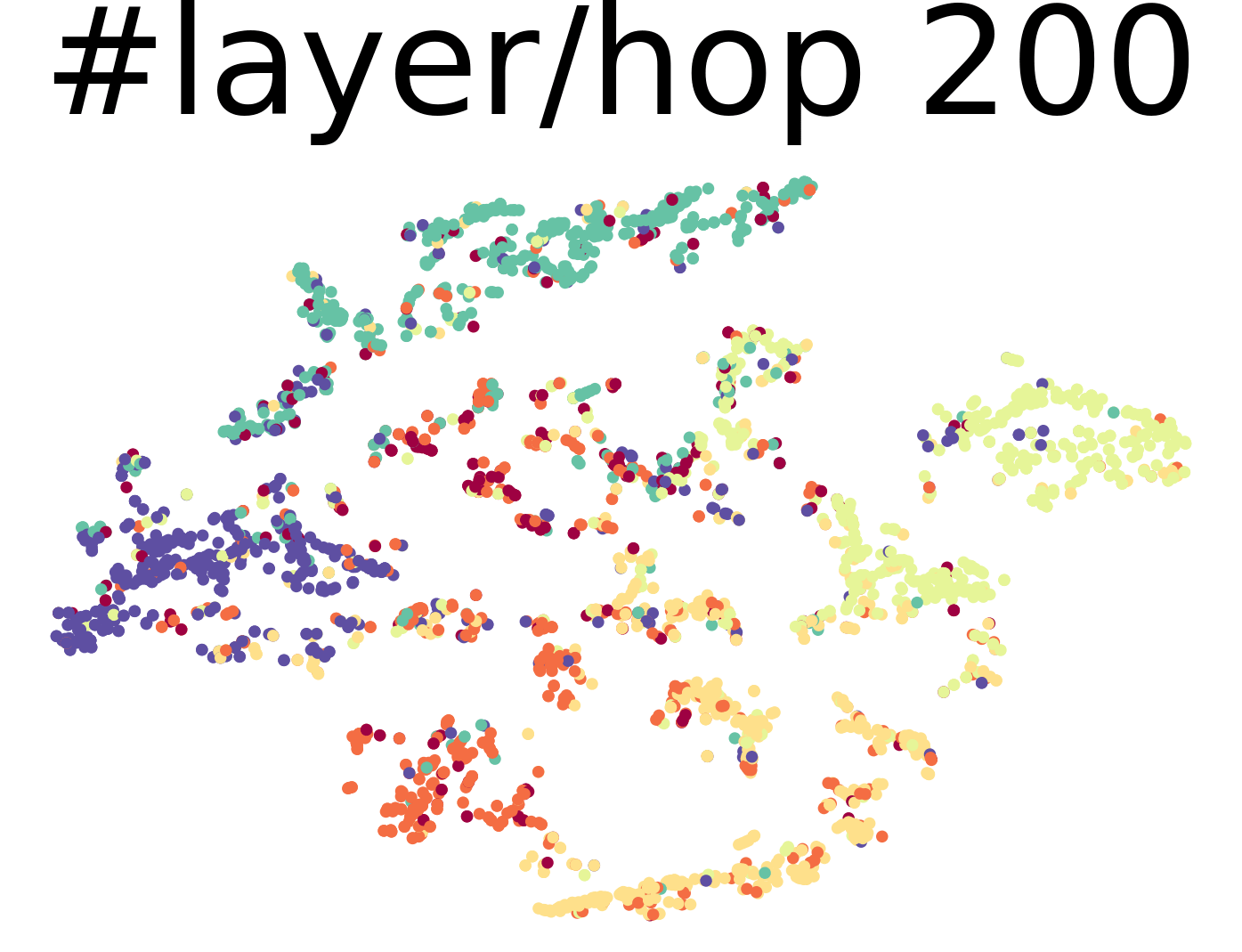}
}
\caption{t-SNE visualization of node representations derived by models as~Eq.(\ref{DetachGCN}) with different numbers of layers on CiteSeer. Colors represent node classes.}
\label{fig:tsne2_citeseer}
\end{figure*}

\begin{figure*}
\centering
\subfigure{
\includegraphics[width=0.205\columnwidth]{./figures/pubmed_ori_tt.png}
}
\subfigure{
\includegraphics[width=0.205\columnwidth]{./figures/pubmed_mlp_tt.png}
}
\subfigure{
\includegraphics[width=0.205\columnwidth]{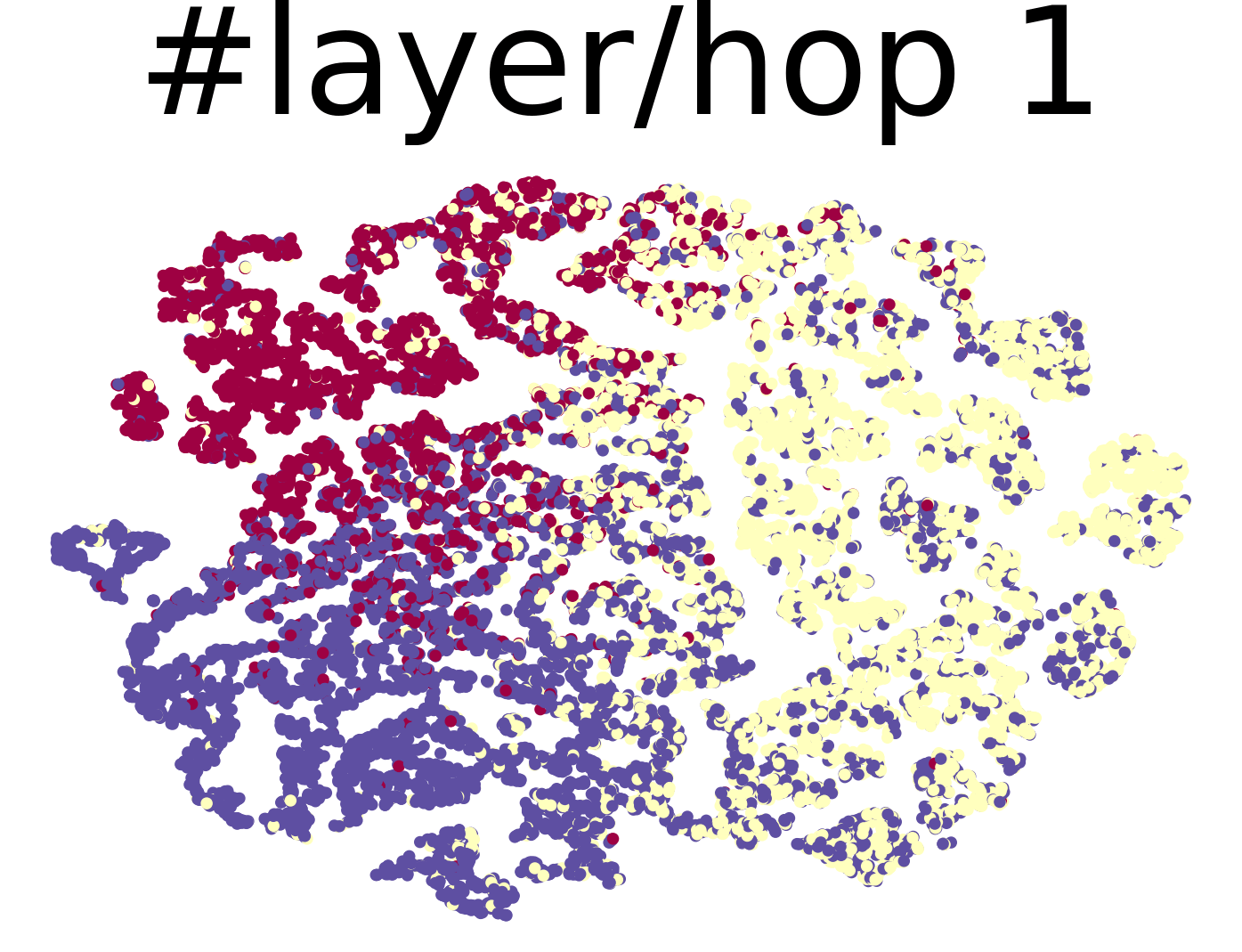}
}
\subfigure{
\includegraphics[width=0.205\columnwidth]{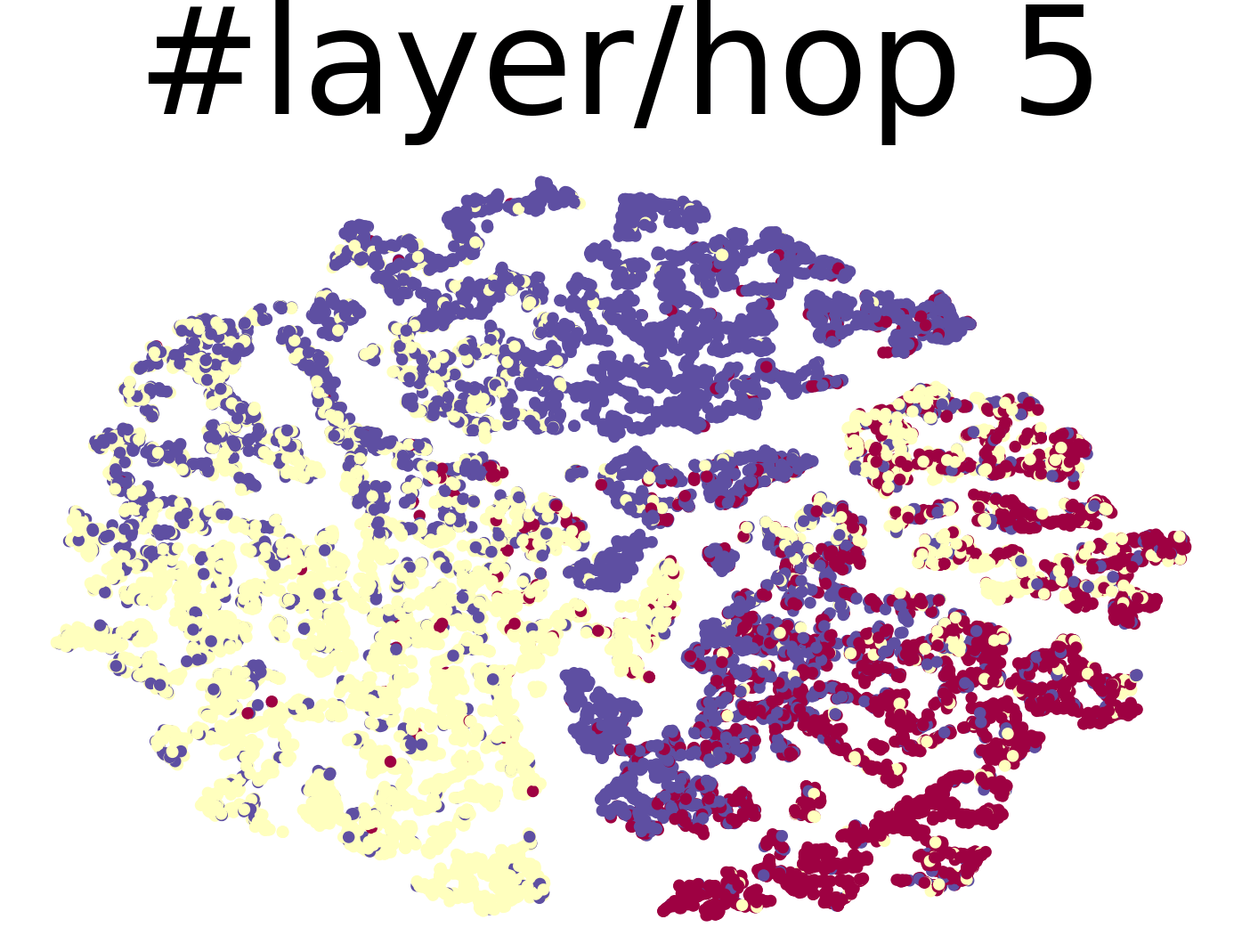}
}
\subfigure{
\includegraphics[width=0.205\columnwidth]{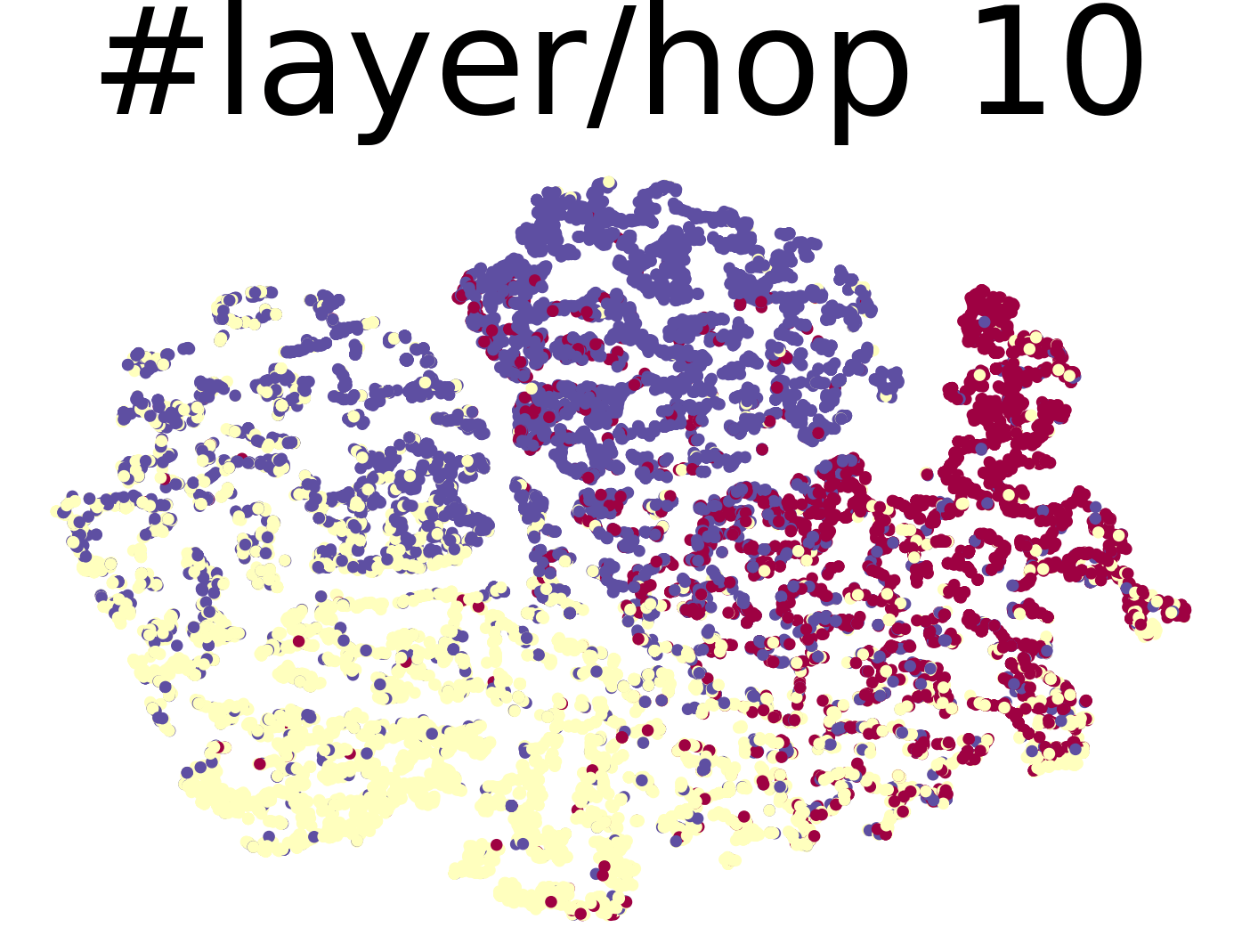}
}
\subfigure{
\includegraphics[width=0.205\columnwidth]{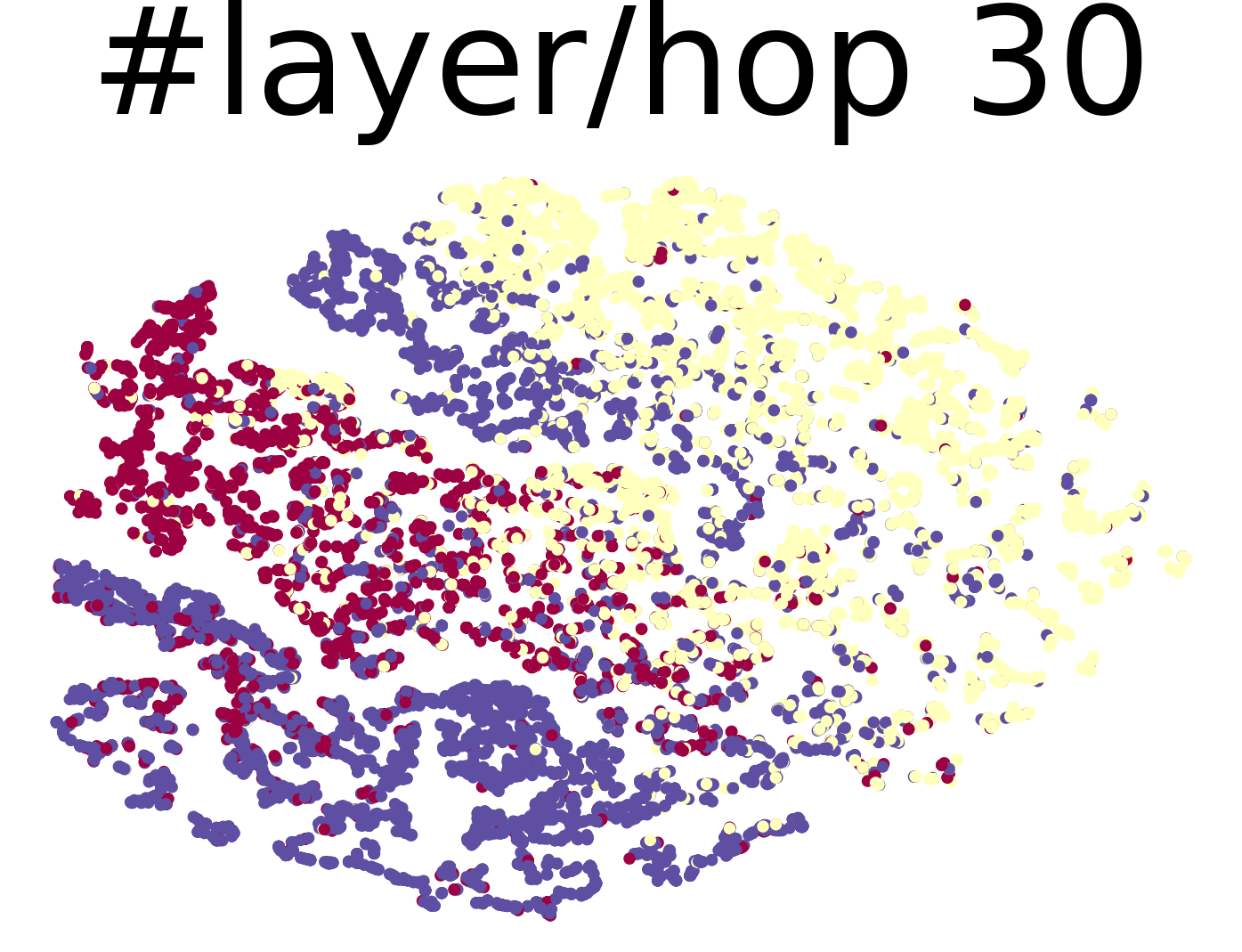}
}
\subfigure{
\includegraphics[width=0.205\columnwidth]{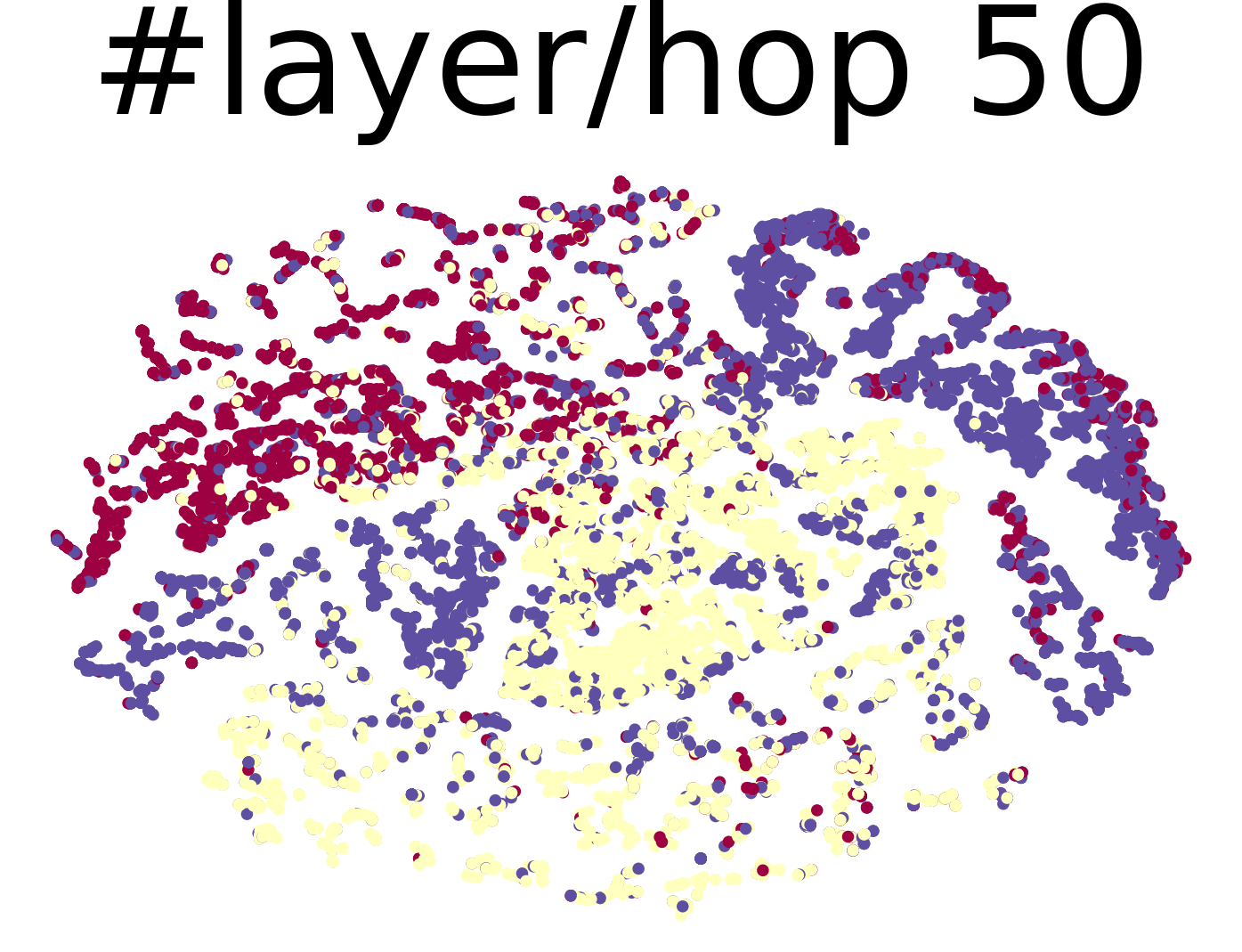}
}
\subfigure{
\includegraphics[width=0.205\columnwidth]{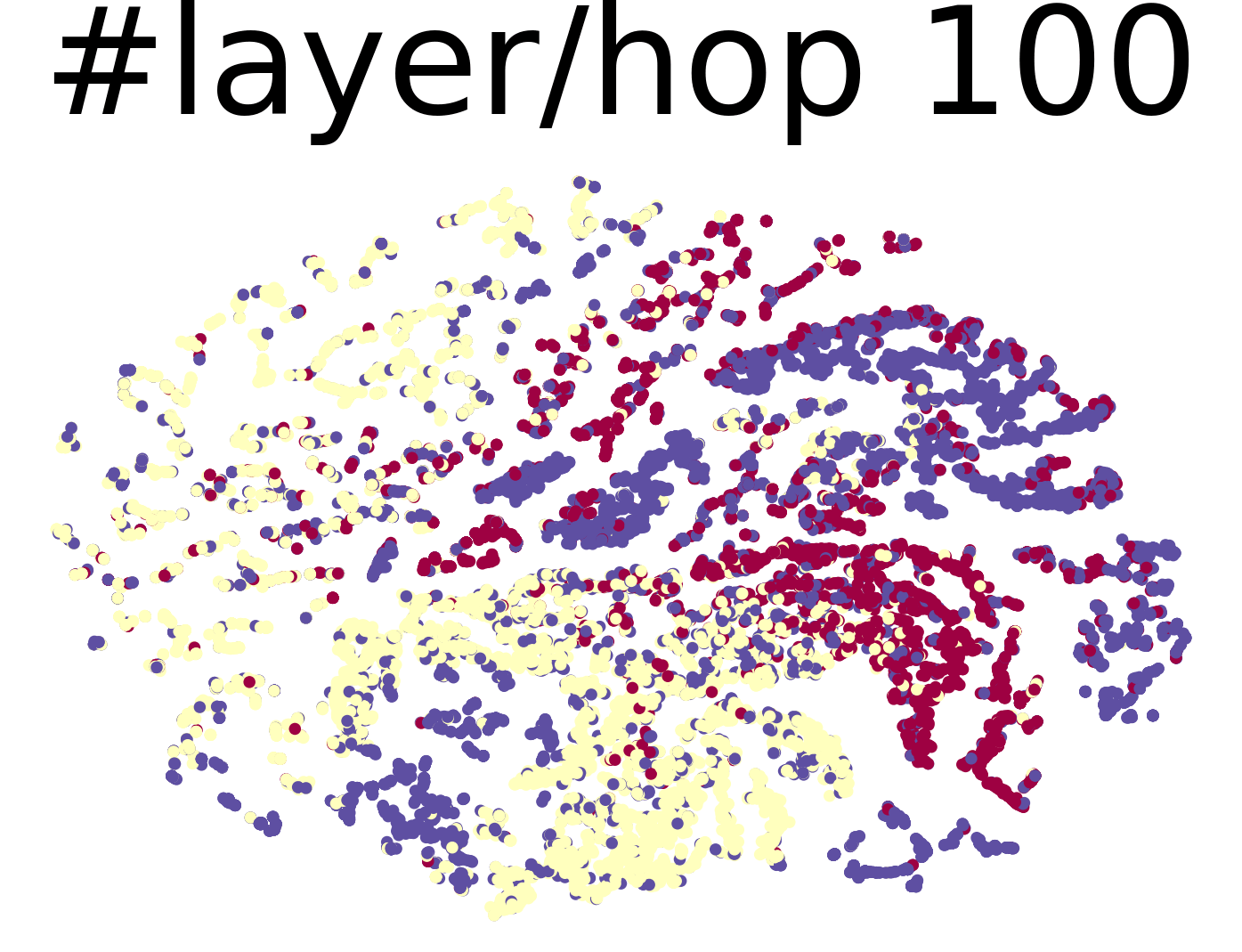}
}
\subfigure{
\includegraphics[width=0.205\columnwidth]{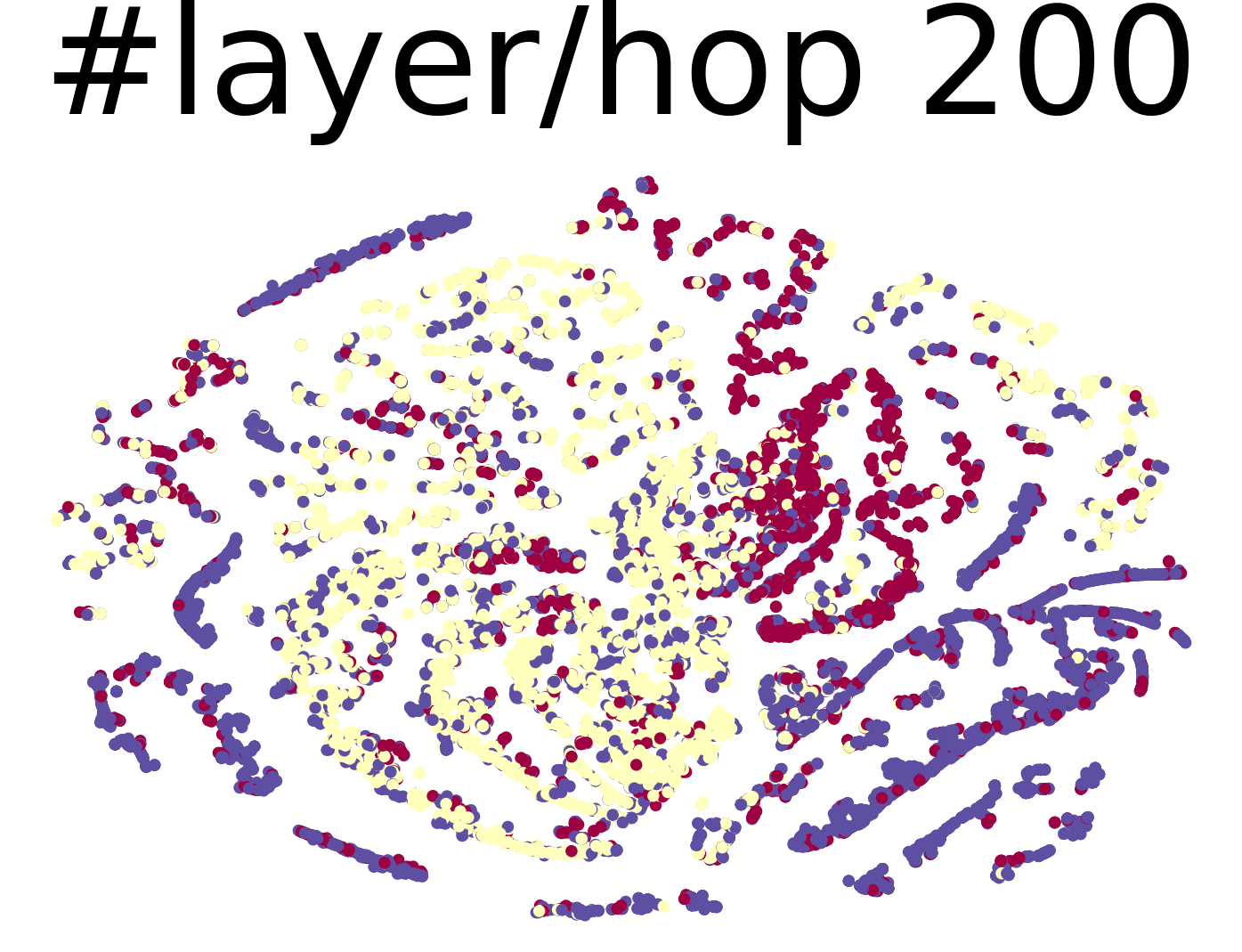}
}
\caption{t-SNE visualization of node representations derived by models as~Eq.(\ref{DetachGCN}) with different numbers of layers on PubMed. Colors represent node classes.}
\label{fig:tsne2_pubmed}
\end{figure*}

The t-SNE visualization of node representations derived by model as~Eq.(\ref{DetachGCN}) with different numbers of layers are shown in Figure \ref{fig:tsne2_citeseer} and \ref{fig:tsne2_pubmed} for CiteSeer and PubMed, respectively. It is shown that after the entanglement of representation transformation and propagation is removed, the model with a large receptive field, such as 50-hop, still generating distinguishable node representations. The over-smoothing issue affects the distinguishability only when an extremely receptive field, like 200-hop, is adopted.

%
%
%
%

\subsection{Proof for Lemma~\ref{lemma1}}\label{Sec:proof1}

\begin{proof}
If $\lambda$ is an eigenvalue of
$\widehat{\boldsymbol{A}}_\oplus$ with left eigenvector
$\boldsymbol{v}_l$ and right eigenvector $\boldsymbol{v}_r$,
we have $\boldsymbol{v}_l
\widehat{\boldsymbol{A}}_\oplus=\lambda\boldsymbol{v}_l$ and
$\widehat{\boldsymbol{A}}_\oplus\boldsymbol{v}_r=\lambda\boldsymbol{v}_r$,
i.e. $\boldsymbol{v}_l
\widetilde{\boldsymbol{D}}^{-1}\widetilde{\boldsymbol{A}}=\lambda\boldsymbol{v}_l$
and
$\widetilde{\boldsymbol{D}}^{-1}\widetilde{\boldsymbol{A}}\boldsymbol{v}_r=\lambda\boldsymbol{v}_r$.
We right multiply the first eigenvalue equation with
$\widetilde{\boldsymbol{D}}^{-\frac{1}{2}}$ and left multiply
the second eigenvalue equation with
$\widetilde{\boldsymbol{D}}^{\frac{1}{2}}$, respectively.
Then we can derive $(\boldsymbol{v}_l
\widetilde{\boldsymbol{D}}^{-\frac{1}{2}})
\widetilde{\boldsymbol{D}}^{-\frac{1}{2}}
\widetilde{\boldsymbol{A}}
\widetilde{\boldsymbol{D}}^{-\frac{1}{2}} =
\lambda(\boldsymbol{v}_l
\widetilde{\boldsymbol{D}}^{-\frac{1}{2}})$ and
$\widetilde{\boldsymbol{D}}^{-\frac{1}{2}}
\widetilde{\boldsymbol{A}}\widetilde{\boldsymbol{D}}^{-\frac{1}{2}}(\widetilde{\boldsymbol{D}}^{\frac{1}{2}}\boldsymbol{v}_r)=\lambda(\widetilde{\boldsymbol{D}}^{\frac{1}{2}}\boldsymbol{v}_r)$.
Hence, $\lambda$ is also an eigenvalue of
$\widehat{\boldsymbol{A}}_\odot$ with left eigenvector
$\boldsymbol{v}_l \widetilde{\boldsymbol{D}}^{-\frac{1}{2}}$
and right eigenvector
$\widetilde{\boldsymbol{D}}^{\frac{1}{2}}\boldsymbol{v}_r$.
From $\widehat{\boldsymbol{A}}_\odot$ to
$\widehat{\boldsymbol{A}}_\oplus$, we can prove it in the
same way.
\end{proof}

\subsection{Proof for Lemma~\ref{lemma2}}\label{Sec:proof2}
\begin{proof}
We first prove that $\widehat{\boldsymbol{A}}_\oplus$ and
$\widehat{\boldsymbol{A}}_\odot$ always have an eigenvalue 1
and all eigenvalues $\lambda$ satisfy $\lvert \lambda \rvert
\leq 1$. We have $\widehat{\boldsymbol{A}}_\oplus
\boldsymbol{e}^T = \boldsymbol{e}^T$ because each row of
$\widehat{\boldsymbol{A}}_\oplus$ sums to $1$. Therefore, $1$
is an eigenvalue of $\widehat{\boldsymbol{A}}_\oplus$.
Suppose that there exists an eigenvalue $\lambda$ that
$\lvert \lambda \rvert > 1$ with eigenvector
$\boldsymbol{v}$, then the length of the right side in
$\widehat{\boldsymbol{A}}_\oplus^k\boldsymbol{v}=\lambda^k\boldsymbol{v}$
grows exponentially when $k$ goes to infinity. This indicates
that some entries of  $\widehat{\boldsymbol{A}}_\oplus^k$
shoulde be larger than $1$. Nevertheless, all entries of
$\widehat{\boldsymbol{A}}_\oplus^k$ are positive and each row
of $\widehat{\boldsymbol{A}}_\oplus^k$ always sums to $1$,
hence no entry of $\widehat{\boldsymbol{A}}_\oplus^k$ can be
larger than $1$, which leads to contradiction. From Lemma \ref{lemma1},
$\widehat{\boldsymbol{A}}_\oplus$ and
$\widehat{\boldsymbol{A}}_\odot$ have the same eigenvalues.
Therefore, $\widehat{\boldsymbol{A}}_\oplus$ and
$\widehat{\boldsymbol{A}}_\odot$ always have an eigenvalue 1
and all eigenvalues $\lambda$ satisfy $\lvert \lambda \rvert
\leq 1$.

According to the Perron-Frobenius Theorem for Primitive
Matrices~\cite{seneta2006non}, there exists an eigenvalue $r$
for an $n\times n$ non-negative primitive matrix such that
$r>\lvert \lambda \rvert$ for any eigenvalue $\lambda\neq r$
and the eigenvectors associated with $r$ are unique. The
property that the given graph is connected can guarantee that
for $\forall i, j$: $\exists k$ s.t.
$\widehat{\boldsymbol{A}}^k_\oplus[i,j]>0$. Furthermore,
there must exsit some $k$ that can make all entries of
$\widehat{\boldsymbol{A}}^k$ to be simultaneously positive
because self-loops are included in the graph. Formally,
$\exists k$: $\widehat{\boldsymbol{A}}^k_\oplus[i,j]>0$ for
$\forall i, j$. Hence, $\widehat{\boldsymbol{A}}_\oplus$ is a
non-negative primitive matrix. From the Perron-Frobenius
Theorem for Primitive Matrices,
$\widehat{\boldsymbol{A}}_\oplus$ always has an eigenvalue
$1$ with unique associated eigenvectors and all other
eigenvalues $\lambda$ satisfy $\lvert \lambda \rvert < 1$.
Based on Lemma \ref{lemma1}, this property can be extended to
$\widehat{\boldsymbol{A}}_\odot$. We then compute the
eigenvectors associated with eigenvalue $1$. Obviously,
$\boldsymbol{e}^T$ is the right eigenvector of
$\widehat{\boldsymbol{A}}_\oplus$ associated with eigenvalue
$1$. Next, assume $\boldsymbol{v}_l$ is the left eigenvector
of $\widehat{\boldsymbol{A}}_\oplus$ associated with
eigenvalue $1$ and thus
$\boldsymbol{v}_l\widetilde{\boldsymbol{D}}^{-\frac{1}{2}}$
is the left eigenvector of $\widehat{\boldsymbol{A}}_\odot$
associated with eigenvalue $1$. We know
$\widehat{\boldsymbol{A}}_\odot$ is a symmetric matrix, whose
left and right eigenvectors associated with the same
eigenvalue are simply each other's transpose. Hence, we
utilize
$\boldsymbol{v}_l\widetilde{\boldsymbol{D}}^{-\frac{1}{2}} =
(\widetilde{\boldsymbol{D}}^{\frac{1}{2}}\boldsymbol{e}^T)^T$
to obtain  $\boldsymbol{v}_l =
\boldsymbol{e}\widetilde{\boldsymbol{D}}$. After deriving the
eigenvectors of $\widehat{\boldsymbol{A}}_\oplus$ associated
with eigenvalue $1$, corresponding eigenvectors of
$\widehat{\boldsymbol{A}}_\odot$ can be computed by Lemma
\ref{lemma1}.
\end{proof}

\subsection{Datasets Description and Statistics}\label{Sec:dataset}

\textbf{Citation datasets.} Cora, CiteSeer and PubMed ~\cite{sen2008collective} are representative citation network datasets where nodes and edges denote documents and their citation relationships, respectively. Node features are formed by bay-of-words representations for documents. Each node has a label indicating what field the corresponding document belongs to.

\textbf{Co-authorship datasets.} Coauthor CS and Coauthor Physics ~\cite{shchur2018pitfalls} are co-authorship graphs datasets. Nodes denote authors, which are connected by an edge if they co-authored a paper. Node features represent paper keywords for each author's papers. Each node has a label denoting the most active fields of study for the corresponding author.

\textbf{Co-purchase datasets.} Amazon Computers and Amazon Photo ~\cite{shchur2018pitfalls} are segments of the Amazon co-purchase graph~\cite{mcauley2015image} where nodes are goods and edges denote that two goods are frequently bought together. Node features are derived from bag-of-words representations for product reviews and class labels are given by the product category.

The datasets statistics are summarized in Table \ref{tab:dataset}.

\end{document}